\documentclass{article} %
\usepackage{iclr2027_conference,times}
\usepackage{amsthm}
\usepackage{url}
\usepackage{enumitem}
\usepackage{bbm}
\usepackage{lipsum}
\usepackage{graphicx}
\usepackage[most]{tcolorbox}
\usepackage{soul}
\usepackage{caption}
\usepackage{subcaption}
\usepackage{tablefootnote}
\usepackage{wrapfig}
\usepackage{lipsum}
\usepackage{enumitem}
\usepackage{multirow}
\usepackage{tikz}
\usepackage{mathtools}
\usepackage{booktabs}
\usepackage{amsmath}
\usepackage{xcolor}
\usetikzlibrary{arrows.meta}
\usepackage{basic_commands}
\usepackage{algorithm}
\usepackage{tabularew}
\usepackage[para,online,flushleft]{threeparttable}
\usepackage[endLComment=,italicComments=false]{algpseudocodex}

\AddToHook{cmd/ttfamily/after}{\frenchspacing}

\theoremstyle{plain}

\theoremstyle{definition}

\theoremstyle{remark}

\definecolor{myred}{HTML}{F54254}
\definecolor{myorange}{HTML}{FFB135}
\definecolor{mygreen}{HTML}{10BD35}
\definecolor{myblue}{HTML}{598BE7}
\definecolor{mypurple}{HTML}{9A1C6B}
\definecolor{plgray}{HTML}{999999}
\renewcommand{\tt}[1]{\texttt{#1}}
\renewcommand{\phi}{\varphi}

\newcommand{\pl}[1]{{\color{plgray} #1}}

\hypersetup{urlcolor=myblue,citecolor=myblue,linkcolor=myblue}

\setlist[itemize]{itemsep=1pt, leftmargin=20pt}

\newcommand{\cutsectionup}{\vspace{0pt}}
\newcommand{\cutsectiondown}{\vspace{0pt}}
\newcommand{\cutsubsectionup}{\vspace{0pt}}
\newcommand{\cutsubsectiondown}{\vspace{0pt}}

\newcommand{\ocbenchrepo}{\url{https://github.com/seohongpark/ocbench}}
\newcommand{\ocbenchblog}{\url{https://seohong.me/blog/behavioral-cloning-mystery}}
\newcommand{\ocbenchwebsite}{\url{https://seohong.me/projects/ocbench}}

\title{Behavioral Cloning Mystery}

\author{Seohong Park$^{1}$ \quad Sergey Levine$^{1}$ \\
$^{1}$University of California, Berkeley \\
\texttt{seohong@berkeley.edu}
}

\iclrfinalcopy %
\begin{document}

\maketitle

\begin{abstract}
Behavioral cloning (BC), despite its simplicity, exhibits many counterintuitive phenomena in the real world.
For example, the performance of BC often keeps increasing as the model overfits more to the dataset,
and fully closed-loop policies often completely fail without action chunking.
Unfortunately, properly studying these anecdotal phenomena (``behavioral cloning mysteries'') is challenging:
in the real world, datasets and experiments are costly and not fully controllable;
in simulation with synthetic data, these phenomena are often not easily observed
partly due to the discrepancy between scripted policies and human demonstrations.
In this work, we propose OCBench,
a robotic manipulation benchmark with controllable scripted policies that have similar properties to human demonstrations.
We show that, by mimicking key properties of human demonstrations,
OCBench reproduces many anecdotal BC-related phenomena in controlled settings.
With its GPU-accelerated environments and scripted policies,
we demonstrate how OCBench enables scientific studies of previously reported BC-related phenomena
by analyzing and refuting various hypotheses.

Blog post: \ocbenchblog

Project page: \ocbenchwebsite

Repository: \ocbenchrepo
\end{abstract}

\cutsectionup
\cutsectionup
\section{Introduction}
\cutsectiondown

Behavioral cloning (BC) is a foundational technique for large-scale robotic control~\citep{brohan2023rt, intelligence2026pi}.
Despite its simplicity,
behavioral cloning exhibits many counterintuitive phenomena (``mysteries''),
especially when trained on complex, large-scale, human-collected data.
For example, researchers have anecdotally found that
BC performance often increases as it overfits more~\citep{mandlekar2021matters, allshire2026scalable},
fully closed-loop policies often fail completely whereas partially open-loop control with action chunking works well~\citep{zhao2023learning, lazzati2026does, zeng2026revisiting},
and history conditioning can hurt performance despite providing strictly more information~\citep{de2019causal, wen2020fighting, mark2026bpp}.

Unfortunately, properly studying these empirical phenomena is challenging.
In the real world, robotic experiments are not perfectly reproducible,
human-collected data are expensive and less controllable,
and different factors cannot be easily isolated.
Simulated benchmarks with scripted policies provide more control over data,
but they often do not reproduce all these real-world phenomena (see \Cref{sec:motivation}),
likely because previous scripted policies do not have the same properties as actual human demonstrations.

The main goal of this work is to provide a suite of simulated tasks and datasets,
which we call \textbf{OCBench}, %
that has similar \emph{structural properties} to human demonstrations,
to enable scientific, controlled studies of realistic behavioral cloning and robotic control.
We show that by mimicking the key properties of human demonstrations
(\eg, trajectories being very narrow, non-Markovian, and smoothly random),
we can reproduce many anecdotal BC-related phenomena in controlled settings,
including those that are not clearly observed in previous synthetic benchmarks.

We implement the OCBench tasks and scripted policies in MJWarp, a GPU-accelerated MuJoCo based on NVIDIA Warp~\citep{macklin2022warp}.
On a single H100 GPU,
OCBench can collect synthetic trajectories highly efficiently, at up to $200$K environment steps per second.
This makes \emph{streaming-data} training practically feasible,
meaning that we can train an agent using a continuous stream of freshly generated data.
This allows us to study the ``limit'' of different algorithms and design choices under effectively unlimited data.

In this paper, based on our controlled reproduction in OCBench,
we discuss four non-intuitive phenomena that happen with real-world behavioral cloning.
We provide (and refute) possible hypotheses and discuss implications of each mystery,
demonstrating how OCBench's controlled datasets and streaming-data training enable scientific studies
of anecdotal phenomena reported in the literature.

\textbf{Contributions.}
The main contributions of this work are twofold.
First, we introduce OCBench,
a new GPU-accelerated benchmark suite accompanied by scripted policies that mimic the structural properties of human demonstrations.
This enables \emph{controlled} studies of diverse phenomena observed in real-world behavioral cloning.
Second, we reproduce four previously reported, non-intuitive phenomena in OCBench,
making them scientifically studyable.
We also provide analyses and hypotheses for these phenomena with concrete evidence in controlled settings.

\section{Related work}

\textbf{Behavioral cloning.}
Modern large-scale robotic control is mainly based on behavioral cloning and imitation learning~\citep{brohan2023rt, kim2024openvla, bjorck2025gr00t, intelligence2026pi}.
Among diverse behavioral cloning techniques, including those based on multi-layer perceptrons (MLPs)~\citep{fu2020d4rl, mandlekar2021matters},
tokenized actions~\citep{shafiullah2022behavior, brohan2023rt},
and energy-based models~\citep{florence2022implicit},
the most popular approaches to date are diffusion~\citep{chi2025diffusion} and flow matching~\citep{black2024pi_0}.
Researchers have scaled up flow behavioral cloning with large transformers,
demonstrating its effectiveness across a wide array of robotic tasks~\citep{bjorck2025gr00t, intelligence2026pi}.
Accordingly, we mainly focus on flow behavioral cloning in this work,
although many of the phenomena studied in this work may apply to other behavioral cloning techniques as well.

\textbf{Analyses of behavioral cloning.}
Several works have analyzed diverse phenomena and failure modes in behavioral cloning.
One line of work has focused on test-time distribution shift and DAgger
from both theoretical and empirical perspectives~\citep{ross2011reduction, spencer2021feedback, park2024value, simchowitz2025pitfalls}.
Another body of work has analyzed the effect of modern techniques,
such as action chunking and open-loop control~\citep{zhang2026action, lazzati2026does, zeng2026revisiting},
as well as history conditioning~\citep{de2019causal, wen2020fighting, mark2026bpp}.
Many of the aforementioned works rely primarily on relatively small, fixed, human-collected datasets
(\eg, $50$--$300$ demonstrations per task).
In contrast, recent large-scale robotic systems train on orders of magnitude more robot data~\citep{o2024open, intelligence2026pi}.
The difference in data scale often serves as a confounder,
making it unclear whether these issues will eventually go away with more data or are fundamental to behavioral cloning.
In this work, rather than relying on limited human-collected data,
we mimic the structural properties of human demonstrations with GPU-accelerated scripted policies,
enabling effectively unlimited data generation at up to $200$K steps per second,
while reproducing many phenomena observed in real-world behavioral cloning.
To the best of our knowledge, this is the first work to systematically reproduce multiple real-world BC phenomena
using GPU-accelerated,
fully controllable scripted policies.

\textbf{Behavioral cloning and reinforcement learning benchmarks.}
A variety of simulated benchmarks have been proposed for behavioral cloning and data-driven reinforcement learning (RL).
Previous behavioral cloning benchmarks typically provide
simulated tasks with tens to a few hundred human-collected demonstrations per task~\citep{mandlekar2021matters, liu2023libero}.
Unlike them, we instead design scripted policies that mimic human data's properties to enable unlimited generation of controlled trajectories.
Among existing BC benchmarks, MimicGen~\citep{mandlekar2023mimicgen} is particularly relevant,
which supports systematic human-like data generation.
Unlike MimicGen,
which generates trajectories by first recording about $10$ human trajectories per task and ``replaying'' their segments in new randomized configurations,
OCBench generates data using fully scripted policies having human-like properties
and provides a host of controllable knobs (\eg, motion speed, mistake probability, coverage degree, etc.),
enabling both diverse data generation and scientific analyses.
Another line of simulated benchmarks~\citep{fu2020d4rl, james2020rlbench, yarats2022don, lu2022challenges, tao2025demonstrating, park2025ogbench}
provides scripted or RL-trained policies to generate data.
Compared to these previous benchmarks, OCBench provides more human-like scripted policies,
enabling us to reproduce diverse phenomena that happen in real-world behavioral cloning
but are not easily observed in previous scripted benchmarks (\Cref{sec:motivation}).
We also provide GPU-accelerated environments and scripted policies based on MJWarp.
OCBench's closest predecessor is OGBench~\citep{park2025ogbench}, and we compare them in detail in \Cref{sec:prev_bench}.
\cutsectionup
\section{Motivation}
\label{sec:motivation}
\cutsectiondown

Behavioral cloning, despite its simplicity, exhibits several counterintuitive phenomena (``mysteries''),
especially when using real-world data in robotics and autonomous driving.
Below are some of these phenomena that have been anecdotally reported in the literature.

\textbf{Mystery 1: Overfitting is not necessarily bad.}
The performance of behavioral cloning often increases with more overfitting~\citep{mandlekar2021matters, allshire2026scalable},
and a smaller dataset can sometimes lead to better performance than a larger one~\citep{codevilla2019exploring, bai2025rethinking}.

\textbf{Mystery 2: Open-loop\footnote{Throughout this work, we use the terminology ``open-loop'' to denote \emph{partially} open-loop control with action chunking~\citep{zhao2023learning}.} is often better than closed-loop.}
Open-loop policies with action chunking can substantially outperform closed-loop ones even when there is no inference delay~\citep{zhao2023learning, lazzati2026does, zeng2026revisiting}.
Adding history conditioning to closed-loop policies does not necessarily fix this issue~\citep{de2019causal, wen2020fighting, mark2026bpp}.

\textbf{Mystery 3: Policies should be very large.}
Behavioral cloning on human-collected data often requires a much larger network
than behavioral cloning on synthetic data, even on state-based tasks~\citep{pan2026much}.

\textbf{Mystery 4: Features matter even with enough data.}
Behavioral cloning performance often depends heavily on the input features~\citep{de2019causal, mandlekar2021matters, lu2026would}.
Scaling data does not necessarily reduce this dependence.

\begin{figure}[t!]
    \centering
    \includegraphics[width=1.0\textwidth]{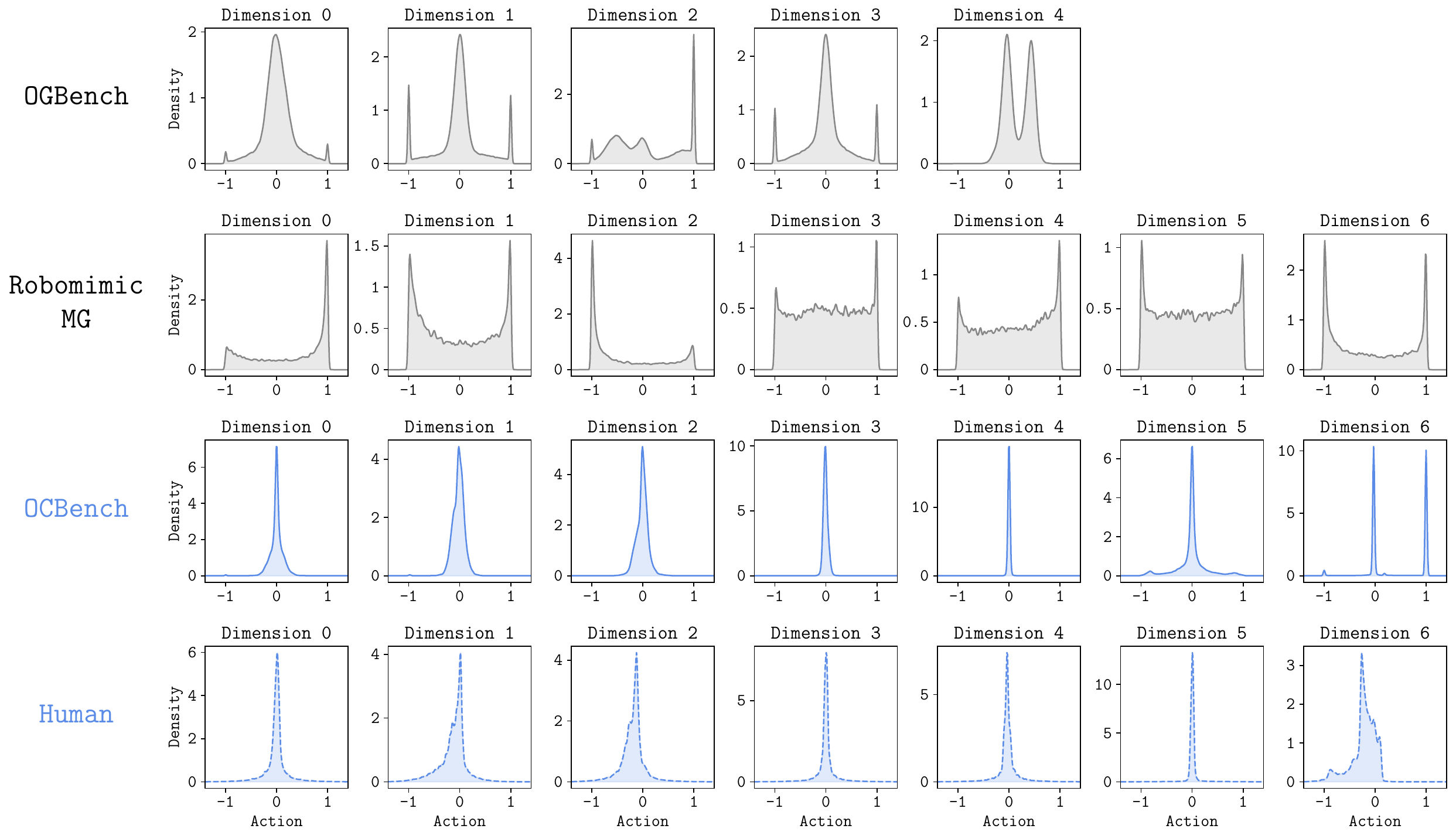}
    \caption{
    \footnotesize
    OCBench's scripted policy has more human-like action statistics
    than previous scripted (or RL-trained) policies,
    which tend to saturate the action range (\ie, $-1$ or $+1$) or have overly uniform distributions.
    }
    \label{fig:action_hist}
\end{figure}

The phenomena listed above have mostly been observed with actual human demonstration data.
Unfortunately, scientifically studying them is challenging in practice.
In the real world, we typically have limited control over such human-generated data.
Existing simulated BC and RL benchmarks based on scripted or RL-trained policies provide more control over data collection,
but they do not consistently reproduce all these phenomena.
For example, in many popular benchmarks based on systematic data generation
(\eg, OGBench manipulation, Robomimic MG, etc.),
the gap between closed-loop and open-loop policy is often not as significant as in real data,
and they do not require as large policies (see \Cref{sec:prev_bench}).

We hypothesize that this discrepancy mainly stems from the difference in the \emph{structural properties}
of scripted (or RL-trained) policies and human demonstrations.
For example, previous scripted policies are often
not as temporally correlated, not as multi-modal,
and mostly rely on Gaussian noise to diversify trajectories~\citep{fu2020d4rl, park2025ogbench}.
Moreover, their action statistics are typically biased toward boundary actions ($-1$ or $1$) or excessively uniform, unlike human demonstrations,
as scripted or RL-trained policies often saturate the action range or are often randomized via Gaussian noise (\Cref{fig:action_hist}).
OCBench aims to bridge the gap between scripted policies and human demonstrations.

\cutsectionup
\section{OCBench: A controllable robotic manipulation benchmark}
\label{sec:ocbench}
\cutsectiondown

The main goal of OCBench is to provide a diverse set of tasks and datasets generated by controllable scripted policies
that have similar \emph{structural properties} to human demonstrations,
so that researchers can scientifically study diverse BC- (and potentially RL-) related phenomena in this benchmark.
Moreover, we aim to implement \emph{both} the tasks and data-generating scripts of OCBench in GPU-accelerated MuJoCo (MJWarp)
to enable highly efficient data generation and simulation.

In this section, we first describe our approach to mimic the properties of human demonstrations (\Cref{sec:datasets})
and the tasks provided in OCBench (\Cref{sec:tasks}).

\cutsubsectionup
\subsection{Scripted policies and datasets}
\label{sec:datasets}
\cutsubsectiondown

One of the core contributions of OCBench is its scripted policies
designed to mimic the properties of human demonstrations.
We hypothesize that human demonstrations have the following key properties:
they are (1) very narrowly distributed (as opposed to Gaussian-noise-augmented trajectories),
(2) highly non-Markovian and temporally correlated,
and (3) semantically diverse (\eg, grabbing objects with diverse poses and strategies).

To mimic these core properties of human demonstrations,
we rely on randomized piecewise $C^1$-continuous Hermite splines as our data-generation backbone.
Given a task, we first sample random control points for semantic phases (\eg, grasp, lift, transport, etc.),
and smoothly interpolate them with Hermite splines with Catmull-Rom tangent matching to ensure $C^1$ continuity.
Importantly, we randomize a \emph{wide} range of factors,
including waypoints, grasping angles and directions, contact points, motion speeds,
gripper rolls and pitches, trajectory curvatures, and high-level task-solving strategies,
to enhance diversity as much as possible.
We also let the scripted policy occasionally make mistakes and recover from them.

As we show in \Cref{sec:mysteries},
we find that scripted policies designed in this way allow us to reproduce many counterintuitive phenomena happening with human-data behavioral cloning in OCBench,
unlike previous scripted benchmarks (\Cref{sec:prev_bench}).
\Cref{fig:action_hist} also shows that the OCBench data has action statistics more similar to actual human demonstrations than previous scripted or RL-trained policies.

\begin{figure}[t!]
    \centering
    \includegraphics[width=1.0\textwidth]{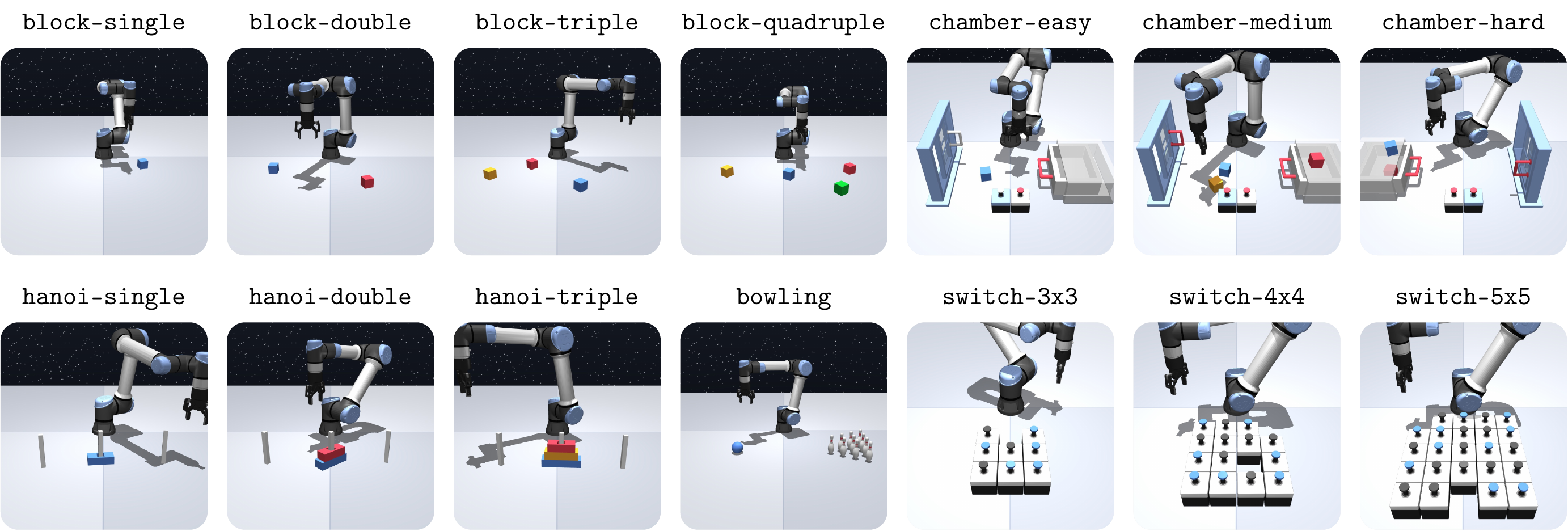}
    \caption{
    \footnotesize
    OCBench provides $28$ tasks across five types of environments.
    }
    \label{fig:envs}
\end{figure}

\cutsubsectionup
\subsection{Tasks}
\label{sec:tasks}
\cutsubsectiondown

OCBench provides $28$ tasks across five types of environments (\tt{block}, \tt{chamber}, \tt{switch}, \tt{hanoi}, and \tt{bowling})
with diverse manipulation challenges.
We provide an illustration in \Cref{fig:envs} and their detailed descriptions in \Cref{sec:appx_tasks}.

All tasks and environments are implemented in MuJoCo~\citep{todorov2012mujoco}.
We provide both CPU and GPU-accelerated (MJWarp) versions of the tasks and their accompanying scripted policies.
The CPU and MJWarp versions have \emph{logically} identical dynamics and data generation strategies,
although they are not bit-identical due to differences in numerical solvers and implementation details.

OCBench environments operate at $50$Hz and have a $7$-dimensional joint delta (relative) action space.
We mainly consider state-based environments in this work, but we provide multi-view pixel-based observations as well.
For convenience, we also provide \tt{lite} versions of the environments (except \tt{bowling}),
which operate at $10$Hz and have demonstrations that run twice as fast,
to enable lightweight experimental iterations with smaller networks.

\cutsectionup
\section{Studying behavioral cloning in OCBench}
\label{sec:mysteries}
\cutsectiondown

Now, we study the four mysteries mentioned in \Cref{sec:motivation},
based on our reproductions in OCBench.
Here, our goal is \textbf{not} to completely demystify these phenomena, which we leave for future work on OCBench.
Rather, we aim to demonstrate examples of
how OCBench's controlled, GPU-accelerated scripts enable scientific studies of them.

\textbf{Behavioral cloning methodology.}
We mainly consider flow behavioral cloning~\citep{black2024pi_0} with action chunking~\citep{zhao2023learning} in this work,
as it is a standard objective in current large-scale robotic systems~\citep{bjorck2025gr00t, intelligence2026pi}.
Given a dataset $\gD = \{\tau^{(i)}\}_i$ consisting of state-action trajectories $\tau = (s_0, a_0, s_1, \ldots, s_T)$,
flow behavioral cloning trains an action-chunking policy $\pi(\pl{a_{t:t+k-1}} \mid \pl{s_t}): \gS \to \Delta(\gA^k)$ with flow matching~\citep{lipman2022flow, albergo2022building, liu2022flow},
where $\gS$ is the state space, $\gA$ is the action space, $\Delta(\gX)$ denotes the set of probability distributions over a space $\gX$,
and $a_{t:t+k-1} = (a_t, a_{t+1}, \ldots, a_{t+k-1})$ denotes the length-$k$ action chunk.
At test time, we by default execute a full action chunk $a_{t:t+k-1}$ sampled from the policy in an open-loop manner, without re-planning.
For our experiments, we average over two random seeds and denote $95$\% confidence intervals in the plots, unless mentioned otherwise.

\cutsubsectionup
\subsection{Mystery 1: Overfitting is not necessarily bad}
\label{sec:mystery1}
\cutsubsectiondown

The first counterintuitive phenomenon we study is that overfitting is not always harmful in behavioral cloning.
In fact, it turns out that overfitting is often \emph{beneficial}.
This phenomenon has been anecdotally reported in previous works in real-world robotics and next-token prediction~\citep{ouyang2022training, allshire2026scalable}.

\begin{figure}[h!]
    \centering
    \begin{subfigure}[t]{0.49\textwidth}
        \centering
        \includegraphics[width=1.0\textwidth]{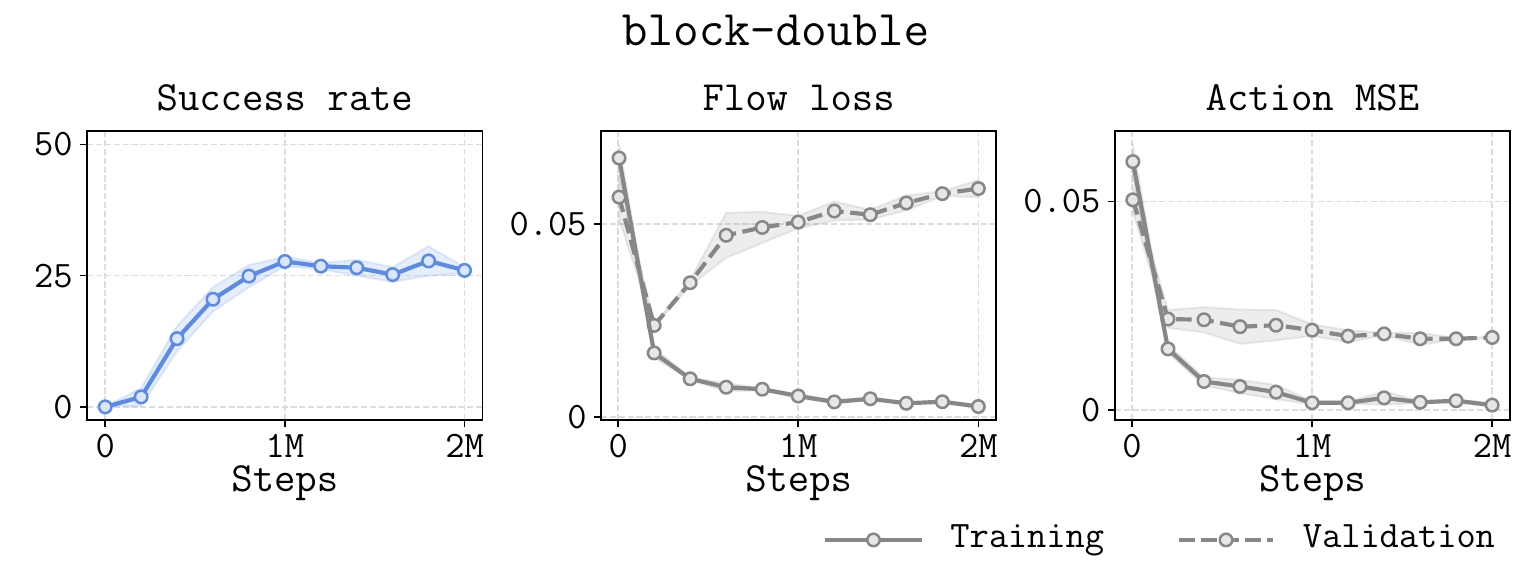}
    \end{subfigure}%
    \hfill%
    \begin{subfigure}[t]{0.49\textwidth}
        \centering
        \includegraphics[width=1.0\textwidth]{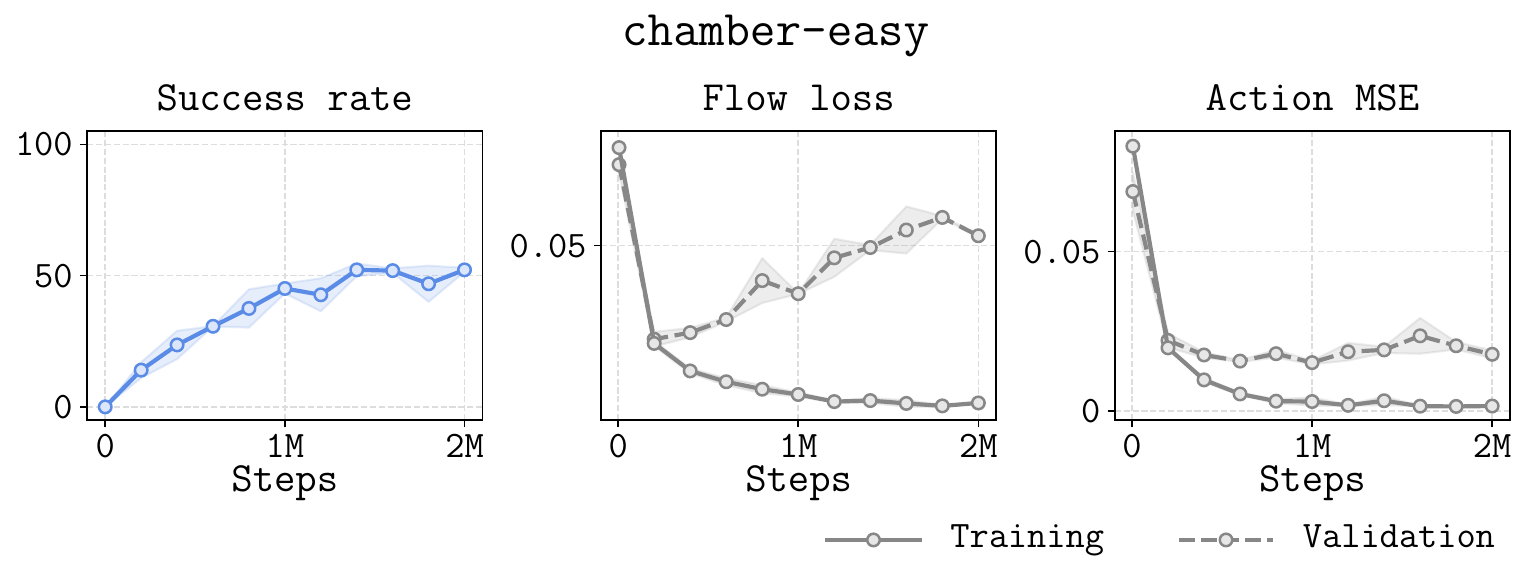}
    \end{subfigure}
    \caption{
    \footnotesize
    Performance can increase as the agent overfits more.
    }
    \label{fig:overfitting}
\end{figure}

\Cref{fig:overfitting} shows the performance and loss curves of flow BC agents trained
on \tt{block-double-task1} with $30$K episodes and \tt{chamber-easy-task2} with $10$K episodes.
The result shows that the performance generally does not drop over time,
even though the validation flow loss keeps growing.
In particular, the performance when the agent starts to overfit (\ie, the step with the minimum validation loss)
is substantially lower than the final performance, on both tasks.

\begin{figure}[h!]
    \centering
    \begin{subfigure}[t]{0.49\textwidth}
        \centering
        \includegraphics[width=1.0\textwidth]{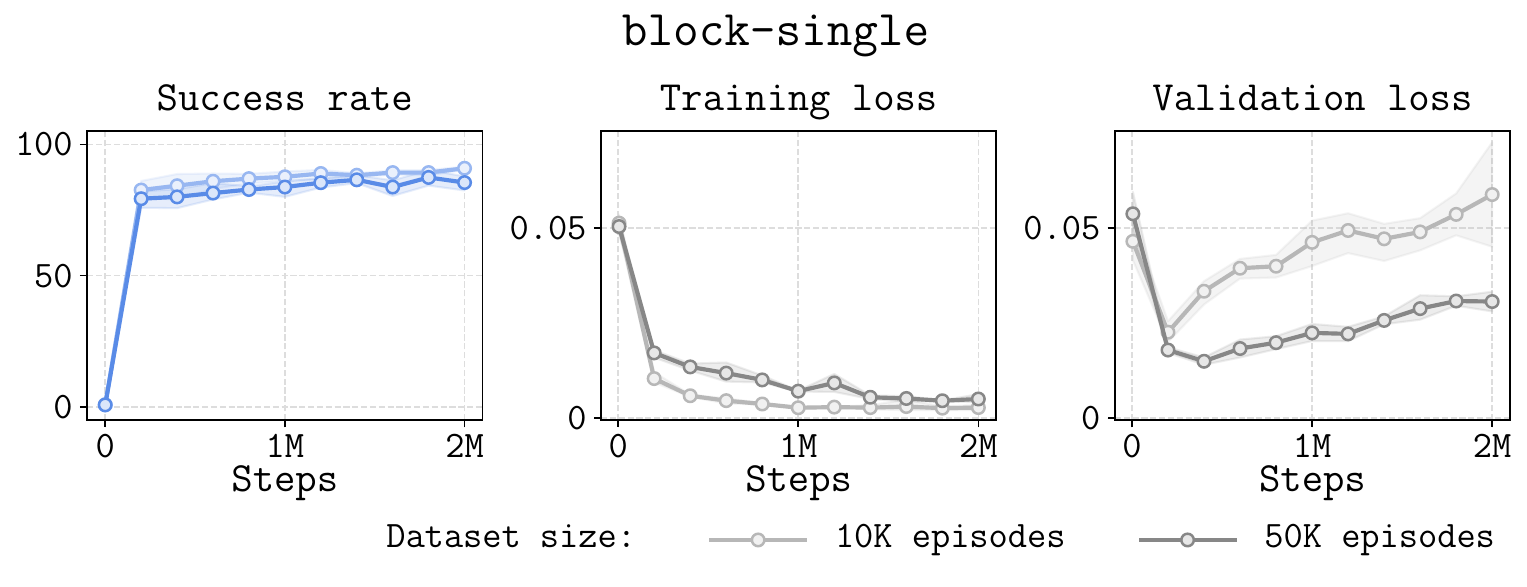}
    \end{subfigure}%
    \hfill%
    \begin{subfigure}[t]{0.49\textwidth}
        \centering
        \includegraphics[width=1.0\textwidth]{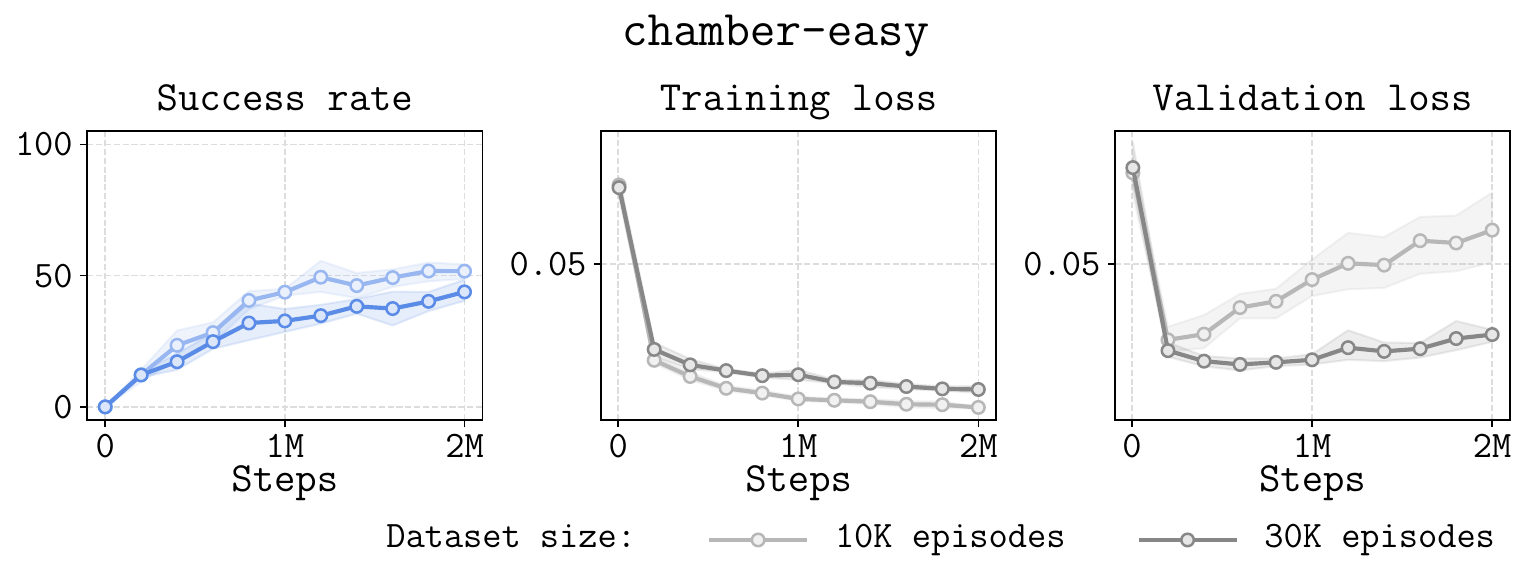}
    \end{subfigure}
    \caption{
    \footnotesize
    Using a larger dataset can \emph{hurt} performance, even when sampled from the same data distribution.
    }
    \label{fig:dataset_size}
\end{figure}

\Cref{fig:dataset_size} shows an even more counterintuitive phenomenon.
Here, we compare the performance of flow BC trained on two datasets that differ \emph{only} in sizes
($10$K and $50$K for \tt{block-single-task1} and $10$K and $30$K for \tt{chamber-easy-task2}).
The results show that \textbf{using a larger dataset can hurt performance} by a small yet statistically significant margin,
even when the datasets are sampled from the \emph{same} distribution.
In this plot, the curves are averaged over $4$ independently sampled datasets for each setting.
This indicates that overfitting might potentially be \emph{desirable}, at least in some cases.

This size-reversal phenomenon was anecdotally reported in BC in autonomous driving~\citep{codevilla2019exploring} and robotics~\citep{bai2025rethinking}.
Although the authors of these works~\citep{codevilla2019exploring, bai2025rethinking} suspect that it may be due to non-identical data distributions and biases,
our result shows that this is a \emph{reproducible} phenomenon that happens even in a fully controlled setting with independent and identically distributed samples.
This is an example of how OCBench enables a scientific study of BC-related phenomena.

While we do not have a definitive answer to this overfitting phenomenon,
especially to the size-reversal one (which we leave for future studies on OCBench),
we discuss three potential hypotheses and some refutations below.

\textbf{Hypothesis 1: Flow loss alone does not capture performance.}
One hypothesis is that flow matching's velocity loss is not necessarily correlated with performance.
For example, \citet{allshire2026scalable} recently reported that action mean squared error (MSE)
(\ie, the squared distance between dataset actions and generated actions)
correlates more strongly with performance than flow loss.

The third and sixth plots in \Cref{fig:overfitting} show the MSE metric.
While this metric indeed correlates more strongly with performance than flow loss,
it still does not perfectly explain performance
(\eg, the performance keeps increasing beyond $200$K steps while MSE flatlines).
Moreover, it does not fully explain the size-reversal phenomenon (\Cref{fig:dataset_size_mse}),
so we believe this hypothesis alone cannot fully account for our observations.

\begin{wrapfigure}{r}{0.3388\textwidth}
    \centering
    \vspace{-5pt}
    \raisebox{0pt}[\dimexpr\height-1.0\baselineskip\relax]{
        \begin{subfigure}[t]{1.0\linewidth}
        \includegraphics[width=\linewidth]{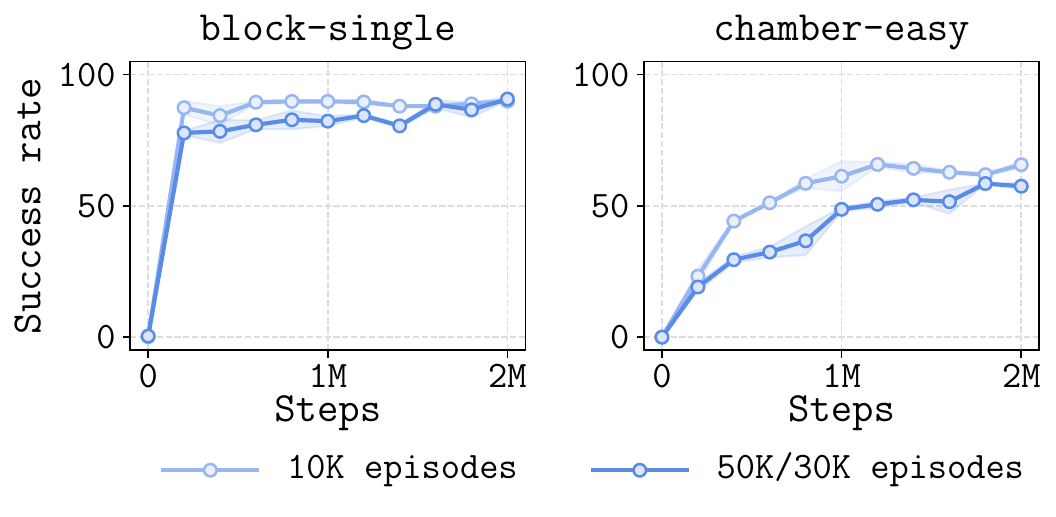}
        \end{subfigure}
    }
    \caption{
    \footnotesize
    \Cref{fig:dataset_size} with larger models.
    }
    \vspace{-10pt}
    \label{fig:dataset_size_capacity}
\end{wrapfigure}
\textbf{Hypothesis 2: It is a network capacity issue.}
Another hypothesis is that the size-reversal issue is due to a network capacity issue (\ie, the model size is the confounder).
To examine this possibility, we train $4\times$ larger, $540$M-sized models in the same setting ($2$ seeds each).
\Cref{fig:dataset_size_capacity} shows the results.
Even with half-billion-sized models, the size-reversal phenomenon does not vanish.
This suggests that limited model capacity alone is insufficient to explain this phenomenon.

\textbf{Hypothesis 3: Overfitting prevents test-time distribution shift.}
The third hypothesis is that it is mainly related to the test-time distribution shift issue (``DAgger issue''~\citep{ross2011reduction}) in behavioral cloning.
By overfitting to the dataset, the agent can more closely follow the ``nearest-neighbor'' segment in the dataset.
Although the task is randomized in each episode, following the nearest-neighbor motion often provides reasonable behavior~\citep{pari2021surprising},
which could save the agent from compounding errors and distribution shift.
As long as this neural nearest-neighbor retrieval is based on \emph{learned}, generalizable representations,
the agent could still perform well at test time despite overfitting.
While we leave quantitative investigations about this point for future work,
\Cref{fig:test_time_generalization} in \Cref{sec:mystery4} indirectly supports this hypothesis.

\cutsubsectionup
\subsection{Mystery 2: Open-loop is often better than closed-loop}
\label{sec:mystery2}
\cutsubsectiondown

The second phenomenon is that partially open-loop control (\ie, action chunking) is often \emph{crucial} in behavioral cloning,
even when there is no inference delay.
In the literature,
\citet{zhao2023learning} show that a closed-loop policy completely fails while action chunking can achieve reasonable performance.
More recent works~\citep{zhang2026action, lazzati2026does, zeng2026revisiting} report similar phenomena and discuss them in more diverse settings.
We reproduce this on two OCBench tasks,
\tt{block-single-task1} and \tt{hanoi-single-task1}, with streaming data.

\begin{wrapfigure}{r}{0.3388\textwidth}
    \centering
    \vspace{-2pt}
    \raisebox{0pt}[\dimexpr\height-1.0\baselineskip\relax]{
        \begin{subfigure}[t]{1.0\linewidth}
        \includegraphics[width=\linewidth]{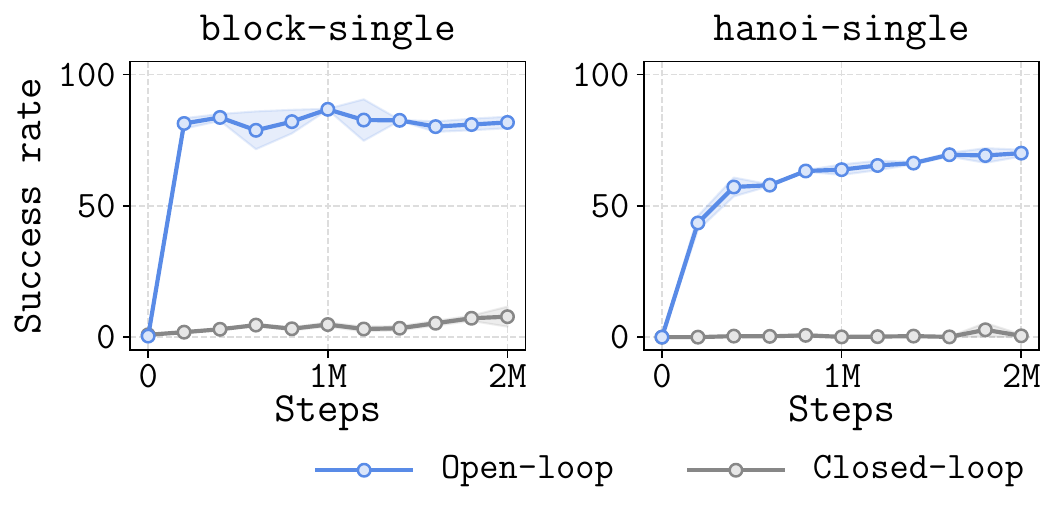}
        \end{subfigure}
    }
    \vspace{-5pt}
    \caption{
    \footnotesize
    Closed-loop control fails.
    }
    \label{fig:open_closed}
    \vspace{-14pt}
\end{wrapfigure}
\Cref{fig:open_closed} compares the performance of
closed-loop policies $\pi(\pl{a_t} \mid \pl{s_t})$ and open-loop (action-chunking) policies $\pi(\pl{a_{t:t+24}} \mid \pl{s_t})$.
The results suggest that open-loop control is indeed essential on these tasks,
whereas closed-loop policies almost completely fail.
Note that the policies are trained with \emph{streaming data} in this experiment,
meaning that the failure of closed-loop policies is not due to data scarcity or overfitting.

There are two main hypotheses for why open-loop control is essential:
one is about compounding errors and the other is about non-Markovianity.

\textbf{Hypothesis 1: Closed-loop control compounds errors.}
One hypothesis is that closed-loop policies sample (stochastic) actions more often,
so errors can accumulate faster than under open-loop control.
On the other hand, the policy becomes more complex as the action chunk length increases,
so there is a trade-off.
The results suggest that the sweet spot might be around length-$25$ action chunks.

While plausible, this hypothesis is challenged by a recent work~\citep{lazzati2026does},
which shows that a closed-loop, ``delayed'' policy $\pi(\pl{a_t} \mid \pl{s_{t-k}})$
can match an open-loop policy's performance on human-collected data.
Such a policy still samples an action at every step and thus has the compounding error problem,
but can capture temporal dependencies by conditioning on a past state.

\begin{wrapfigure}{r}{0.3317\textwidth}
    \centering
    \vspace{-5pt}
    \raisebox{0pt}[\dimexpr\height-1.0\baselineskip\relax]{
        \begin{subfigure}[t]{1.0\linewidth}
        \includegraphics[width=\linewidth]{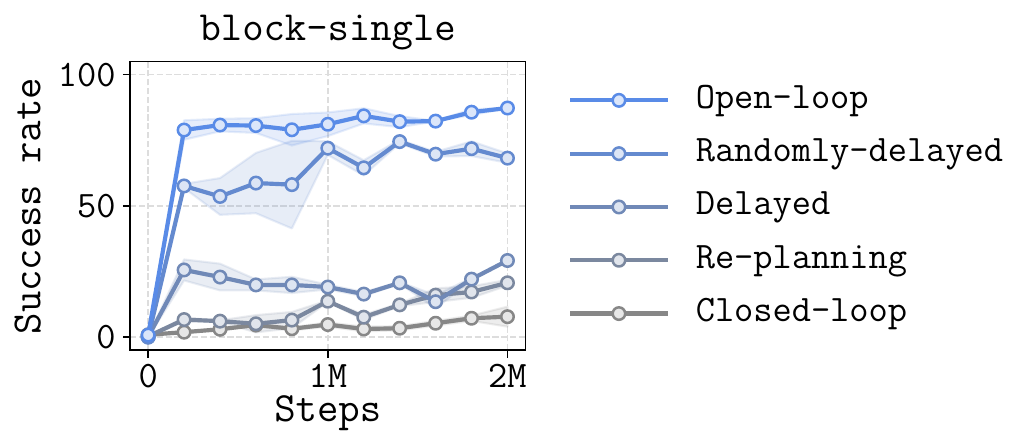}
        \end{subfigure}
    }
    \vspace{-13pt}
    \caption{
    \footnotesize
    Diverse control strategies.
    }
    \vspace{-10pt}
    \label{fig:control_strategies}
\end{wrapfigure}

We test whether this finding also holds in OCBench.
\Cref{fig:control_strategies} compares the performance of the following five different control strategies on \tt{block-single-task1}:
\begin{itemize}
    \item \tt{Closed-loop}, a standard closed-loop policy $\pi(\pl{a_t} \mid \pl{s_t})$.
    \item \tt{Open-loop}, a standard open-loop (action-chunking) policy $\pi(\pl{a_{t:t+24}} \mid \pl{s_t})$.
    \item \tt{Re-planning}, a closed-loop policy derived from an open-loop policy $\pi(\pl{a_{t:t+24}} \mid \pl{s_t})$ by executing only the first action and re-planning at every step.
    \item \tt{Delayed}, a delayed policy $\pi(\pl{a_t} \mid \pl{s_{t-24}})$.
    \item \tt{Randomly-delayed}, a randomly delayed policy $\pi(\pl{a_t} \mid \pl{s_{t-k}})$, where $k$ is sampled between $0$ and $9$ (inclusive).
\end{itemize}

The results show that a randomly delayed closed-loop policy can achieve strong performance
(though it still falls behind open-loop control),
suggesting that compounding errors alone may not fully explain the gap between closed-loop and open-loop control.
This also shows that OCBench's synthetic demo-like trajectories can reproduce
a similar result to that observed with real human-demonstration data~\citep{lazzati2026does}.

\textbf{Hypothesis 2: Closed-loop policies do not capture non-Markovianity in data.}
Another hypothesis for the necessity of open-loop control
is that closed-loop policies fail to capture non-Markovianity in temporally correlated demonstrations.
This expressivity mismatch can lead to distribution shift at test time,
which in turn leads to poor performance.

One way to test this hypothesis is to compare closed- and open-loop control
on ``Markovianized'' data,
where the scripted policy randomly generates a new plan at every step,
breaking the temporal dependencies~\citep{zeng2026revisiting}.

\clearpage

\begin{wrapfigure}{r}{0.3388\textwidth}
    \centering
    \vspace{5pt}
    \raisebox{0pt}[\dimexpr\height-1.0\baselineskip\relax]{
        \begin{subfigure}[t]{1.0\linewidth}
        \includegraphics[width=\linewidth]{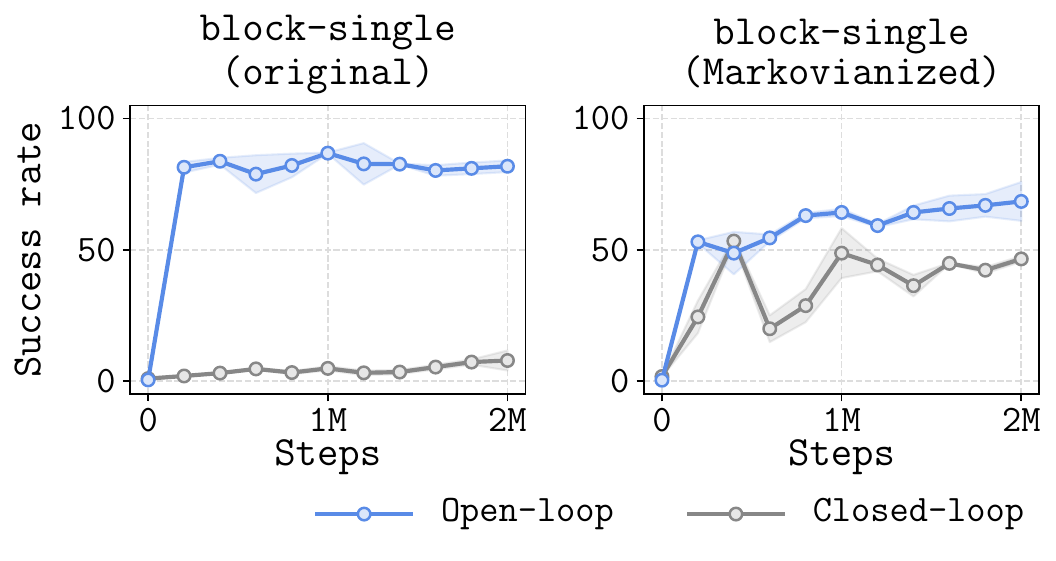}
        \end{subfigure}
    }
    \vspace{-15pt}
    \caption{
    \footnotesize
    Markovianized datasets.
    }
    \vspace{-12pt}
    \label{fig:markovianized}
\end{wrapfigure}
\Cref{fig:markovianized} shows the training curves of closed- and open-loop policies
trained on the original and Markovianized datasets.
The results suggest that Markovianizing the data can indeed reduce the gap between closed- and open-loop control.
However, there still exists a gap on Markovianized data,
suggesting that both compounding errors \emph{and} non-Markovianity might contribute to the performance difference.

\cutsubsectionup
\subsubsection{Aside: What about history conditioning?}
\cutsubsectiondown

A different way to handle non-Markovianity with a closed-loop policy is
to condition on \emph{history}, \mbox{$\pi(\pl{a_t} \mid \pl{s_{t-24:t}})$}.
This allows the policy to capture a similar degree of non-Markovianity
to that of the corresponding open-loop policy $\pi(\pl{a_{t:t+24}} \mid \pl{s_t})$.

\begin{figure}[h!]
    \centering
    \begin{subfigure}[t]{0.4934\textwidth}
        \centering
        \includegraphics[width=1.0\textwidth]{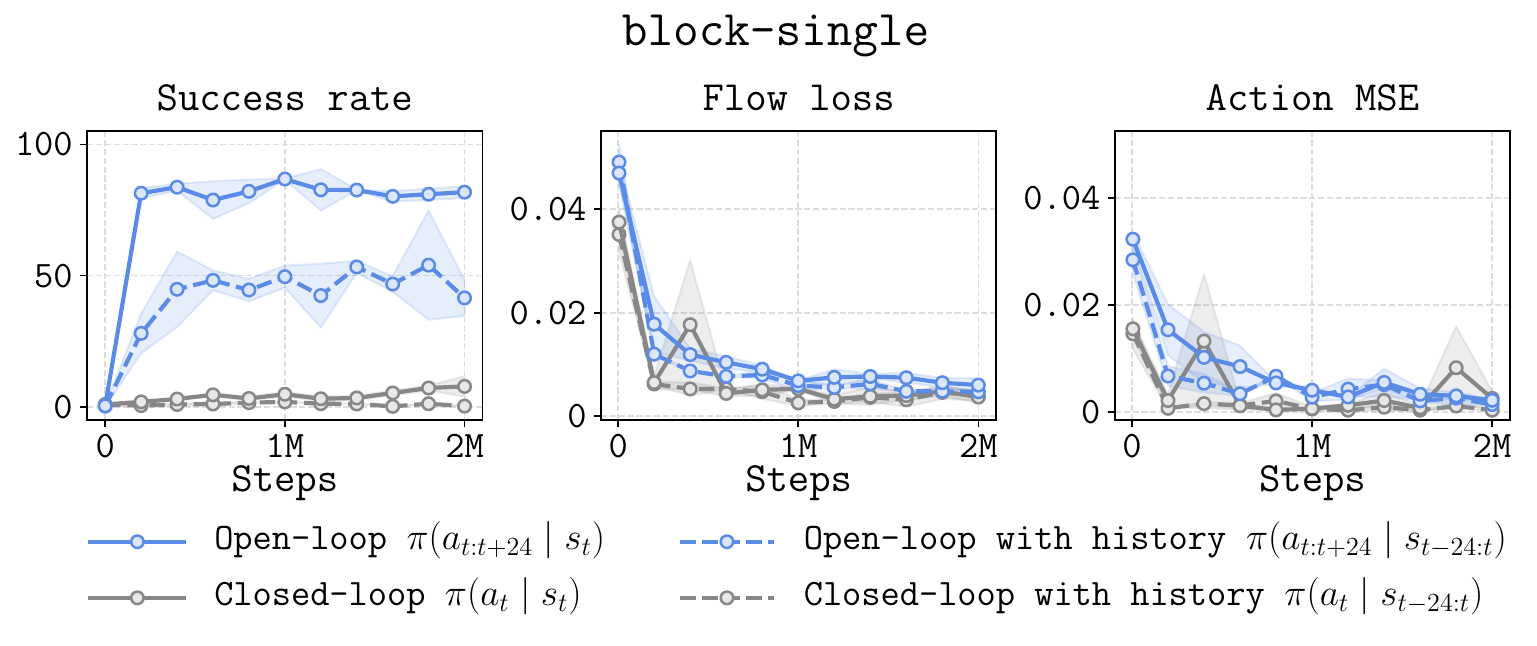}
    \end{subfigure}%
    \hfill%
    \begin{subfigure}[t]{0.4934\textwidth}
        \centering
        \includegraphics[width=1.0\textwidth]{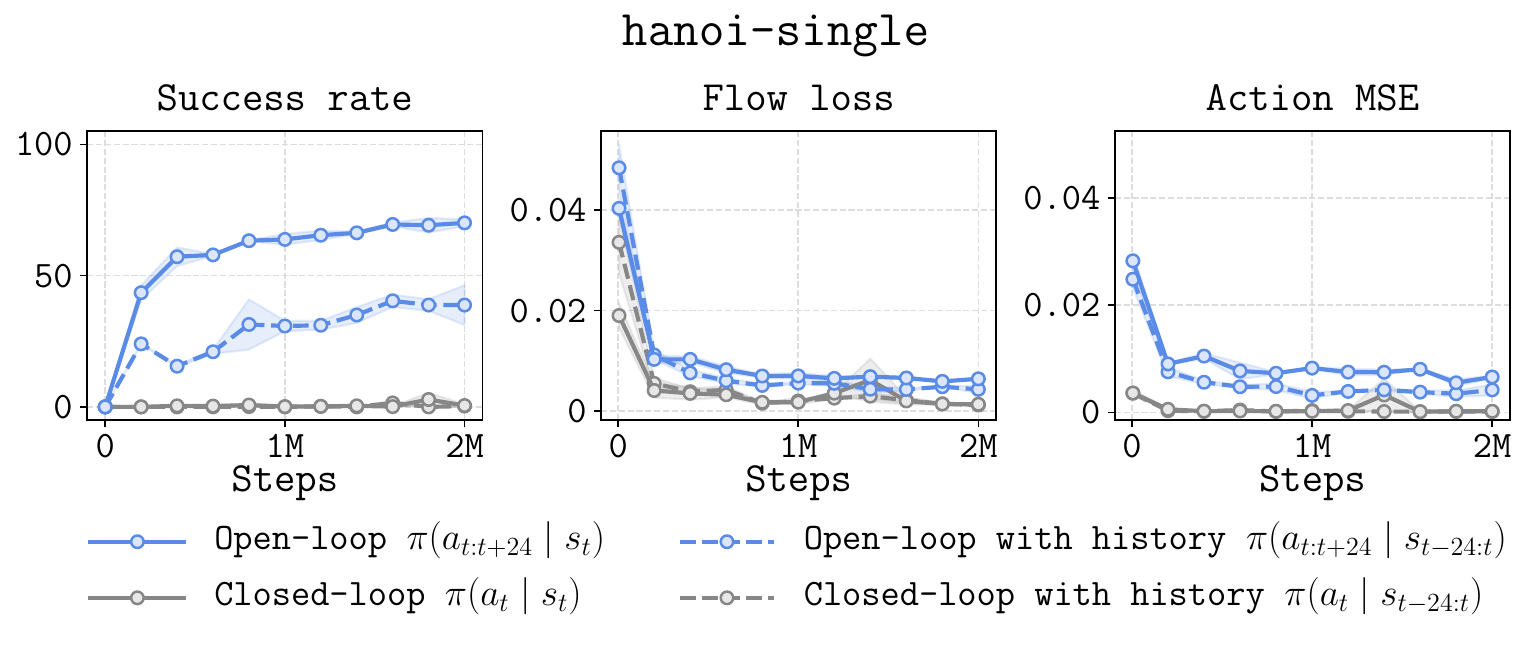}
    \end{subfigure}
    \caption{
    \footnotesize
    History conditioning can hurt performance despite the lower loss.
    }
    \label{fig:history}
\end{figure}

\Cref{fig:history} compares the performance of closed- and open-loop policies with and without history conditioning,
under the streaming-data setting.
Perhaps a bit unexpectedly, the results show that history conditioning can \emph{hurt} in practice.
Note that this is not due to overfitting, as we use streaming data in this experiment.
In fact, history-conditioned policies achieve lower flow loss and action MSE,
meaning that they \emph{do} fit the dataset better.
However, this does not lead to better performance at test time.

History conditioning hurting performance is a previously reported phenomenon in robotics~\citep{de2019causal, mandlekar2021matters, mark2026bpp}.
There are two possible explanations.
One is causal confusion~\citep{de2019causal}:
a history-conditioned policy might learn to ``cheat'' by simply outputting the previous action,
which is often highly (yet spuriously) correlated with the next action.
Another explanation is that history-conditioned policies are more susceptible to test-time distribution shift in general,
as they simply have a larger input space.
While we do not have a conclusive answer yet,
we note that this issue happens even in this basic task with streaming data.
We expect that our minimal reproduction will make it easier to study history conditioning in future work.

\cutsubsectionup
\subsection{Mystery 3: Policies should be (very) large}
\label{sec:mystery3}
\cutsubsectiondown

The next phenomenon we study is that we generally need \emph{much} larger networks to clone human behaviors
than to clone synthetic trajectories generated by (previous) scripted policies.
A typical network used to handle synthetic data in D4RL~\citep{fu2020d4rl} and OGBench~\citep{park2025ogbench}
is a $[512, 512, 512]$-sized
multi-layer perceptron (MLP)~\citep{park2025ogbench},
which has roughly $0.5$M trainable parameters.
In such tasks, we typically see limited performance improvement with larger models (\Cref{fig:prev_model_size}).
On the other hand, cloning \emph{human} behaviors often benefits from much larger networks (\eg, up to $100$M-sized models),
especially when using flow matching or diffusion,
even in \emph{state-based} environments of similar dimensionality and difficulty to OGBench tasks~\citep{pan2026much}.
We test whether OCBench is similar to previous synthetic data or human-generated data in terms of the best model size.

\begin{figure}[h!]
    \centering
    \begin{subfigure}[t]{0.5\textwidth}
        \centering
        \includegraphics[width=1.0\textwidth]{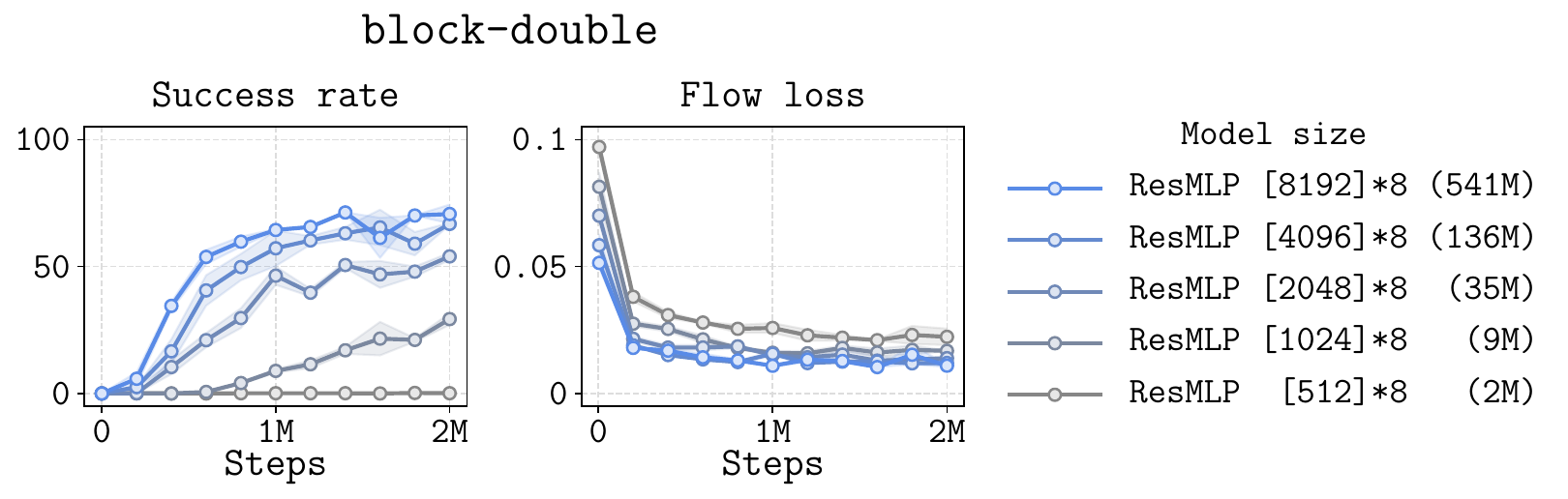}
    \end{subfigure}%
    \hfill%
    \begin{subfigure}[t]{0.5\textwidth}
        \centering
        \includegraphics[width=1.0\textwidth]{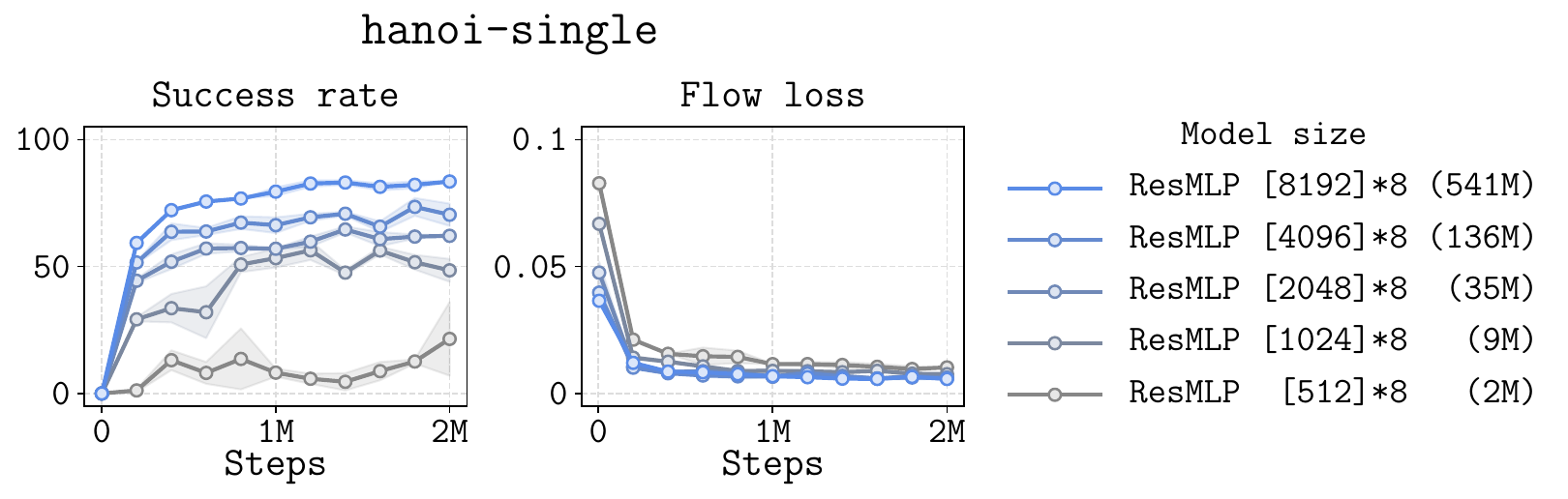}
    \end{subfigure}
    \caption{
    \footnotesize
    Behavioral cloning from human-like data can require $500$M-sized models even on state-based tasks.
    }
    \label{fig:model_size}
\end{figure}

\Cref{fig:model_size} presents a model size ablation on two OCBench tasks.
The results show that OCBench tasks require much larger networks than previous synthetic benchmarks,
as is typically the case with behavioral cloning from human demonstrations~\citep{pan2026much}.
\Cref{fig:model_size} shows performance keeps improving up to $[8192] \times 8$-sized residual MLPs,
which have about $0.5$ \emph{billion} parameters,
even on these basic state-based manipulation tasks.

We discuss two hypotheses for this phenomenon.

\textbf{Hypothesis 1: Human-data behavioral cloning is simply hard.}
One hypothesis is that BC from human (or human-like) data is just difficult.
Since human behaviors are highly complex and diverse,
BC may naturally require a large model to fit the dataset,
especially given that overfitting is often beneficial (\Cref{sec:mystery1}).
This hypothesis could help explain why vision-language-action (VLA) models require a large ``action expert,''
as also reported by \citet{black2024pi_0}.
Our result suggests that we may need a billion-scale action expert even with \emph{perfect} perception and task understanding and in a single task.

\textbf{Hypothesis 2: Flow-matching requires larger networks in general.}
Another hypothesis is that flow matching (or diffusion) is slower to train and requires a larger capacity~\citep{pertsch2025fast, driess2026knowledge} compared to alternatives
(\eg, autoregressive action prediction),
possibly due to the high variance of noise coupling.

\begin{wrapfigure}{r}{0.4850\textwidth}
    \centering
    \vspace{-2pt}
    \raisebox{0pt}[\dimexpr\height-1.0\baselineskip\relax]{
        \begin{subfigure}[t]{1.0\linewidth}
        \includegraphics[width=\linewidth]{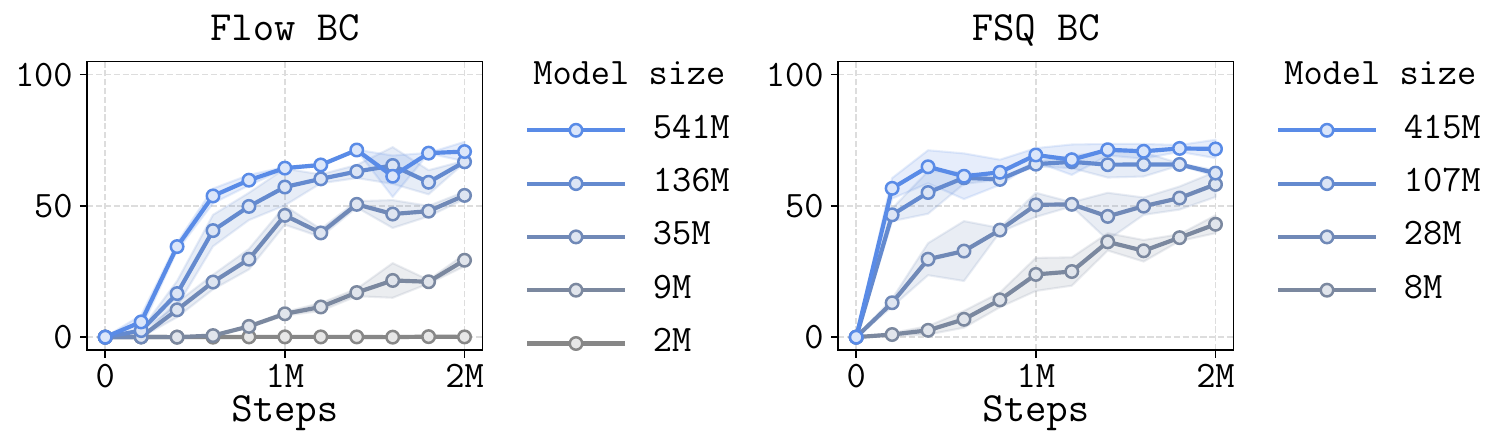}
        \end{subfigure}
    }
    \vspace{-5pt}
    \caption{
    \footnotesize
    Flow BC vs. FSQ BC.
    }
    \vspace{-5pt}
    \label{fig:flow_fsq}
\end{wrapfigure}
To test this hypothesis, we perform a similar model size ablation with FSQ~\citep{mentzer2024finite},
a discrete variant of VAE~\citep{kingma2013auto}.
Specifically, we tokenize action chunks with FSQ and train an autoregressive next-token policy over the tokenized actions.
We present the ablation results in \Cref{fig:flow_fsq}.
The results suggest that
smaller models indeed perform better with autoregressive policies than with flow-based ones.
However, performance still increases with network size even with FSQ tokens,
suggesting that the use of flow matching alone does not fully explain the need for large-capacity networks.

\cutsubsectionup
\subsection{Mystery 4: Features matter even with enough data}
\label{sec:mystery4}
\cutsubsectiondown

The last phenomenon we study is that input features substantially affect BC performance, even with virtually unlimited data.
A number of works have reported the sensitivity of behavioral cloning to input features~\citep{de2019causal, mandlekar2021matters, lu2026would}.
Since the results from these works are mostly drawn from a fixed set of human-collected trajectories,
it is less clear whether this is an issue that will go away with more data or one that is fundamental to behavioral cloning.
To understand this phenomenon, we study how different features affect performance under the streaming-data setting in OCBench.
In particular,
we consider input features that contain identical information but differ only in their relative scales.

\begin{figure}[h!]
    \centering
    \begin{subfigure}[t]{0.49\textwidth}
        \centering
        \includegraphics[width=1.0\textwidth]{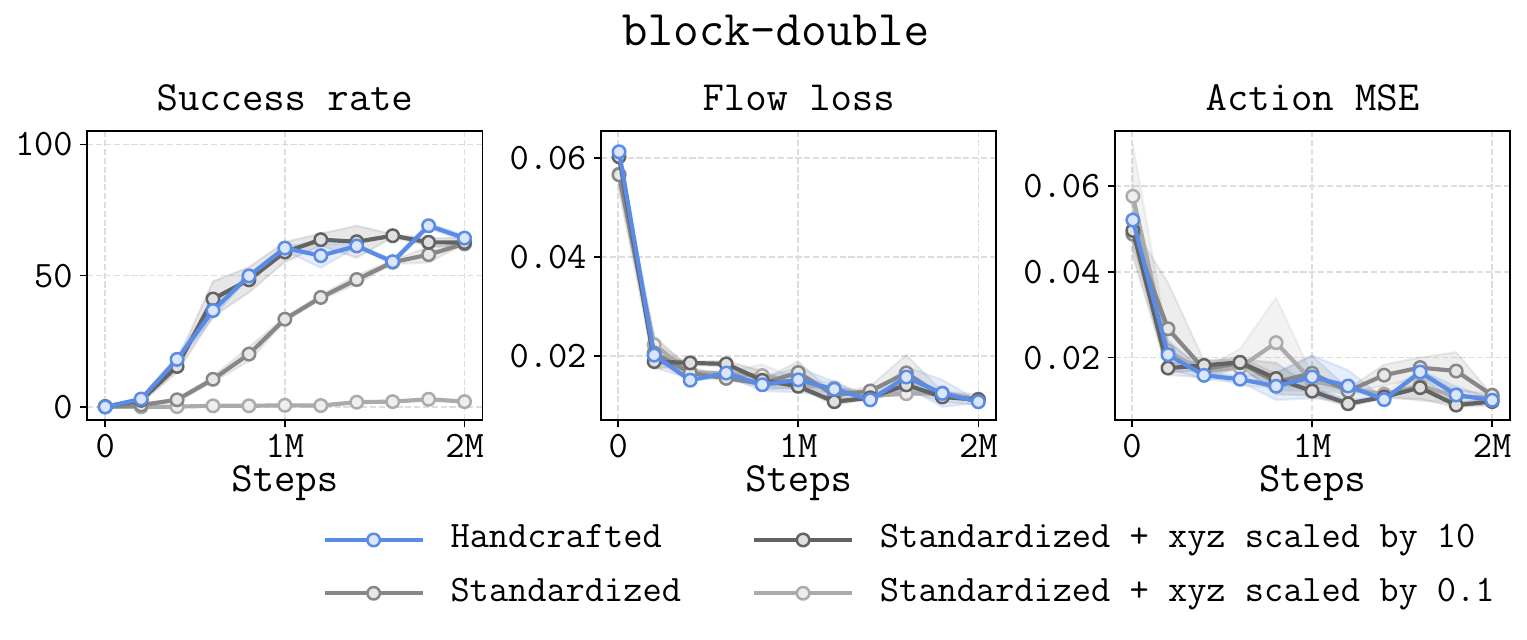}
    \end{subfigure}%
    \hfill%
    \begin{subfigure}[t]{0.49\textwidth}
        \centering
        \includegraphics[width=1.0\textwidth]{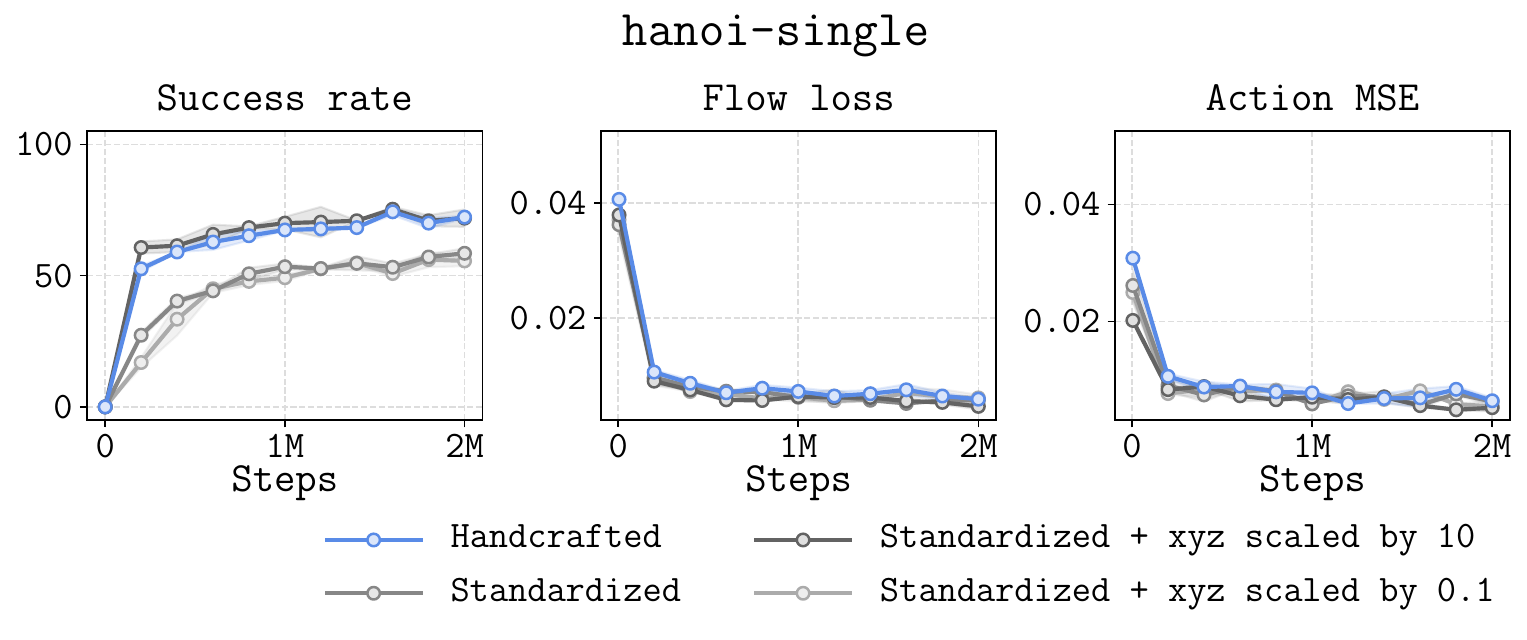}
    \end{subfigure}
    \caption{
    \footnotesize
    Different input features can lead to different performance despite almost identical training metrics.
    }
    \label{fig:input_features}
\end{figure}

\Cref{fig:input_features} shows the performance curves of the following input feature configurations
on \tt{block-double-task1} and \tt{hanoi-single-task1} with streaming data:
\begin{itemize}
    \item \tt{Handcrafted} means that we manually choose the scale of each state dimension.
    \item \tt{Standardized} means that each state dimension is standardized to the same range.
    \item \tt{Standardized + xyz scaled by \{0.1, 10\}} means that after standardizing each state dimension,
    we scale the dimensions corresponding to $xyz$ coordinates by $0.1$ or $10$.
\end{itemize}
The results suggest that performance depends \emph{heavily} on which input features are emphasized,
even with unlimited data.
Note that all four input feature configurations contain the \emph{same} information, as they only differ by scaling.
Indeed, their training and validation metrics (\eg, flow losses and MSEs) are almost indistinguishable.
However, their task performance differs substantially.

For this phenomenon, we are relatively confident in the following hypothesis.

\textbf{Hypothesis: Different features lead to different test-time generalization.}
Our hypothesis is that this is related to \emph{test-time} policy generalization~\citep{de2019causal, park2024value}.
Even when two policies have identical training metrics,
they could generalize differently to unseen states at test time.
Ultimately, BC performance is determined by how the policy behaves at states it encounters at test time.
For example, in manipulation tasks,
one can imagine that policies that focus on the object would generalize better than those that focus on internal joint angles.

\begin{wrapfigure}{r}{0.49\textwidth}
    \centering
    \raisebox{0pt}[\dimexpr\height-1.0\baselineskip\relax]{
        \begin{subfigure}[t]{1.0\linewidth}
        \includegraphics[width=\linewidth]{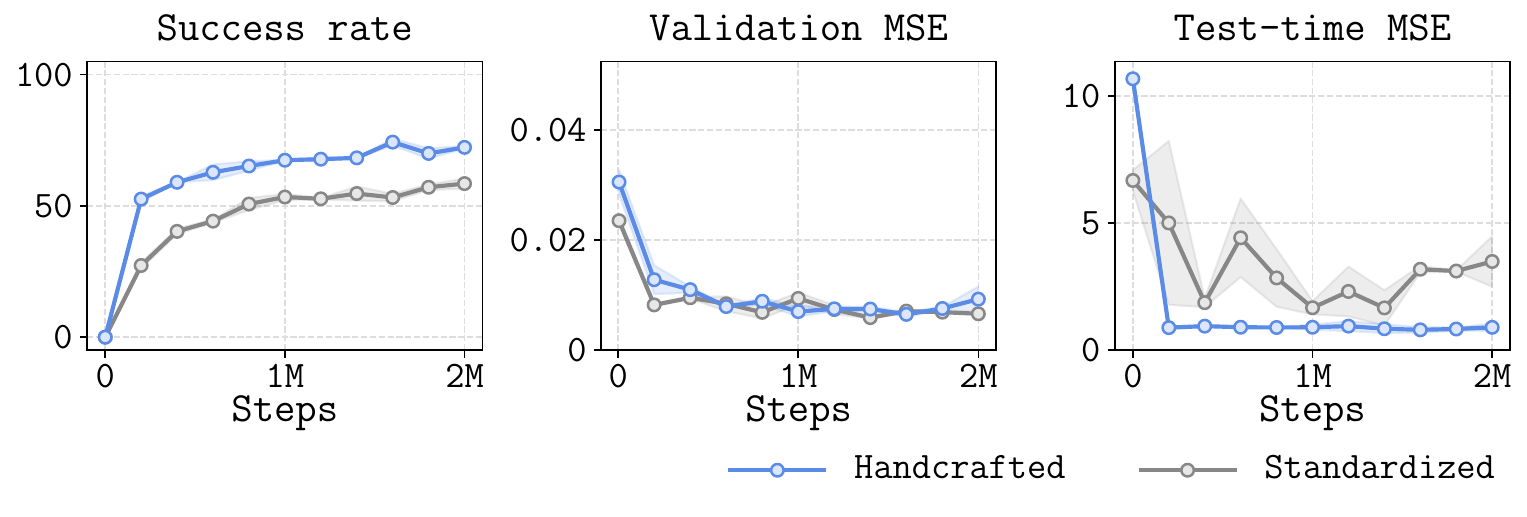}
        \end{subfigure}
    }
    \caption{
    \footnotesize
    Test-time MSE correlates better with performance than validation MSE.
    }
    \label{fig:test_time_generalization}
\end{wrapfigure}
To support this point, we measure the MSE between policy and oracle actions on \emph{test-time} states of \tt{hanoi-single-task1}.
Specifically, at each test-time state, we generate a fresh oracle plan with OCBench's scripted policy,
and measure the difference against the policy action.
We present the results in \Cref{fig:test_time_generalization}.
The results suggest that policies with different input features do behave differently at test time,
despite their almost identical training metrics.
Moreover, test-time MSE correlates better with performance than training or validation MSEs~\citep{park2024value}.
Although test-time MSE is still an imperfect metric (as it ignores multimodality),
these results nevertheless suggest that different features can lead to different \emph{generalization} behavior at test time.

\cutsectionup
\section{Closing remarks}
\cutsectiondown

In this work, we have introduced OCBench, a controllable robotic manipulation benchmark,
accompanied by GPU-accelerated scripted policies that have similar structural properties to human demonstrations.
Throughout our analyses, we have experimentally shown that OCBench reproduces diverse real-world BC-related phenomena
and enables their scientific studies via its controllable and efficient data generation scripts.

\textbf{What about reinforcement learning?}
While RL is outside the scope of this work,
OCBench's scripted policies and datasets are designed to enable future studies in RL too.
For example, OCBench's scripted policies generate trajectories at different speeds
(the fastest is around $2\times$ faster than the slowest),
just like how human robot operators have different levels of proficiency.
We expect that this makes it easier to study RL in terms of not only success rates but also throughput (\ie, task completion speed).
We also provide oracle subtask boundaries and labels to help studies in hierarchical control and RL (see \Cref{sec:appx_tasks}).
Given the current discrepancy between real-world and synthetic tasks in data-driven RL~\citep{mandlekar2021matters, huang2025using},
we hope that OCBench can help close this gap in robotics RL research too.

\textbf{Limitations.}
While we design OCBench's scripted policies to have similar properties to human demonstrations,
they are still scripted and do not perfectly mimic human behavior.
Robotics research should ultimately be done in the real world using real data.
Nonetheless, we believe OCBench complements real-world research
by serving as a fast, convenient playground
where robotics researchers can quickly test and verify their ideas across diverse tasks
and conduct controlled studies that would be difficult with real data.

\section*{Acknowledgments}
We would like to thank Kwanyoung Park for designing the camera poses for the visual tasks,
and Charles Xu, Zhiyuan (Paul) Zhou, Kwanyoung Park, and Nitish Dashora for helpful feedback on earlier versions of OCBench.
This work was partly supported by the Korea Foundation for Advanced Studies (KFAS),
NSF IIS-2150826, and ONR N00014-25-1-2060.
This research used the Savio computational cluster resource provided by the Berkeley Research Computing program at UC Berkeley.

\bibliography{iclr2027_conference}
\bibliographystyle{iclr2027_conference}

\clearpage

\appendix

\clearpage

\section{Tasks and datasets}
\label{sec:appx_tasks}

OCBench provides $28$ tasks across five types of environments (\tt{block}, \tt{chamber}, \tt{switch}, \tt{hanoi}, and \tt{bowling}).
We provide basic specifications for the tasks in \Cref{table:task_specs,table:visual_task_specs}
and the datasets in \Cref{table:dataset_specs}.

\begin{figure}[h!]
    \centering
    
    \begin{subfigure}[t]{0.30\textwidth}
        \centering
        \includegraphics[width=1.0\textwidth]{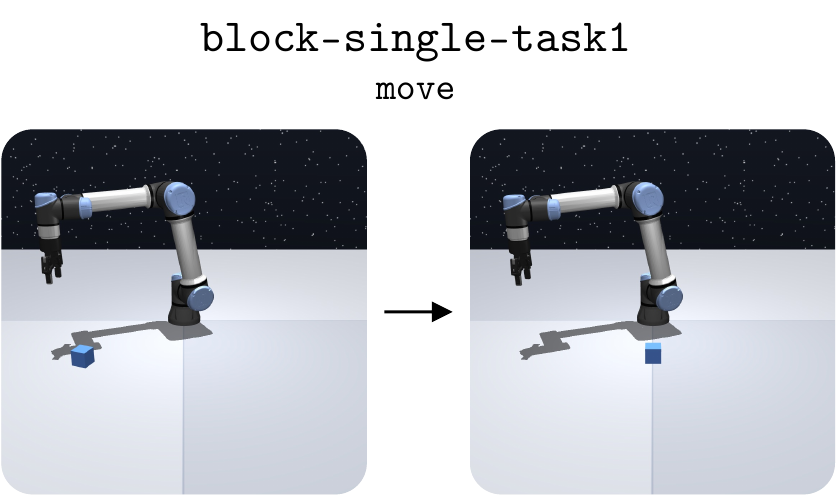}
    \end{subfigure}%

    \vspace{5pt}

    \begin{subfigure}[t]{0.30\textwidth}
        \centering
        \includegraphics[width=1.0\textwidth]{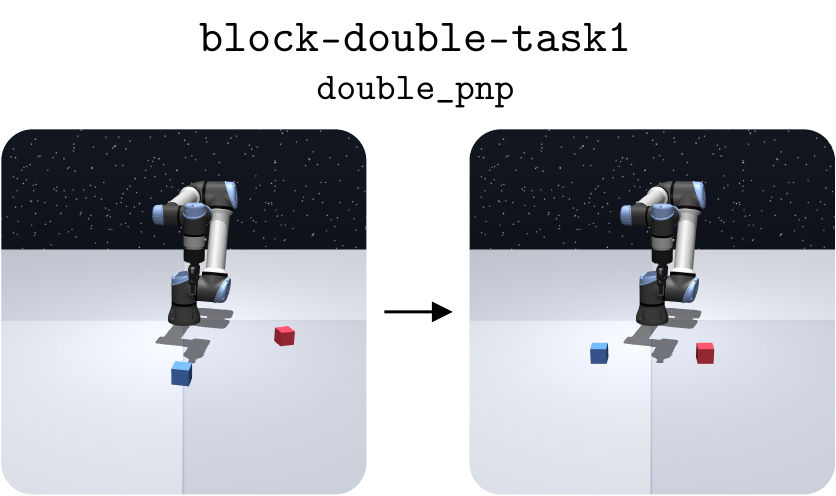}
    \end{subfigure}%
    \hspace{\dimexpr(\textwidth-0.30\textwidth*3-1sp)/2\relax}%
    \begin{subfigure}[t]{0.30\textwidth}
        \centering
        \includegraphics[width=1.0\textwidth]{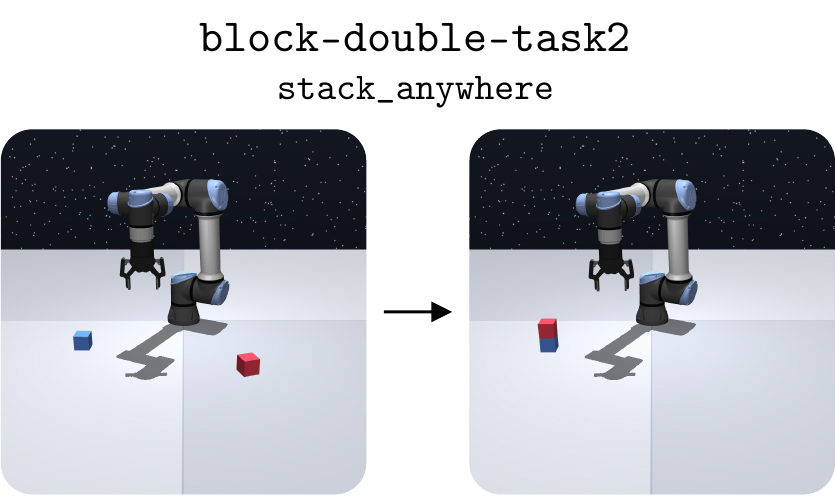}
    \end{subfigure}%

    \vspace{5pt}

    \begin{subfigure}[t]{0.30\textwidth}
        \centering
        \includegraphics[width=1.0\textwidth]{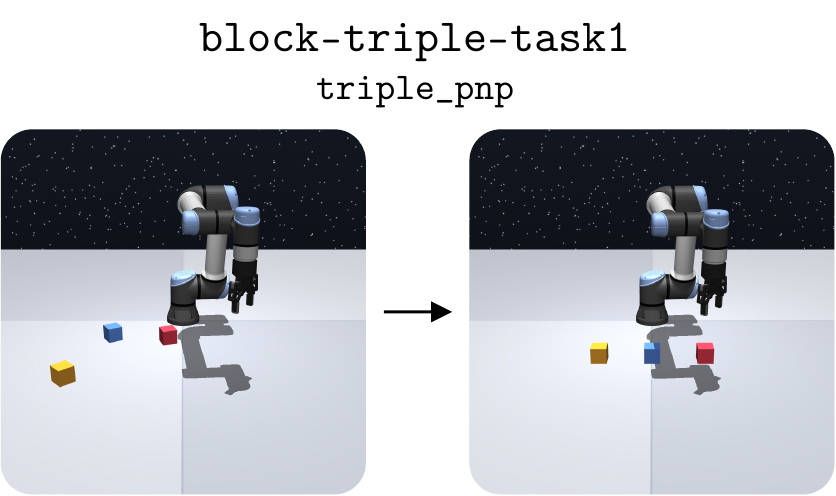}
    \end{subfigure}%
    \hspace{\dimexpr(\textwidth-0.30\textwidth*3-1sp)/2\relax}%
    \begin{subfigure}[t]{0.30\textwidth}
        \centering
        \includegraphics[width=1.0\textwidth]{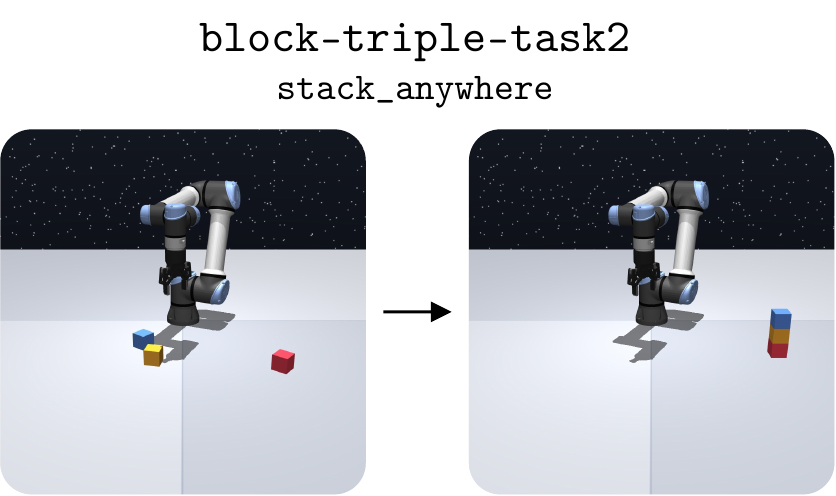}
    \end{subfigure}%

    \vspace{5pt}

    \begin{subfigure}[t]{0.30\textwidth}
        \centering
        \includegraphics[width=1.0\textwidth]{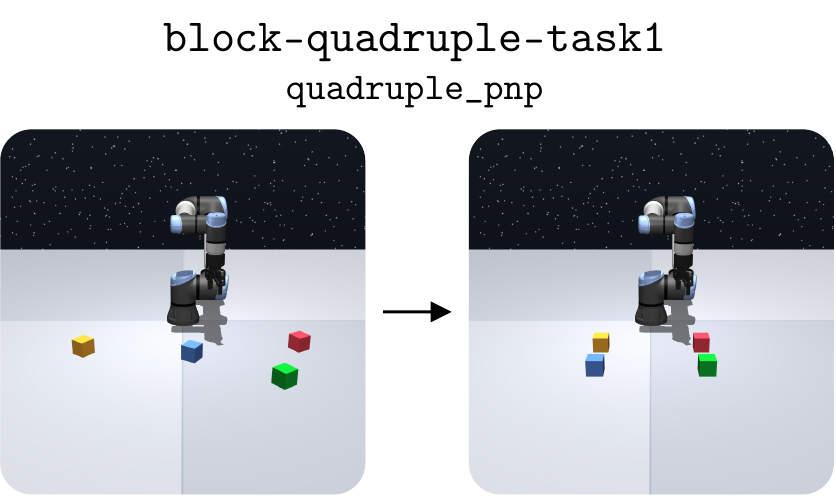}
    \end{subfigure}%
    \hspace{\dimexpr(\textwidth-0.30\textwidth*3-1sp)/2\relax}%
    \begin{subfigure}[t]{0.30\textwidth}
        \centering
        \includegraphics[width=1.0\textwidth]{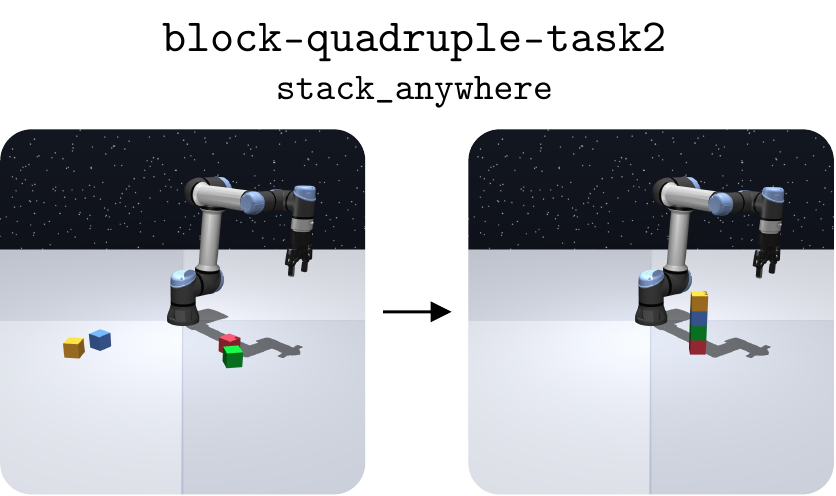}
    \end{subfigure}%
    \hspace{\dimexpr(\textwidth-0.30\textwidth*3-1sp)/2\relax}%
    \begin{subfigure}[t]{0.30\textwidth}
        \centering
        \includegraphics[width=1.0\textwidth]{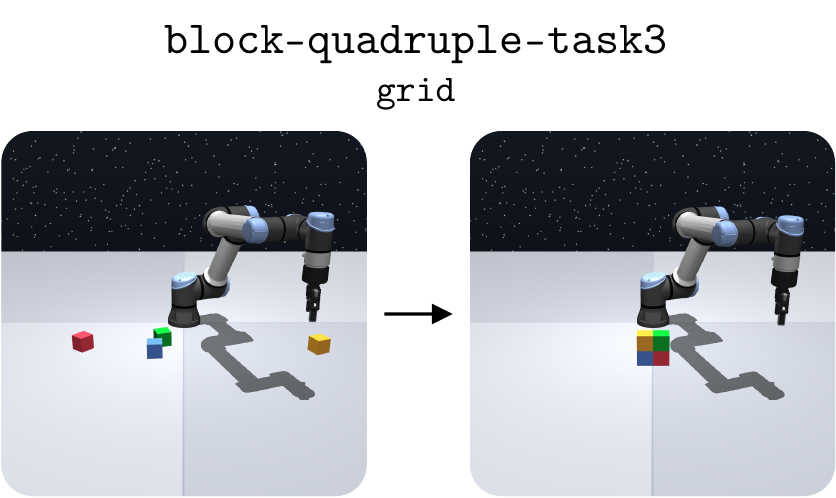}
    \end{subfigure}%

    \caption{
    \footnotesize
    \tt{block} tasks.
    }
    \label{fig:tasks_block}
\end{figure}

\begin{figure}[h!]
    \centering
    
    \begin{subfigure}[t]{0.30\textwidth}
        \centering
        \includegraphics[width=1.0\textwidth]{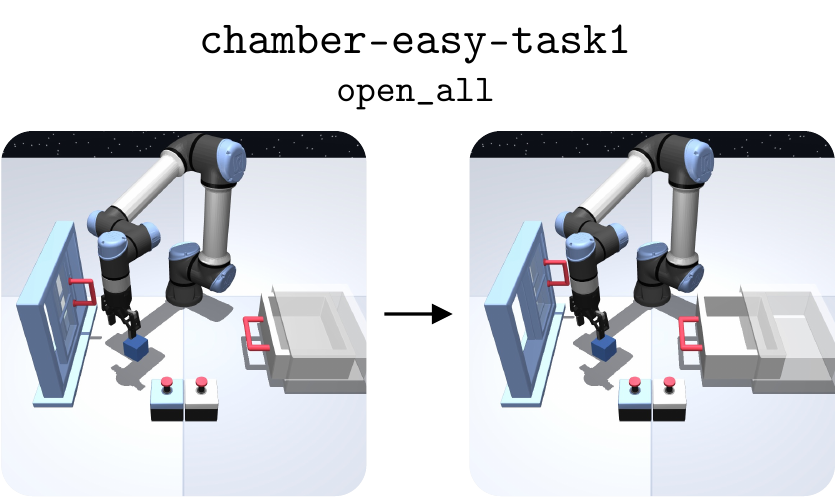}
    \end{subfigure}%
    \hspace{\dimexpr(\textwidth-0.30\textwidth*3-1sp)/2\relax}%
    \begin{subfigure}[t]{0.30\textwidth}
        \centering
        \includegraphics[width=1.0\textwidth]{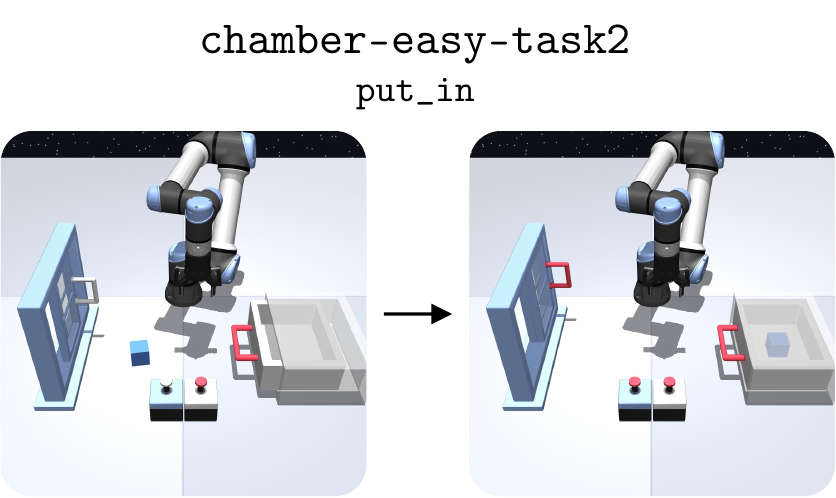}
    \end{subfigure}%
    \hspace{\dimexpr(\textwidth-0.30\textwidth*3-1sp)/2\relax}%
    \begin{subfigure}[t]{0.30\textwidth}
        \centering
        \includegraphics[width=1.0\textwidth]{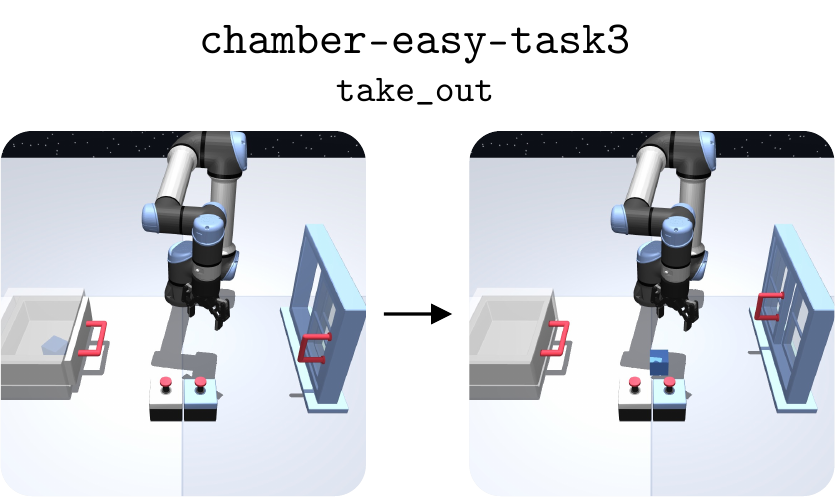}
    \end{subfigure}%

    \vspace{5pt}

    \begin{subfigure}[t]{0.30\textwidth}
        \centering
        \includegraphics[width=1.0\textwidth]{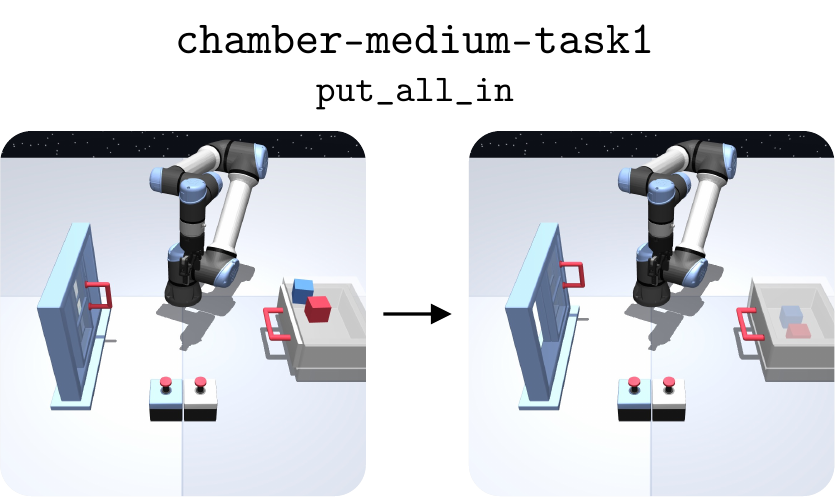}
    \end{subfigure}%
    \hspace{\dimexpr(\textwidth-0.30\textwidth*3-1sp)/2\relax}%
    \begin{subfigure}[t]{0.30\textwidth}
        \centering
        \includegraphics[width=1.0\textwidth]{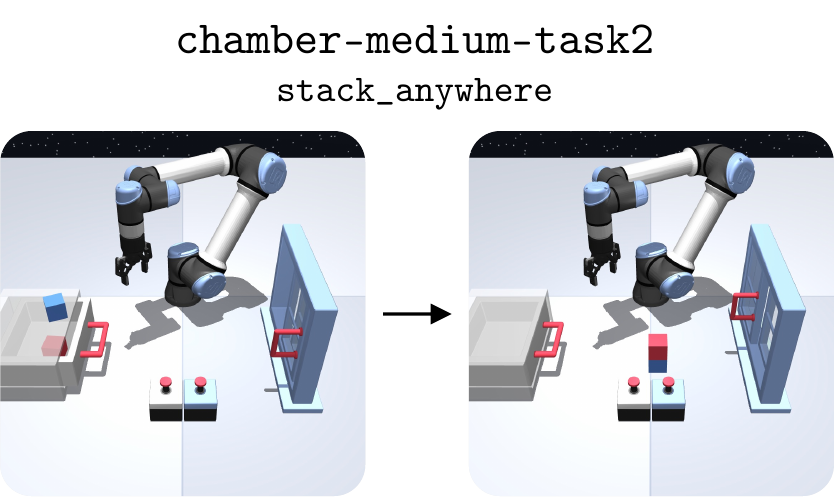}
    \end{subfigure}%

    \vspace{5pt}

    \begin{subfigure}[t]{0.30\textwidth}
        \centering
        \includegraphics[width=1.0\textwidth]{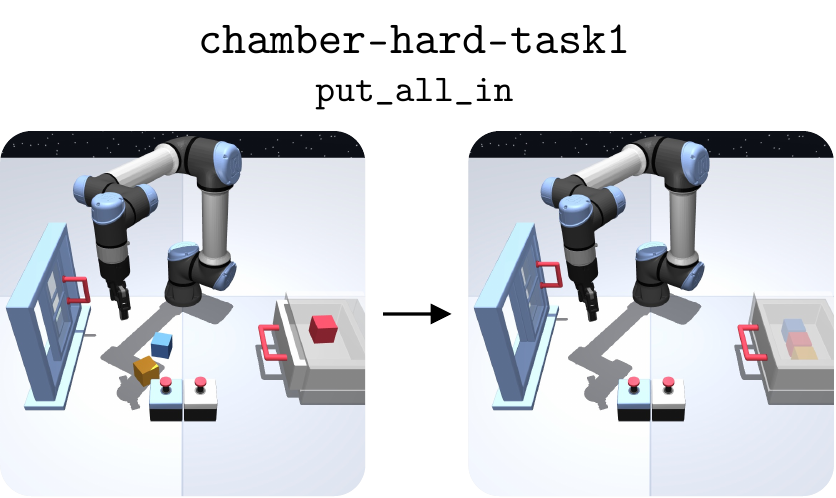}
    \end{subfigure}%
    \hspace{\dimexpr(\textwidth-0.30\textwidth*3-1sp)/2\relax}%
    \begin{subfigure}[t]{0.30\textwidth}
        \centering
        \includegraphics[width=1.0\textwidth]{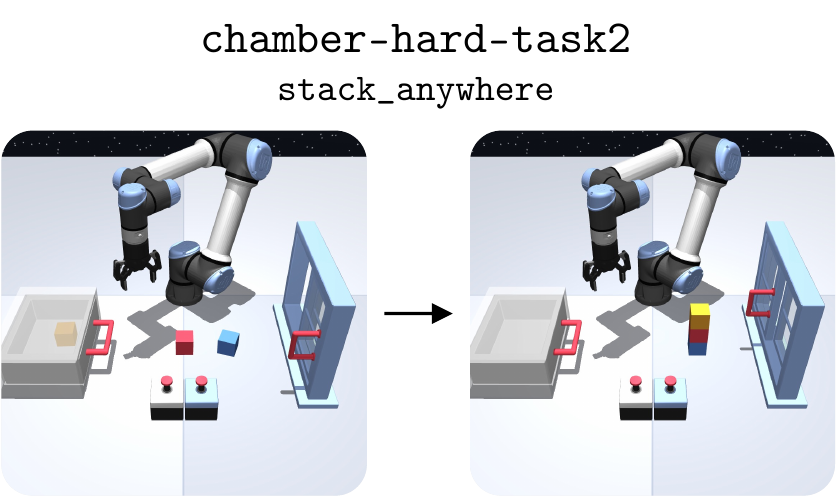}
    \end{subfigure}%

    \caption{
    \footnotesize
    \tt{chamber} tasks.
    }
    \label{fig:tasks_chamber}
\end{figure}

\begin{figure}[h!]
    \centering
    
    \begin{subfigure}[t]{0.30\textwidth}
        \centering
        \includegraphics[width=1.0\textwidth]{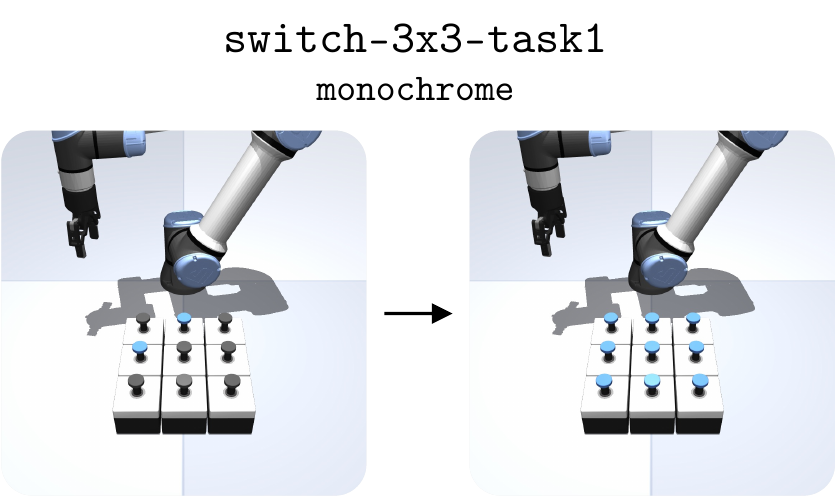}
    \end{subfigure}%
    \hspace{\dimexpr(\textwidth-0.30\textwidth*3-1sp)/2\relax}%
    \begin{subfigure}[t]{0.30\textwidth}
        \centering
        \includegraphics[width=1.0\textwidth]{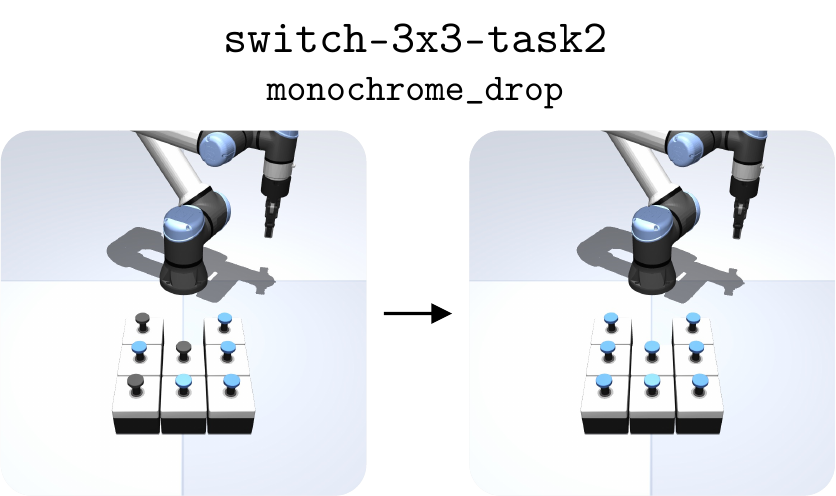}
    \end{subfigure}%

    \vspace{5pt}

    \begin{subfigure}[t]{0.30\textwidth}
        \centering
        \includegraphics[width=1.0\textwidth]{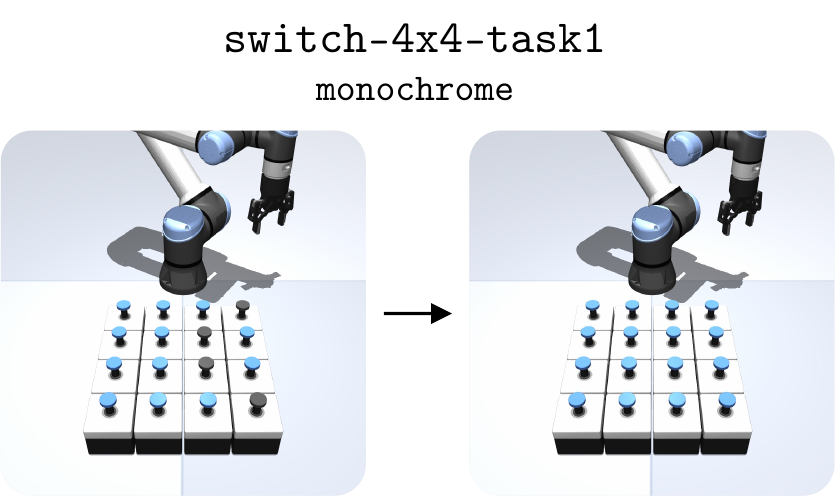}
    \end{subfigure}%
    \hspace{\dimexpr(\textwidth-0.30\textwidth*3-1sp)/2\relax}%
    \begin{subfigure}[t]{0.30\textwidth}
        \centering
        \includegraphics[width=1.0\textwidth]{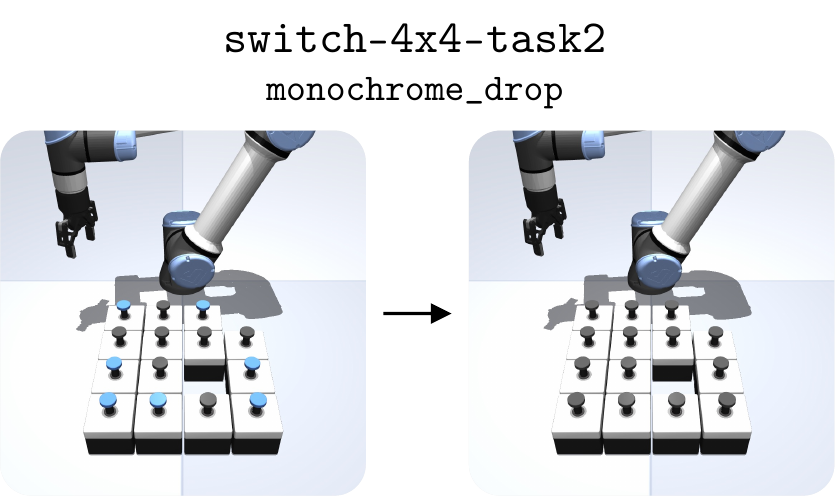}
    \end{subfigure}%

    \vspace{5pt}

    \begin{subfigure}[t]{0.30\textwidth}
        \centering
        \includegraphics[width=1.0\textwidth]{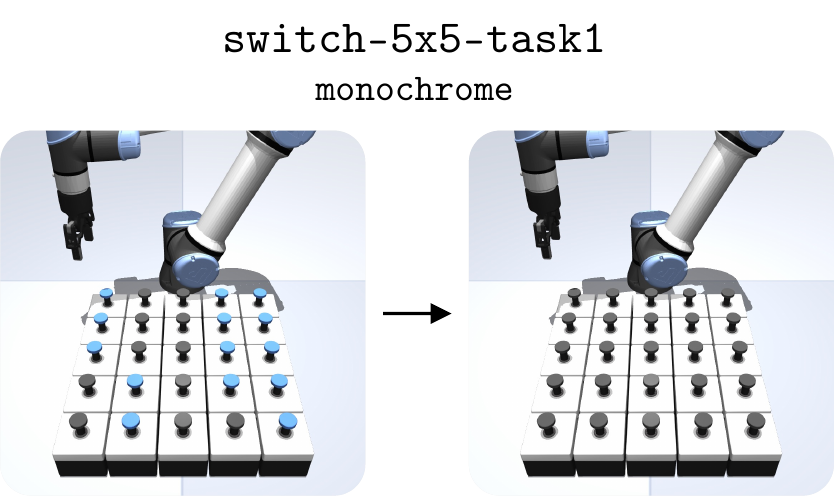}
    \end{subfigure}%
    \hspace{\dimexpr(\textwidth-0.30\textwidth*3-1sp)/2\relax}%
    \begin{subfigure}[t]{0.30\textwidth}
        \centering
        \includegraphics[width=1.0\textwidth]{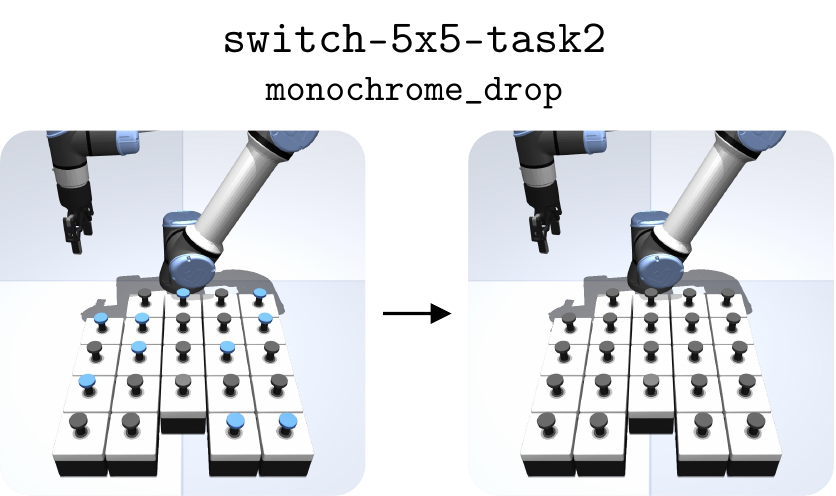}
    \end{subfigure}%

    \caption{
    \footnotesize
    \tt{switch} tasks.
    }
    \label{fig:tasks_switch}
\end{figure}

\begin{figure}[h!]
    \centering
    
    \begin{subfigure}[t]{0.30\textwidth}
        \centering
        \includegraphics[width=1.0\textwidth]{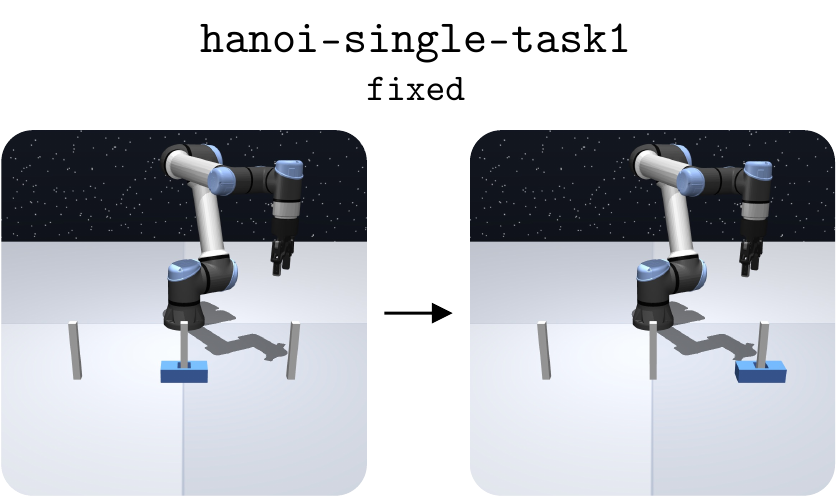}
    \end{subfigure}%
    \hspace{\dimexpr(\textwidth-0.30\textwidth*3-1sp)/2\relax}%
    \begin{subfigure}[t]{0.30\textwidth}
        \centering
        \includegraphics[width=1.0\textwidth]{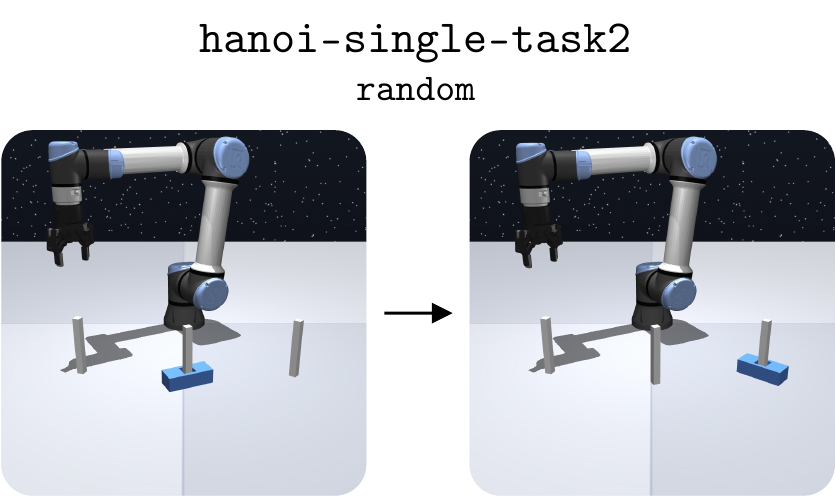}
    \end{subfigure}%

    \vspace{5pt}

    \begin{subfigure}[t]{0.30\textwidth}
        \centering
        \includegraphics[width=1.0\textwidth]{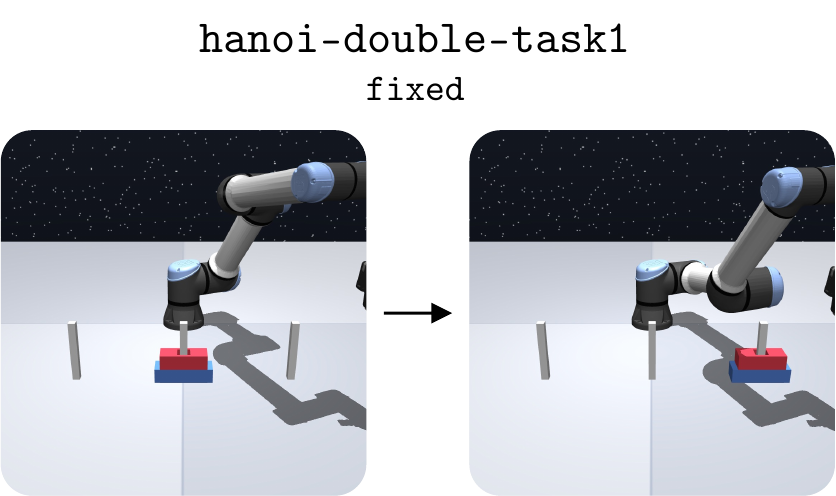}
    \end{subfigure}%
    \hspace{\dimexpr(\textwidth-0.30\textwidth*3-1sp)/2\relax}%
    \begin{subfigure}[t]{0.30\textwidth}
        \centering
        \includegraphics[width=1.0\textwidth]{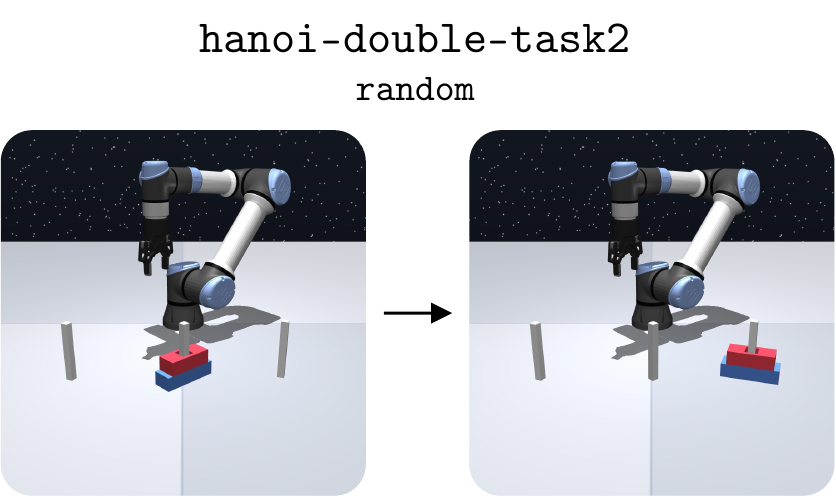}
    \end{subfigure}%

    \vspace{5pt}

    \begin{subfigure}[t]{0.30\textwidth}
        \centering
        \includegraphics[width=1.0\textwidth]{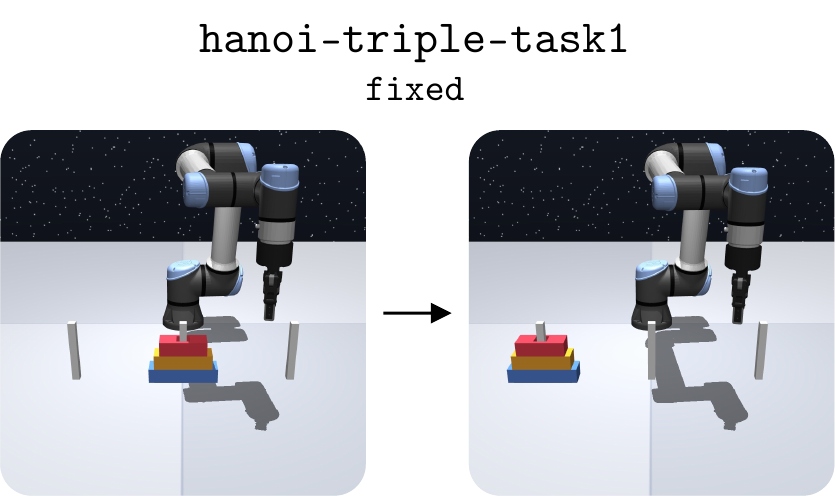}
    \end{subfigure}%
    \hspace{\dimexpr(\textwidth-0.30\textwidth*3-1sp)/2\relax}%
    \begin{subfigure}[t]{0.30\textwidth}
        \centering
        \includegraphics[width=1.0\textwidth]{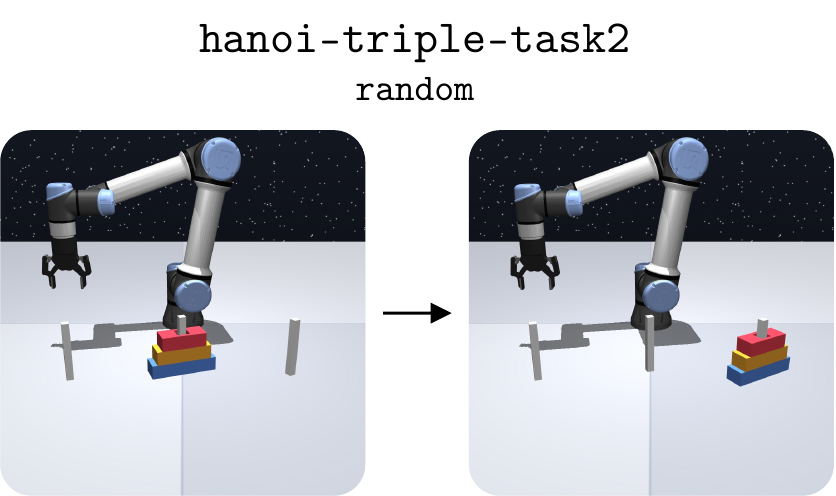}
    \end{subfigure}%

    \caption{
    \footnotesize
    \tt{hanoi} tasks.
    }
    \label{fig:tasks_hanoi}
\end{figure}

\begin{figure}[h!]
    \centering
    
    \begin{subfigure}[t]{0.30\textwidth}
        \centering
        \includegraphics[width=1.0\textwidth]{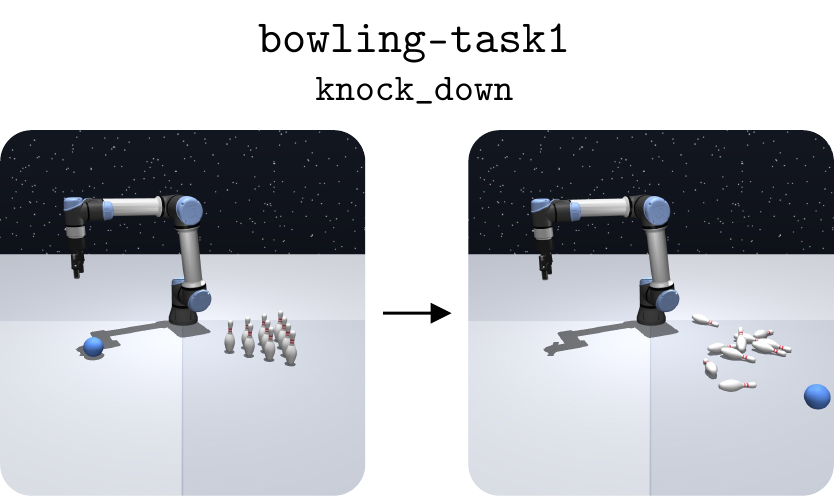}
    \end{subfigure}%

    \caption{
    \footnotesize
    \tt{bowling} task.
    }
    \label{fig:tasks_bowling}
\end{figure}

\begin{itemize}
\item \tt{block} ($8$ tasks in total) is a standard block manipulation environment (\Cref{fig:tasks_block}).
We provide four variants (\tt{single}, \tt{double}, \tt{triple}, and \tt{quadruple}) with one to four blocks,
with each environment variant defining up to three different tasks (\tt{task1}--\tt{task3}).

Across four environments, \tt{task1} is a fixed pick-and-place task,
where the agent must place the blocks at pre-defined positions,
starting from randomized initial states.
\tt{task2} is a ``stack anywhere'' task, where the goal is to stack the blocks anywhere on the table in any order.
\tt{task3} is available only in \tt{block-quadruple} and requires the agent to arrange the blocks in a grid at a fixed position.

\item \tt{chamber} ($7$ tasks in total) is an environment with multiple objects
(blocks, a drawer, a window, and button locks) that require different manipulation strategies
(\Cref{fig:tasks_chamber}).
We provide three variants (\tt{easy}, \tt{medium}, and \tt{hard}) with up to three different tasks (\tt{task1}--\tt{task3}).
The position and orientation of the objects are randomized,
the drawer and window are randomly swapped,
and demonstrations complete the subtasks in a random order, with occasional high-level mistakes.
Hence, the agent must deal with multimodal demonstrations over a long horizon (up to $5500$ steps).

\tt{chamber-easy-task1} is the simplest task and does not require block manipulation;
its goal is to open both the drawer and window (after unlocking them if necessary), and lock them again.
\tt{chamber-easy-task2} and \tt{chamber-\{medium, hard\}-task1} require the agent to put all blocks in the drawer and lock it,
and open the window and lock it.
\tt{chamber-easy-task3} and \tt{chamber-\{medium, hard\}-task2} require the agent to pull all blocks out of the drawer,
stack them anywhere on the table in any order,
and leave both the window and drawer closed and locked.

\item \tt{switch} ($6$ tasks in total) is an environment based on the Lights Out puzzle, which requires combinatorial reasoning (\Cref{fig:tasks_switch}).
We provide three board sizes (\tt{3x3}, \tt{4x4}, and \tt{5x5}) and two variants:
\tt{task1} provides a fixed grid, and \tt{task2} provides a randomized grid
where at most one (\tt{3x3}) or two (\tt{4x4} and \tt{5x5}) randomly selected buttons are omitted.
When the agent presses a button, it toggles the color of the pressed button and up to four adjacent buttons.
The goal is to make all buttons the same color (either all blue or all black).
The demonstrations for these tasks are \emph{highly} suboptimal, unlike those for the other environments in OCBench,
so this environment examines the ability to learn from suboptimal data.

\item \tt{hanoi} ($6$ tasks in total) is an environment based on the Tower of Hanoi puzzle (\Cref{fig:tasks_hanoi}).
We provide three variants (\tt{single}, \tt{double}, and \tt{triple}) with one to three disks,
with each variant defining two tasks:
\tt{task1} uses fixed peg positions, and \tt{task2} randomizes the positions of the pegs.
The disks are initially stacked on the middle peg,
and the goal is to move them to \emph{either} the first or third peg,
while following the rules of the Tower of Hanoi.
Since the disks fit tightly around the pegs,
these tasks require highly precise manipulation maneuvers over up to $7000$ environment steps.

\item \tt{bowling} ($1$ task) is a task whose goal is to push a bowling ball to knock down as many pins as possible (\Cref{fig:tasks_bowling}).
Since the dynamics are not easily predictable due to the shape of the bowling pins,
this task requires understanding of the environment dynamics from highly diverse outcomes.
\end{itemize}

Among them, the first three types of environments are modified from
the \tt{cube}, \tt{scene}, and \tt{puzzle} environments in OGBench~\citep{park2025ogbench},
although we made substantial modifications
to make them more suitable for standard robotic manipulation studies (see \Cref{sec:prev_bench}).
As in OGBench, we use assets from MuJoCo Menagerie~\citep{zakka2022mujoco} and Meta-World~\citep{yu2020meta} for the UR5e robot arm and objects.

\textbf{Lite tasks.}
In OCBench, we also provide \tt{lite} versions of all original tasks except for \tt{bowling}.
These tasks have shorter horizons and simplified scripted policies.
The main differences are threefold:
(1) \tt{lite} tasks use $10$Hz control whereas the original ones use $50$Hz,
(2) the scripted policies for \tt{lite} tasks move approximately $2\times$ as fast as the original ones
(so if we combine this with the reduction from $50$Hz to $10$Hz, \tt{lite} tasks have approximately $10\times$ shorter horizons in total),
and (3) the scripted policies for \tt{lite} tasks are simpler (they do not randomize the gripper's roll and pitch).
As a result, \tt{lite} tasks can be learned with a smaller network (\eg, $[1024] \times 8$-sized networks, as opposed to $[4096] \times 8$ for full versions).
We expect \tt{lite} tasks to be useful for faster initial iteration on algorithmic ideas.

\begin{wrapfigure}{r}{0.5\textwidth}
    \centering
    \raisebox{0pt}[\dimexpr\height-1.0\baselineskip\relax]{
        \begin{subfigure}[t]{1.0\linewidth}
        \includegraphics[width=\linewidth]{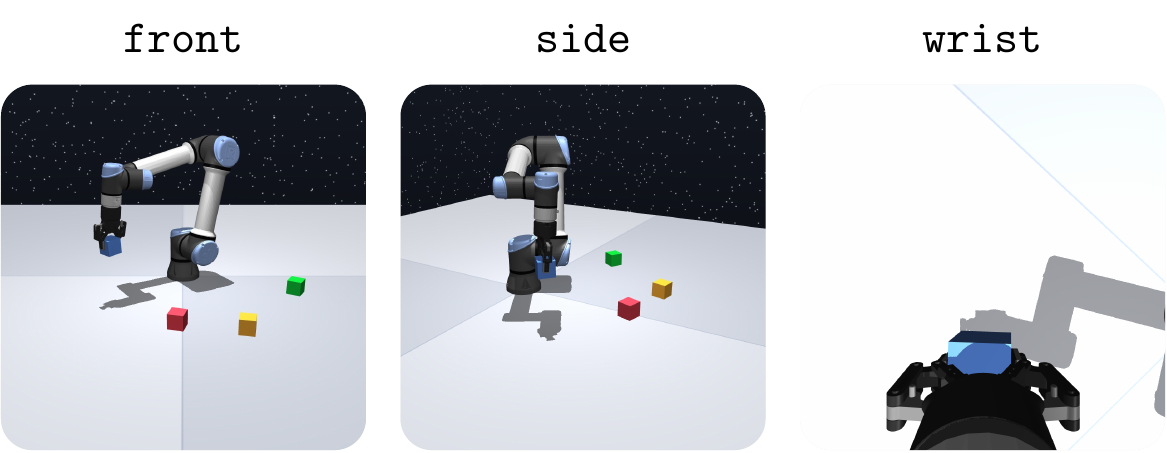}
        \end{subfigure}
    }
    \caption{
    \footnotesize
    Camera observations ($224 \times 224 \times 3$).
    }
    \label{fig:visual}
\end{wrapfigure}
\textbf{Pixel-based tasks.}
We also provide \tt{visual} versions of OCBench tasks for pixel-based control.
While we only provide datasets for selected tasks due to storage constraints,
one can easily generate data using OCBench's data generation scripts.
A visual observation consists of three camera images of size $224 \times 224 \times 3$ (front, side, and wrist; see \Cref{fig:visual}).
In the provided datasets, images are recorded every $25$ steps (\ie, one image set per length-$25$ action chunk) to save space.
We found that this still allows reasonably high performance while substantially reducing the dataset size and speeding up data generation.

\textbf{Rewards and metrics.}
All tasks except for \tt{bowling} use binary sparse rewards.
If the agent successfully completes the task, it receives a reward of $1$ and the episode immediately terminates.
Otherwise, it receives a reward of $0$.
On \tt{bowling}, the agent receives a reward of $0.1$ for each pin it knocks down for the first time.
Across all environments, the sum of rewards is always bounded by $[0, 1]$.

For all tasks except for \tt{bowling}, we mainly consider two metrics: success rate and throughput.
Success rate is the ratio of successful episodes to total episodes.
Throughput is defined as follows:
\begin{align}
(\mathrm{Throughput}) =
(\mathrm{Success\ rate}) \times
\frac{(\mathrm{Maximum\ episode\ length})}{(\mathrm{Mean\ episode\ length})},
\end{align}
where the mean episode length is computed over all episodes, including failures.
For \tt{bowling}, we consider the normalized average return, defined as the average return multiplied by $100$.
Throughout the tables (\Cref{table:dataset_specs,table:benchmark_sr}),
\tt{bowling} numbers denote this normalized average return instead of the success rate.

\textbf{Auxiliary information.}
OCBench datasets provide additional information beyond standard observations and actions.
First, we provide full simulation states
(\tt{qpos}, \tt{qvel}, \verb|button_states|, and \verb|dynamics_info|)
from which observations can be perfectly reconstructed.
Second, we provide segment information for each atomic subtask in a trajectory
(\eg, opening a drawer, pressing a button, etc.),
including subtask types, target information, subtask success indicators,
and high-level mistake indicators.
We expect this auxiliary information to be useful for
research in RL and hierarchical control
as well as for more fine-grained analyses.

\section{How is OCBench different from previous synthetic benchmarks?}
\label{sec:prev_bench}

One of the main contributions of OCBench is that it features more human-like scripted policies.
This enables reproducing many BC-related phenomena (see \Cref{sec:motivation}) that happen in the real world.
In this section, we discuss
how OCBench's scripted policies differ from those of previous benchmarks based on synthetic data (collected by scripted or RL-trained policies) in detail,
showing how previous synthetic benchmarks might not fully reproduce all these phenomena.

\begin{figure}[h!]
    \centering
    \includegraphics[width=0.5038\textwidth]{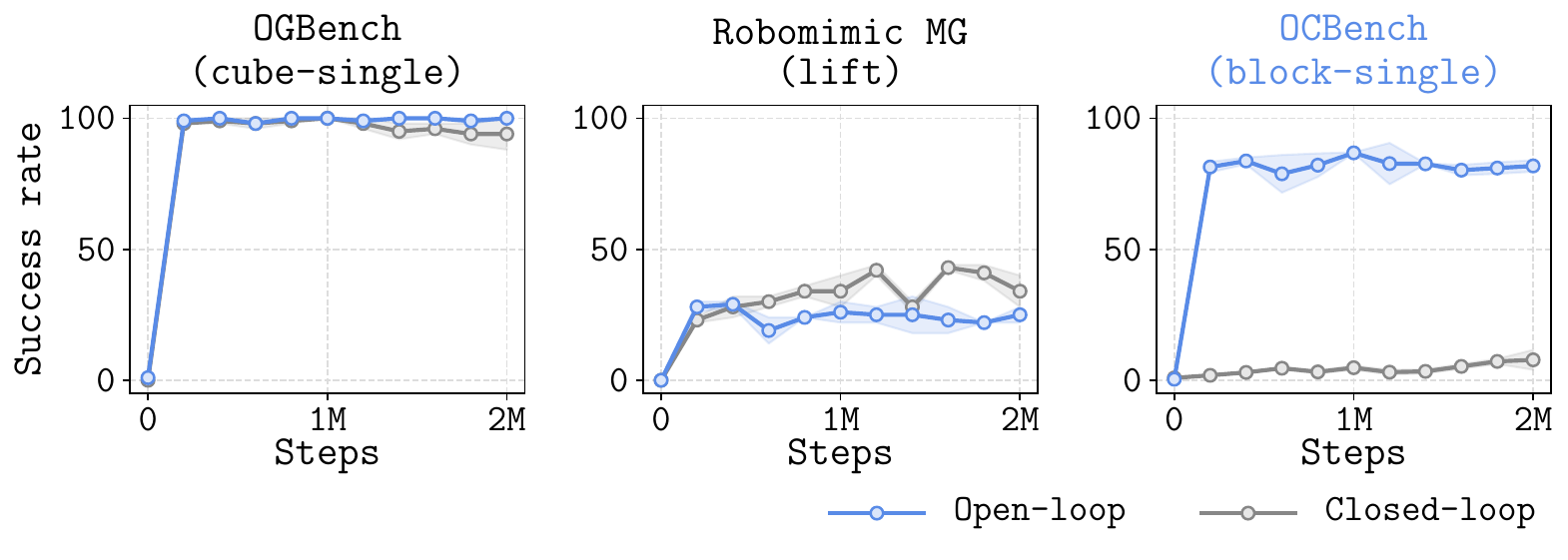}
    \caption{
    \footnotesize
    The difference between closed-loop and open-loop policies is less pronounced in previous synthetic benchmarks.
    }
    \label{fig:prev_open_closed}
\end{figure}

\textbf{Open-loop vs. closed-loop control (\Cref{sec:mystery2}).}
\Cref{fig:prev_open_closed} compares the performance of fully closed-loop policies and partially open-loop (action-chunking) policies on
standard single-cube pick-and-place tasks from three benchmarks:
OGBench~\citep{park2025ogbench}, Robomimic~\citep{mandlekar2021matters}, and OCBench.
For Robomimic, we use machine-generated (MG) datasets, as our goal in this experiment is to compare policies trained on synthetic data.
The results suggest that the gap between closed- and open-loop policies is much less pronounced in previous benchmarks,
than in OCBench or with human-generated data~\citep{zhao2023learning, lazzati2026does, zeng2026revisiting}.
Although action chunking does benefit long-horizon tasks in OGBench~\citep{li2026reinforcement},
synthetic policies in previous benchmarks typically generate trajectories that have (much) less temporal correlation than human demonstrations,
which could explain why the gap between closed- and open-loop control is generally smaller in these previous benchmarks.

\begin{figure}[h!]
    \centering
    \includegraphics[width=1.0\textwidth]{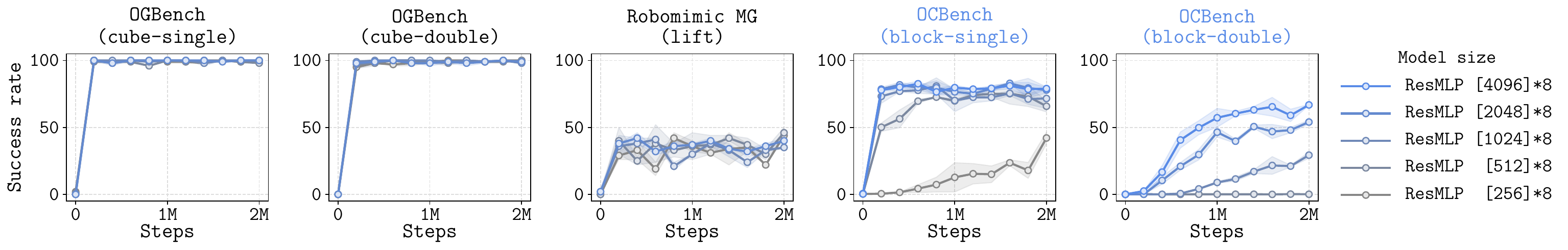}
    \caption{
    \footnotesize
    A smaller network ($[256] \times 8$) can saturate performance on previous synthetic benchmarks.
    In contrast, OCBench requires larger networks (as is the case with human demonstrations) even for tasks of similar semantic difficulty.
    }
    \label{fig:prev_model_size}
\end{figure}

\textbf{Model size ablation (\Cref{sec:mystery3}).}
\Cref{fig:prev_model_size} shows how different model sizes affect performance
on five single- or double-cube pick-and-place tasks across three benchmarks.
We use \tt{cube-\{single, double\}} from OGBench~\citep{park2025ogbench}, \tt{lift} from Robomimic MG~\citep{mandlekar2021matters},
and \tt{block-\{single, double\}} from OCBench.
The results show that relatively small networks ($[256] \times 8$) already saturate performance on previous benchmarks,
whereas OCBench tasks benefit from larger networks of up to hundreds of millions of parameters
even for tasks of similar semantic complexity (\eg, double pick-and-place).
This model-size scaling property resembles that observed with human demonstrations~\citep{pan2026much}.
We believe this is mainly because OCBench's trajectories are more narrowly distributed and temporally correlated
than the Gaussian noise-augmented trajectories from the previous benchmarks.

\textbf{Action distribution comparison (\Cref{fig:action_hist}).}
In the main paper, we qualitatively compare the action distributions
from previous synthetic benchmarks (OGBench and Robomimic MG),
OCBench, and human demonstrations (\Cref{fig:action_hist}).
Specifically, we plot per-dimension action distributions computed
on \tt{cube-double-play} for OGBench,
\tt{can} for Robomimic MG,
\tt{block-double-task1} for OCBench,
and BiPlay~\citep{dasari2025ingredients} for human demonstrations.
For OGBench, since the original dataset is designed for goal-conditioned RL,
we re-generate a single-task dataset on OCBench's \tt{block-double-task1} using OGBench's scripted policy
to ensure compatibility.
For BiPlay, we convert the original absolute action targets into relative ones
to make them compatible with the relative action representations in the other benchmarks.

While the absolute action values in the action distributions in \Cref{fig:action_hist}
may not be directly comparable due to the difference in joint ranges from different robots,
the results nevertheless highlight that previous synthetic data
often have qualitatively different \emph{properties} from those of human demonstrations.
For example, they often have highly bimodal actions peaked at $-1$ or $+1$
or have overly uniform distributions,
unlike human demonstrations.
On the other hand, OCBench's action distributions more closely resemble those of human demonstrations.

\begin{table}[t!]
\caption{
\footnotesize
\textbf{OGBench vs. OCBench.}
OCBench provides more human-like scripted policies,
is not limited to goal-conditioned tasks, and supports GPU acceleration.
}
\label{table:ogbench}
\centering
\setlength{\tabcolsep}{4pt}
\scalebox{1.0}{%
\begin{tabular}{lcc}
\toprule
\textbf{Property} & \textbf{OGBench}~\citep{park2025ogbench} & \textbf{OCBench (ours)} \\
\midrule
Task type & Goal-reaching tasks & General robotic tasks \\
Action space & $5$-D end effector control & $7$-D joint position control \\
Action frequency & $20$Hz & $50$Hz \\
Environment randomization & No & Yes \\
GPU acceleration & No & Yes \\
Scripted policy & More rule-based & More human-like \\
\bottomrule
\end{tabular}
}
\vspace{-5pt}
\end{table}

\textbf{OGBench vs. OCBench.}
OGBench~\citep{park2025ogbench} is the closest predecessor of OCBench.
Here, we highlight three key differences between OGBench and OCBench.
Unlike OGBench,
(1) OCBench's scripted policies are explicitly designed to mimic the properties of human demonstrations (see \Cref{fig:action_hist,fig:prev_open_closed,fig:prev_model_size}),
(2) it supports MJWarp-based GPU acceleration,
and (3) it provides standard robotic tasks not limited to goal-conditioned ones.
We summarize the main differences in \Cref{table:ogbench}.

\section{Additional results}
\label{sec:additional_results}

\begin{figure}[h!]
    \centering
    \begin{subfigure}[t]{0.8199\textwidth}
        \centering
        \includegraphics[width=1.0\textwidth]{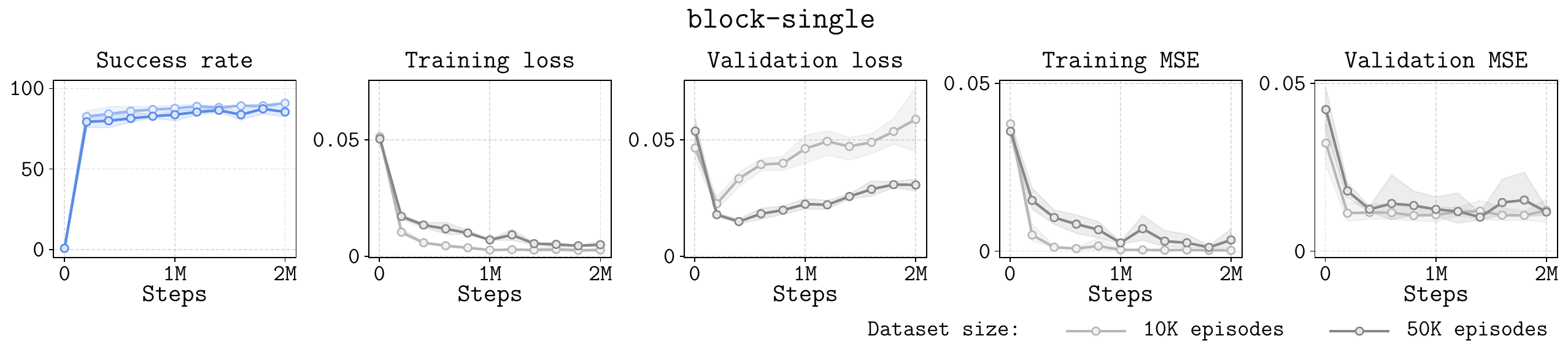}
    \end{subfigure}
    \begin{subfigure}[t]{0.8199\textwidth}
        \centering
        \includegraphics[width=1.0\textwidth]{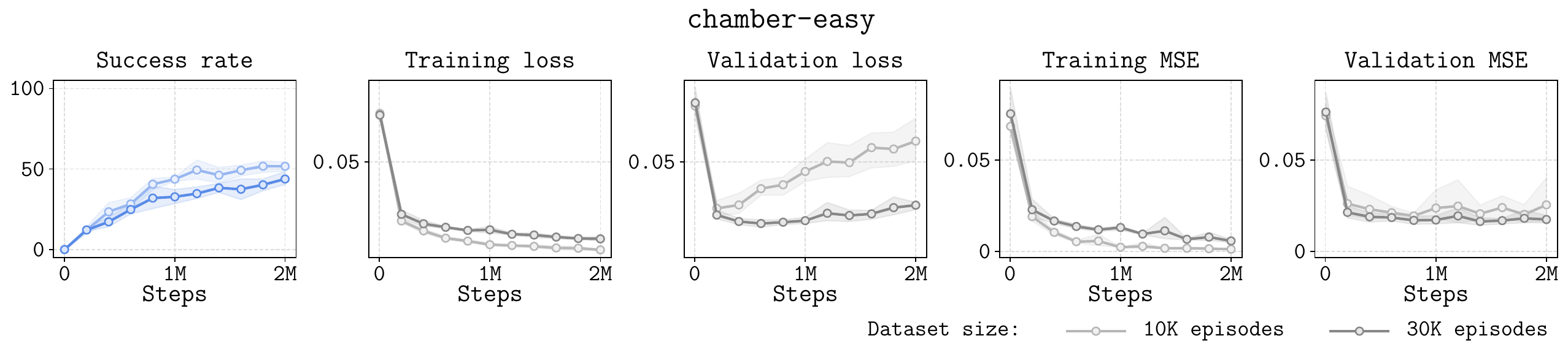}
    \end{subfigure}
    \caption{
    \footnotesize
    The action MSE metric does not fully explain the dataset size-reversal phenomenon.
    }
    \label{fig:dataset_size_mse}
\end{figure}

\textbf{Additional plots.}
\Cref{fig:dataset_size_mse} compares performance, training and validation flow losses, and training and validation action MSEs
on datasets of different sizes.
The results show that the MSE metric still does not fully explain the size-reversal phenomenon (see \Cref{sec:mystery1} for the context).

\begin{figure}[h!]
    \centering
    \begin{subfigure}[t]{0.4962\textwidth}
        \centering
        \includegraphics[width=1.0\textwidth]{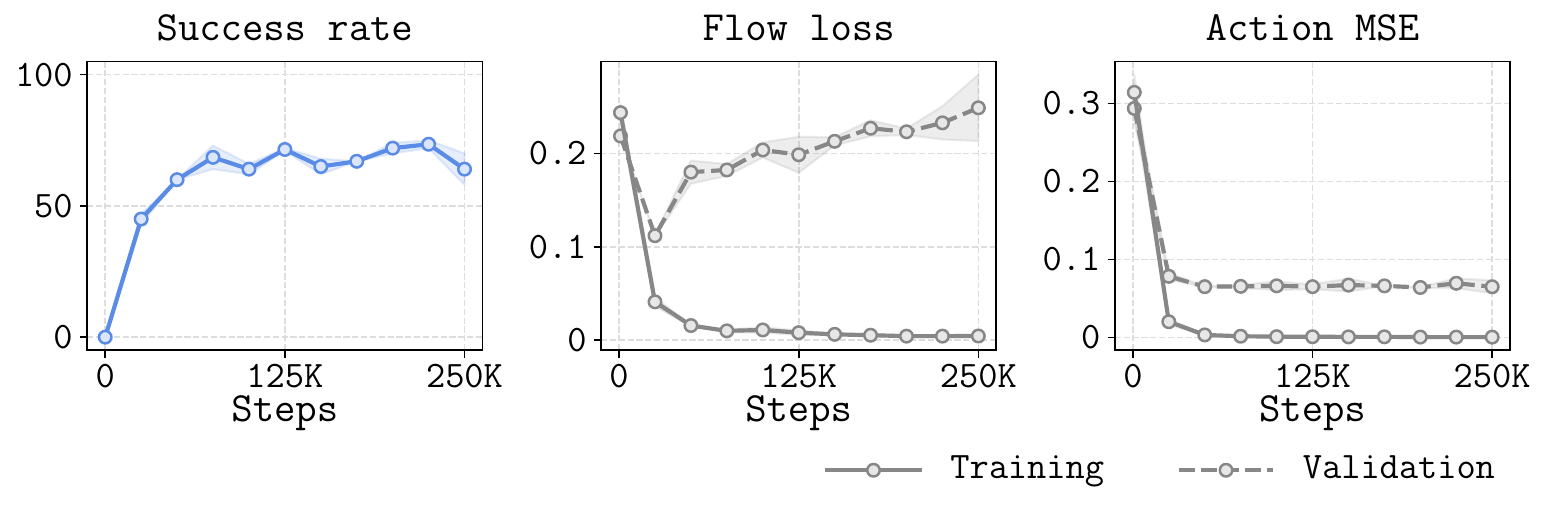}
        \caption{Overfitting is not necessarily bad.}
    \end{subfigure}%
    \quad%
    \begin{subfigure}[t]{0.2462\textwidth}
        \centering
        \includegraphics[width=1.0\textwidth]{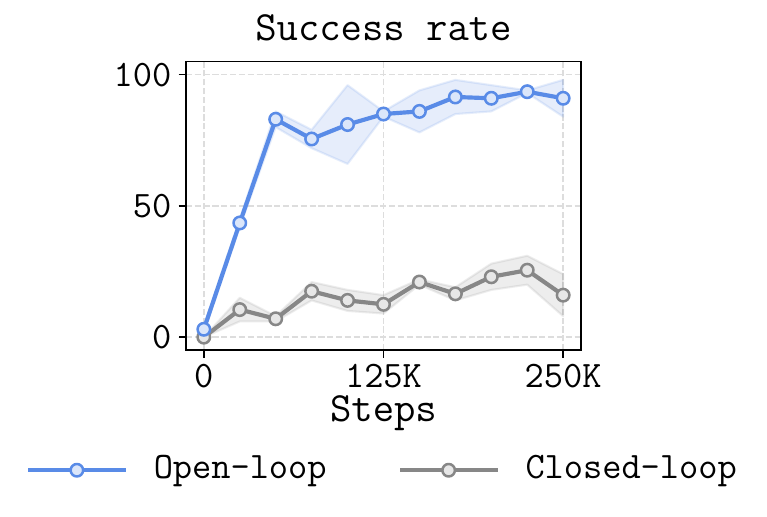}
        \caption{Closed-loop control fails.}
    \end{subfigure}

    \medskip
    \begin{subfigure}[t]{0.5482\textwidth}
        \centering
        \includegraphics[width=1.0\textwidth]{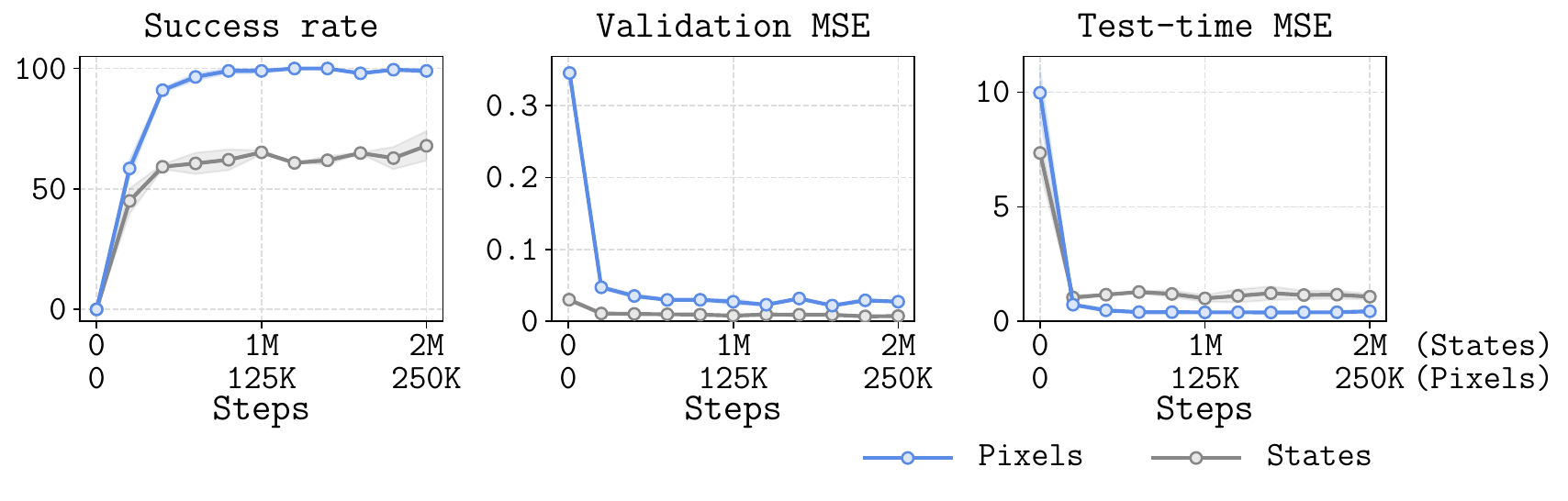}
        \caption{Features matter.}
    \end{subfigure}
    \caption{
    \footnotesize
    Pixel-based OCBench tasks reproduce BC-related phenomena too.
    }
    \label{fig:pixel_results}
\end{figure}

\textbf{Results with pixel-based tasks.}
While we focus on state-based tasks for computational reasons in this work,
we also present preliminary results showing that pixel-based OCBench tasks can be used to study the same BC-related phenomena too.
After all, the phenomena studied in this work are ``reproductions'' of real-world phenomena observed with visual data~\citep{zhao2023learning, mark2026bpp, allshire2026scalable}.

\Cref{fig:pixel_results} reproduces three phenomena
on visual OCBench tasks (\tt{visual-block-single-task1} for (a) and (b), and \tt{visual-hanoi-single-task1} for (c)).
\Cref{fig:pixel_results}a shows that overfitting does not necessarily hurt performance (\Cref{fig:overfitting}).
\Cref{fig:pixel_results}b shows that open-loop control substantially improves performance over closed-loop control (\Cref{fig:open_closed}).
\Cref{fig:pixel_results}c shows that test-time MSE can have an opposite trend to validation MSE for two different features (states vs. pixels),
and that this test-time generalization metric can better explain the actual performance (\Cref{fig:test_time_generalization}).

\textbf{Benchmark results.}
While the main focus of this paper is analysis rather than benchmarking and comparing different algorithms,
we provide a preliminary benchmark result for three BC algorithms (BC, flow BC, and FSQ BC)
in \Cref{table:benchmark_sr,table:benchmark_throughput} as a reference.
The results are averaged over two seeds and $\pm$ denotes a standard deviation.

\section{Implementation details}
\label{sec:impl_details}

We describe the full implementation details in this section.
We also provide the code, datasets, and exact command-line arguments for reproducing the benchmark tables (\Cref{table:benchmark_sr,table:benchmark_throughput}) at
\mbox{\ocbenchrepo}.

\subsection{Methods}

We implement all methods in JAX~\citep{bradbury2018jax}.
By default, we train BC agents only using successful episodes (except for \tt{bowling})
with length-$25$ action chunks (length-$3$ action chunks for \tt{lite} tasks).
We use residual MLPs (ResMLPs) with residual streams over every two adjacent layers.
We follow the action normalization scheme used in $\pi_{0.5}$~\citep{intelligence2025pi_}:
we first compute the $1$st and $99$th percentiles for each action dimension except for the gripper dimension,
map them to $-1$ and $1$ respectively,
and clip normalized values to $[-5, 5]$.
For the gripper dimension, we simply clip the values to $[-1, 1]$ without normalization.
In the benchmark table, we train agents with $2$M steps, %
evaluate each agent with $500$ rollouts every $50$K steps,
and report the performance averaged over the last $5$ evaluation steps
(\{$1.8$M, $1.85$M, $1.9$M, $1.95$M, $2$M\} steps).
For streaming-data runs, we collect $25$K episodes every $100$K steps (and discard old data) by default.

\textbf{Behavioral cloning (BC).}
Vanilla behavioral cloning trains a deterministic policy with a standard MSE loss.

\textbf{Flow behavioral cloning (FBC).}
Flow behavioral cloning trains a velocity field with standard flow matching~\citep{lipman2022flow, liu2022flow, albergo2022building}.
We use $\mathrm{Beta}(1, 1.5)$ (\ie, emphasizing the noisy part)
for the flow time distribution, similarly to \citet{black2024pi_0}.
This differs from previous flow-based works on OGBench~\citep{park2025flow, li2026reinforcement},
which use a uniform time distribution.
We found that the use of the beta distribution substantially improves training speed in OCBench,
whereas it makes no clear difference in OGBench.

\textbf{FSQ behavioral cloning (FSQBC).}
FSQ behavioral cloning jointly trains two models:
(1) a state-conditional finite scalar quantization (FSQ)~\citep{mentzer2024finite} autoencoder
that encodes an action chunk into discrete tokens and reconstructs it using an $\ell_1$ loss,
and (2) an autoregressive BC network that predicts these tokens given a state.
By default, we encode each action chunk into four tokens, each with quantization levels of $[8, 5, 5, 5]$.
This gives us up to $(8 \times 5 \times 5 \times 5)^4 = 1000^4$ possible action chunks given a state.

\textbf{Pixel-based behavioral cloning.}
To handle visual observations, we use a ConvNeXt encoder~\citep{woo2023convnext}.
Specifically, we encode each of the three images using a shared ConvNeXt V2-N encoder, %
concatenate the three $640$-dimensional representations into a $1920$-dimensional vector,
and feed it to a standard $[4096] \times 8$-sized residual MLP.
For training, we apply random-crop image augmentation by default.

\subsection{Hyperparameters}
We provide the default hyperparameters used in our experiments in \Cref{table:hyp}.
We refer the reader to the code for details.

\begin{table}[t!]
\caption{
\footnotesize
\textbf{Default hyperparameters.}
}
\vspace{-5pt}
\label{table:hyp}
\begin{center}
\scalebox{1.0}
{
\begin{tabular}{ll}
    \toprule
    \textbf{Hyperparameter} & \textbf{Value} \\
    \midrule
    Gradient steps & $2$M (default), $250$K (\tt{visual}) \\
    Optimizer & Adam~\citep{kingma2014adam} \\
    Learning rate & $0.0001$ \\
    Batch size & $1024$ (default), $256$ (FSQBC, \tt{visual}) \\
    ResMLP size (BC, FBC) & $[4096] \times 8$ (default), $[1024] \times 8$ (\tt{lite}) \\
    ResMLP size (FSQBC) & $[2048] \times 8$ (default), $[512] \times 8$ (\tt{lite}) \\
    MLP normalization & Layer normalization~\citep{ba2016layer} \\
    Action chunk length & $25$ (default), $3$ (\tt{lite}) \\
    \bottomrule
\end{tabular}
}
\end{center}
\end{table}

\clearpage

\clearpage

\begin{table}[t!]
\caption{
\footnotesize
\textbf{State-based task specifications.}
}
\label{table:task_specs}
\centering
\small
\setlength{\tabcolsep}{4pt}
\scalebox{0.84}{%
\begin{tabular}{lrrr}
\toprule
\texttt{Task} & \texttt{State dimensions} & \texttt{Action dimensions} & \texttt{Maximum episode length} \\
\midrule
\texttt{block-single-task1-v0} & \texttt{28} & \texttt{7} & \texttt{1250} \\
\texttt{block-double-task1-v0} & \texttt{37} & \texttt{7} & \texttt{2500} \\
\texttt{block-double-task2-v0} & \texttt{37} & \texttt{7} & \texttt{2500} \\
\texttt{block-triple-task1-v0} & \texttt{46} & \texttt{7} & \texttt{3750} \\
\texttt{block-triple-task2-v0} & \texttt{46} & \texttt{7} & \texttt{3750} \\
\texttt{block-quadruple-task1-v0} & \texttt{55} & \texttt{7} & \texttt{5000} \\
\texttt{block-quadruple-task2-v0} & \texttt{55} & \texttt{7} & \texttt{5000} \\
\texttt{block-quadruple-task3-v0} & \texttt{55} & \texttt{7} & \texttt{5000} \\
\midrule
\texttt{chamber-easy-task1-v0} & \texttt{53} & \texttt{7} & \texttt{4000} \\
\texttt{chamber-easy-task2-v0} & \texttt{53} & \texttt{7} & \texttt{4000} \\
\texttt{chamber-easy-task3-v0} & \texttt{53} & \texttt{7} & \texttt{4000} \\
\texttt{chamber-medium-task1-v0} & \texttt{62} & \texttt{7} & \texttt{5000} \\
\texttt{chamber-medium-task2-v0} & \texttt{62} & \texttt{7} & \texttt{5000} \\
\texttt{chamber-hard-task1-v0} & \texttt{71} & \texttt{7} & \texttt{5500} \\
\texttt{chamber-hard-task2-v0} & \texttt{71} & \texttt{7} & \texttt{5500} \\
\midrule
\texttt{switch-3x3-task1-v0} & \texttt{64} & \texttt{7} & \texttt{1500} \\
\texttt{switch-3x3-task2-v0} & \texttt{64} & \texttt{7} & \texttt{1500} \\
\texttt{switch-4x4-task1-v0} & \texttt{99} & \texttt{7} & \texttt{1500} \\
\texttt{switch-4x4-task2-v0} & \texttt{99} & \texttt{7} & \texttt{1500} \\
\texttt{switch-5x5-task1-v0} & \texttt{144} & \texttt{7} & \texttt{4000} \\
\texttt{switch-5x5-task2-v0} & \texttt{144} & \texttt{7} & \texttt{4000} \\
\midrule
\texttt{hanoi-single-task1-v0} & \texttt{39} & \texttt{7} & \texttt{1000} \\
\texttt{hanoi-single-task2-v0} & \texttt{39} & \texttt{7} & \texttt{1000} \\
\texttt{hanoi-double-task1-v0} & \texttt{44} & \texttt{7} & \texttt{3000} \\
\texttt{hanoi-double-task2-v0} & \texttt{44} & \texttt{7} & \texttt{3000} \\
\texttt{hanoi-triple-task1-v0} & \texttt{49} & \texttt{7} & \texttt{7000} \\
\texttt{hanoi-triple-task2-v0} & \texttt{49} & \texttt{7} & \texttt{7000} \\
\midrule
\texttt{bowling-task1-v0} & \texttt{102} & \texttt{7} & \texttt{500} \\
\midrule
\texttt{block-lite-single-task1-v0} & \texttt{28} & \texttt{7} & \texttt{200} \\
\texttt{block-lite-double-task1-v0} & \texttt{37} & \texttt{7} & \texttt{400} \\
\texttt{block-lite-double-task2-v0} & \texttt{37} & \texttt{7} & \texttt{400} \\
\texttt{block-lite-triple-task1-v0} & \texttt{46} & \texttt{7} & \texttt{600} \\
\texttt{block-lite-triple-task2-v0} & \texttt{46} & \texttt{7} & \texttt{600} \\
\texttt{block-lite-quadruple-task1-v0} & \texttt{55} & \texttt{7} & \texttt{800} \\
\texttt{block-lite-quadruple-task2-v0} & \texttt{55} & \texttt{7} & \texttt{800} \\
\texttt{block-lite-quadruple-task3-v0} & \texttt{55} & \texttt{7} & \texttt{800} \\
\midrule
\texttt{chamber-lite-easy-task1-v0} & \texttt{53} & \texttt{7} & \texttt{500} \\
\texttt{chamber-lite-easy-task2-v0} & \texttt{53} & \texttt{7} & \texttt{500} \\
\texttt{chamber-lite-easy-task3-v0} & \texttt{53} & \texttt{7} & \texttt{500} \\
\texttt{chamber-lite-medium-task1-v0} & \texttt{62} & \texttt{7} & \texttt{600} \\
\texttt{chamber-lite-medium-task2-v0} & \texttt{62} & \texttt{7} & \texttt{600} \\
\texttt{chamber-lite-hard-task1-v0} & \texttt{71} & \texttt{7} & \texttt{700} \\
\texttt{chamber-lite-hard-task2-v0} & \texttt{71} & \texttt{7} & \texttt{700} \\
\midrule
\texttt{switch-lite-3x3-task1-v0} & \texttt{64} & \texttt{7} & \texttt{300} \\
\texttt{switch-lite-3x3-task2-v0} & \texttt{64} & \texttt{7} & \texttt{300} \\
\texttt{switch-lite-4x4-task1-v0} & \texttt{99} & \texttt{7} & \texttt{300} \\
\texttt{switch-lite-4x4-task2-v0} & \texttt{99} & \texttt{7} & \texttt{300} \\
\texttt{switch-lite-5x5-task1-v0} & \texttt{144} & \texttt{7} & \texttt{800} \\
\texttt{switch-lite-5x5-task2-v0} & \texttt{144} & \texttt{7} & \texttt{800} \\
\midrule
\texttt{hanoi-lite-single-task1-v0} & \texttt{39} & \texttt{7} & \texttt{300} \\
\texttt{hanoi-lite-single-task2-v0} & \texttt{39} & \texttt{7} & \texttt{300} \\
\texttt{hanoi-lite-double-task1-v0} & \texttt{44} & \texttt{7} & \texttt{900} \\
\texttt{hanoi-lite-double-task2-v0} & \texttt{44} & \texttt{7} & \texttt{900} \\
\texttt{hanoi-lite-triple-task1-v0} & \texttt{49} & \texttt{7} & \texttt{2100} \\
\texttt{hanoi-lite-triple-task2-v0} & \texttt{49} & \texttt{7} & \texttt{2100} \\
\bottomrule
\end{tabular}
}
\end{table}

\begin{table}[t!]
\caption{
\footnotesize
\textbf{Pixel-based task specifications.}
}
\label{table:visual_task_specs}
\centering
\small
\setlength{\tabcolsep}{4pt}
\scalebox{0.84}{%
\begin{tabular}{lrrr}
\toprule
\texttt{Task} & \texttt{State dimensions} & \texttt{Action dimensions} & \texttt{Maximum episode length} \\
\midrule
\texttt{visual-block-single-task1-v0} & \texttt{3$\times$224$\times$224$\times$3} & \texttt{7} & \texttt{1250} \\
\texttt{visual-block-double-task1-v0} & \texttt{3$\times$224$\times$224$\times$3} & \texttt{7} & \texttt{2500} \\
\texttt{visual-block-double-task2-v0} & \texttt{3$\times$224$\times$224$\times$3} & \texttt{7} & \texttt{2500} \\
\texttt{visual-block-triple-task1-v0} & \texttt{3$\times$224$\times$224$\times$3} & \texttt{7} & \texttt{3750} \\
\texttt{visual-block-triple-task2-v0} & \texttt{3$\times$224$\times$224$\times$3} & \texttt{7} & \texttt{3750} \\
\texttt{visual-block-quadruple-task1-v0} & \texttt{3$\times$224$\times$224$\times$3} & \texttt{7} & \texttt{5000} \\
\texttt{visual-block-quadruple-task2-v0} & \texttt{3$\times$224$\times$224$\times$3} & \texttt{7} & \texttt{5000} \\
\texttt{visual-block-quadruple-task3-v0} & \texttt{3$\times$224$\times$224$\times$3} & \texttt{7} & \texttt{5000} \\
\midrule
\texttt{visual-chamber-easy-task1-v0} & \texttt{3$\times$224$\times$224$\times$3} & \texttt{7} & \texttt{4000} \\
\texttt{visual-chamber-easy-task2-v0} & \texttt{3$\times$224$\times$224$\times$3} & \texttt{7} & \texttt{4000} \\
\texttt{visual-chamber-easy-task3-v0} & \texttt{3$\times$224$\times$224$\times$3} & \texttt{7} & \texttt{4000} \\
\texttt{visual-chamber-medium-task1-v0} & \texttt{3$\times$224$\times$224$\times$3} & \texttt{7} & \texttt{5000} \\
\texttt{visual-chamber-medium-task2-v0} & \texttt{3$\times$224$\times$224$\times$3} & \texttt{7} & \texttt{5000} \\
\texttt{visual-chamber-hard-task1-v0} & \texttt{3$\times$224$\times$224$\times$3} & \texttt{7} & \texttt{5500} \\
\texttt{visual-chamber-hard-task2-v0} & \texttt{3$\times$224$\times$224$\times$3} & \texttt{7} & \texttt{5500} \\
\midrule
\texttt{visual-switch-3x3-task1-v0} & \texttt{3$\times$224$\times$224$\times$3} & \texttt{7} & \texttt{1500} \\
\texttt{visual-switch-3x3-task2-v0} & \texttt{3$\times$224$\times$224$\times$3} & \texttt{7} & \texttt{1500} \\
\texttt{visual-switch-4x4-task1-v0} & \texttt{3$\times$224$\times$224$\times$3} & \texttt{7} & \texttt{1500} \\
\texttt{visual-switch-4x4-task2-v0} & \texttt{3$\times$224$\times$224$\times$3} & \texttt{7} & \texttt{1500} \\
\texttt{visual-switch-5x5-task1-v0} & \texttt{3$\times$224$\times$224$\times$3} & \texttt{7} & \texttt{4000} \\
\texttt{visual-switch-5x5-task2-v0} & \texttt{3$\times$224$\times$224$\times$3} & \texttt{7} & \texttt{4000} \\
\midrule
\texttt{visual-hanoi-single-task1-v0} & \texttt{3$\times$224$\times$224$\times$3} & \texttt{7} & \texttt{1000} \\
\texttt{visual-hanoi-single-task2-v0} & \texttt{3$\times$224$\times$224$\times$3} & \texttt{7} & \texttt{1000} \\
\texttt{visual-hanoi-double-task1-v0} & \texttt{3$\times$224$\times$224$\times$3} & \texttt{7} & \texttt{3000} \\
\texttt{visual-hanoi-double-task2-v0} & \texttt{3$\times$224$\times$224$\times$3} & \texttt{7} & \texttt{3000} \\
\texttt{visual-hanoi-triple-task1-v0} & \texttt{3$\times$224$\times$224$\times$3} & \texttt{7} & \texttt{7000} \\
\texttt{visual-hanoi-triple-task2-v0} & \texttt{3$\times$224$\times$224$\times$3} & \texttt{7} & \texttt{7000} \\
\midrule
\texttt{visual-bowling-task1-v0} & \texttt{3$\times$224$\times$224$\times$3} & \texttt{7} & \texttt{500} \\
\midrule
\texttt{visual-block-lite-single-task1-v0} & \texttt{3$\times$224$\times$224$\times$3} & \texttt{7} & \texttt{200} \\
\texttt{visual-block-lite-double-task1-v0} & \texttt{3$\times$224$\times$224$\times$3} & \texttt{7} & \texttt{400} \\
\texttt{visual-block-lite-double-task2-v0} & \texttt{3$\times$224$\times$224$\times$3} & \texttt{7} & \texttt{400} \\
\texttt{visual-block-lite-triple-task1-v0} & \texttt{3$\times$224$\times$224$\times$3} & \texttt{7} & \texttt{600} \\
\texttt{visual-block-lite-triple-task2-v0} & \texttt{3$\times$224$\times$224$\times$3} & \texttt{7} & \texttt{600} \\
\texttt{visual-block-lite-quadruple-task1-v0} & \texttt{3$\times$224$\times$224$\times$3} & \texttt{7} & \texttt{800} \\
\texttt{visual-block-lite-quadruple-task2-v0} & \texttt{3$\times$224$\times$224$\times$3} & \texttt{7} & \texttt{800} \\
\texttt{visual-block-lite-quadruple-task3-v0} & \texttt{3$\times$224$\times$224$\times$3} & \texttt{7} & \texttt{800} \\
\midrule
\texttt{visual-chamber-lite-easy-task1-v0} & \texttt{3$\times$224$\times$224$\times$3} & \texttt{7} & \texttt{500} \\
\texttt{visual-chamber-lite-easy-task2-v0} & \texttt{3$\times$224$\times$224$\times$3} & \texttt{7} & \texttt{500} \\
\texttt{visual-chamber-lite-easy-task3-v0} & \texttt{3$\times$224$\times$224$\times$3} & \texttt{7} & \texttt{500} \\
\texttt{visual-chamber-lite-medium-task1-v0} & \texttt{3$\times$224$\times$224$\times$3} & \texttt{7} & \texttt{600} \\
\texttt{visual-chamber-lite-medium-task2-v0} & \texttt{3$\times$224$\times$224$\times$3} & \texttt{7} & \texttt{600} \\
\texttt{visual-chamber-lite-hard-task1-v0} & \texttt{3$\times$224$\times$224$\times$3} & \texttt{7} & \texttt{700} \\
\texttt{visual-chamber-lite-hard-task2-v0} & \texttt{3$\times$224$\times$224$\times$3} & \texttt{7} & \texttt{700} \\
\midrule
\texttt{visual-switch-lite-3x3-task1-v0} & \texttt{3$\times$224$\times$224$\times$3} & \texttt{7} & \texttt{300} \\
\texttt{visual-switch-lite-3x3-task2-v0} & \texttt{3$\times$224$\times$224$\times$3} & \texttt{7} & \texttt{300} \\
\texttt{visual-switch-lite-4x4-task1-v0} & \texttt{3$\times$224$\times$224$\times$3} & \texttt{7} & \texttt{300} \\
\texttt{visual-switch-lite-4x4-task2-v0} & \texttt{3$\times$224$\times$224$\times$3} & \texttt{7} & \texttt{300} \\
\texttt{visual-switch-lite-5x5-task1-v0} & \texttt{3$\times$224$\times$224$\times$3} & \texttt{7} & \texttt{800} \\
\texttt{visual-switch-lite-5x5-task2-v0} & \texttt{3$\times$224$\times$224$\times$3} & \texttt{7} & \texttt{800} \\
\midrule
\texttt{visual-hanoi-lite-single-task1-v0} & \texttt{3$\times$224$\times$224$\times$3} & \texttt{7} & \texttt{300} \\
\texttt{visual-hanoi-lite-single-task2-v0} & \texttt{3$\times$224$\times$224$\times$3} & \texttt{7} & \texttt{300} \\
\texttt{visual-hanoi-lite-double-task1-v0} & \texttt{3$\times$224$\times$224$\times$3} & \texttt{7} & \texttt{900} \\
\texttt{visual-hanoi-lite-double-task2-v0} & \texttt{3$\times$224$\times$224$\times$3} & \texttt{7} & \texttt{900} \\
\texttt{visual-hanoi-lite-triple-task1-v0} & \texttt{3$\times$224$\times$224$\times$3} & \texttt{7} & \texttt{2100} \\
\texttt{visual-hanoi-lite-triple-task2-v0} & \texttt{3$\times$224$\times$224$\times$3} & \texttt{7} & \texttt{2100} \\
\bottomrule
\end{tabular}
}
\end{table}

\begin{table}[t!]
\caption{
\footnotesize
\textbf{Dataset specifications.}
}
\label{table:dataset_specs}
\centering
\small
\setlength{\tabcolsep}{4pt}
\scalebox{0.84}{%
\begin{threeparttable}
\begin{tabular}{lrrrrr}
\toprule
\texttt{Task} & \texttt{Episodes} & \texttt{Transitions} & \texttt{Mean episode length} & \texttt{Success rate} & \texttt{Throughput} \\
\midrule
\texttt{block-single-task1-v0} & \texttt{10K} & \texttt{7.3M} & \texttt{729.3} & \texttt{88} & \texttt{1.51} \\
\texttt{block-double-task1-v0} & \texttt{30K} & \texttt{53.4M} & \texttt{1778.9} & \texttt{79} & \texttt{1.11} \\
\texttt{block-double-task2-v0} & \texttt{30K} & \texttt{45.3M} & \texttt{1511.5} & \texttt{88} & \texttt{1.45} \\
\texttt{block-triple-task1-v0} & \texttt{100K} & \texttt{287.3M} & \texttt{2873.0} & \texttt{72} & \texttt{0.94} \\
\texttt{block-triple-task2-v0} & \texttt{100K} & \texttt{279.0M} & \texttt{2790.5} & \texttt{72} & \texttt{0.96} \\
\texttt{block-quadruple-task1-v0} & \texttt{100K} & \texttt{405.7M} & \texttt{4056.8} & \texttt{65} & \texttt{0.80} \\
\texttt{block-quadruple-task2-v0} & \texttt{100K} & \texttt{403.3M} & \texttt{4033.1} & \texttt{60} & \texttt{0.75} \\
\texttt{block-quadruple-task3-v0} & \texttt{100K} & \texttt{452.8M} & \texttt{4527.7} & \texttt{33} & \texttt{0.37} \\
\midrule
\texttt{chamber-easy-task1-v0} & \texttt{10K} & \texttt{19.7M} & \texttt{1968.5} & \texttt{96} & \texttt{1.96} \\
\texttt{chamber-easy-task2-v0} & \texttt{10K} & \texttt{33.0M} & \texttt{3295.4} & \texttt{65} & \texttt{0.79} \\
\texttt{chamber-easy-task3-v0} & \texttt{10K} & \texttt{33.8M} & \texttt{3377.6} & \texttt{58} & \texttt{0.69} \\
\texttt{chamber-medium-task1-v0} & \texttt{30K} & \texttt{128.7M} & \texttt{4288.8} & \texttt{58} & \texttt{0.67} \\
\texttt{chamber-medium-task2-v0} & \texttt{30K} & \texttt{134.0M} & \texttt{4467.8} & \texttt{44} & \texttt{0.49} \\
\texttt{chamber-hard-task1-v0} & \texttt{100K} & \texttt{502.8M} & \texttt{5028.0} & \texttt{42} & \texttt{0.46} \\
\texttt{chamber-hard-task2-v0} & \texttt{100K} & \texttt{490.3M} & \texttt{4902.9} & \texttt{42} & \texttt{0.47} \\
\midrule
\texttt{switch-3x3-task1-v0} & \texttt{10K} & \texttt{14.3M} & \texttt{1433.9} & \texttt{11} & \texttt{0.11} \\
\texttt{switch-3x3-task2-v0} & \texttt{10K} & \texttt{13.5M} & \texttt{1352.9} & \texttt{21} & \texttt{0.24} \\
\texttt{switch-4x4-task1-v0} & \texttt{30K} & \texttt{43.6M} & \texttt{1454.1} & \texttt{9} & \texttt{0.10} \\
\texttt{switch-4x4-task2-v0} & \texttt{30K} & \texttt{43.9M} & \texttt{1463.6} & \texttt{8} & \texttt{0.08} \\
\texttt{switch-5x5-task1-v0} & \texttt{100K} & \texttt{395.1M} & \texttt{3951.0} & \texttt{5} & \texttt{0.05} \\
\texttt{switch-5x5-task2-v0} & \texttt{100K} & \texttt{395.7M} & \texttt{3956.7} & \texttt{4} & \texttt{0.05} \\
\midrule
\texttt{hanoi-single-task1-v0} & \texttt{10K} & \texttt{7.1M} & \texttt{706.8} & \texttt{100} & \texttt{1.41} \\
\texttt{hanoi-single-task2-v0} & \texttt{10K} & \texttt{7.1M} & \texttt{712.1} & \texttt{99} & \texttt{1.40} \\
\texttt{hanoi-double-task1-v0} & \texttt{30K} & \texttt{77.9M} & \texttt{2595.3} & \texttt{71} & \texttt{0.82} \\
\texttt{hanoi-double-task2-v0} & \texttt{30K} & \texttt{78.5M} & \texttt{2615.1} & \texttt{69} & \texttt{0.79} \\
\texttt{hanoi-triple-task1-v0} & \texttt{100K} & \texttt{618.2M} & \texttt{6181.6} & \texttt{58} & \texttt{0.65} \\
\texttt{hanoi-triple-task2-v0} & \texttt{100K} & \texttt{618.9M} & \texttt{6188.9} & \texttt{53} & \texttt{0.60} \\
\midrule
\texttt{bowling-task1-v0} & \texttt{10K} & \texttt{5.0M} & \texttt{500.0} & \texttt{30} & \texttt{-} \\
\midrule
\texttt{block-lite-single-task1-v0} & \texttt{10K} & \texttt{0.8M} & \texttt{77.2} & \texttt{97} & \texttt{2.50} \\
\texttt{block-lite-double-task1-v0} & \texttt{30K} & \texttt{5.8M} & \texttt{193.7} & \texttt{95} & \texttt{1.96} \\
\texttt{block-lite-double-task2-v0} & \texttt{30K} & \texttt{4.8M} & \texttt{159.6} & \texttt{98} & \texttt{2.45} \\
\texttt{block-lite-triple-task1-v0} & \texttt{100K} & \texttt{31.6M} & \texttt{315.7} & \texttt{94} & \texttt{1.79} \\
\texttt{block-lite-triple-task2-v0} & \texttt{100K} & \texttt{32.1M} & \texttt{321.1} & \texttt{89} & \texttt{1.66} \\
\texttt{block-lite-quadruple-task1-v0} & \texttt{100K} & \texttt{45.5M} & \texttt{454.8} & \texttt{93} & \texttt{1.63} \\
\texttt{block-lite-quadruple-task2-v0} & \texttt{100K} & \texttt{48.9M} & \texttt{489.0} & \texttt{80} & \texttt{1.30} \\
\texttt{block-lite-quadruple-task3-v0} & \texttt{100K} & \texttt{62.9M} & \texttt{628.7} & \texttt{52} & \texttt{0.66} \\
\midrule
\texttt{chamber-lite-easy-task1-v0} & \texttt{10K} & \texttt{2.1M} & \texttt{210.5} & \texttt{96} & \texttt{2.28} \\
\texttt{chamber-lite-easy-task2-v0} & \texttt{10K} & \texttt{3.6M} & \texttt{359.0} & \texttt{82} & \texttt{1.14} \\
\texttt{chamber-lite-easy-task3-v0} & \texttt{10K} & \texttt{3.8M} & \texttt{377.6} & \texttt{72} & \texttt{0.96} \\
\texttt{chamber-lite-medium-task1-v0} & \texttt{30K} & \texttt{14.0M} & \texttt{467.1} & \texttt{74} & \texttt{0.95} \\
\texttt{chamber-lite-medium-task2-v0} & \texttt{30K} & \texttt{15.1M} & \texttt{502.2} & \texttt{59} & \texttt{0.70} \\
\texttt{chamber-lite-hard-task1-v0} & \texttt{100K} & \texttt{57.8M} & \texttt{577.8} & \texttt{65} & \texttt{0.78} \\
\texttt{chamber-lite-hard-task2-v0} & \texttt{100K} & \texttt{56.9M} & \texttt{569.4} & \texttt{60} & \texttt{0.73} \\
\midrule
\texttt{switch-lite-3x3-task1-v0} & \texttt{10K} & \texttt{2.7M} & \texttt{267.6} & \texttt{22} & \texttt{0.25} \\
\texttt{switch-lite-3x3-task2-v0} & \texttt{10K} & \texttt{2.4M} & \texttt{243.9} & \texttt{34} & \texttt{0.42} \\
\texttt{switch-lite-4x4-task1-v0} & \texttt{30K} & \texttt{8.1M} & \texttt{271.7} & \texttt{21} & \texttt{0.24} \\
\texttt{switch-lite-4x4-task2-v0} & \texttt{30K} & \texttt{8.3M} & \texttt{276.7} & \texttt{18} & \texttt{0.20} \\
\texttt{switch-lite-5x5-task1-v0} & \texttt{100K} & \texttt{74.8M} & \texttt{748.0} & \texttt{18} & \texttt{0.19} \\
\texttt{switch-lite-5x5-task2-v0} & \texttt{100K} & \texttt{75.5M} & \texttt{754.5} & \texttt{16} & \texttt{0.17} \\
\midrule
\texttt{hanoi-lite-single-task1-v0} & \texttt{10K} & \texttt{1.4M} & \texttt{141.5} & \texttt{100} & \texttt{2.12} \\
\texttt{hanoi-lite-single-task2-v0} & \texttt{10K} & \texttt{1.5M} & \texttt{145.9} & \texttt{97} & \texttt{1.99} \\
\texttt{hanoi-lite-double-task1-v0} & \texttt{30K} & \texttt{15.2M} & \texttt{506.7} & \texttt{94} & \texttt{1.68} \\
\texttt{hanoi-lite-double-task2-v0} & \texttt{30K} & \texttt{15.5M} & \texttt{516.4} & \texttt{89} & \texttt{1.56} \\
\texttt{hanoi-lite-triple-task1-v0} & \texttt{100K} & \texttt{122.5M} & \texttt{1224.8} & \texttt{75} & \texttt{1.29} \\
\texttt{hanoi-lite-triple-task2-v0} & \texttt{100K} & \texttt{123.4M} & \texttt{1234.2} & \texttt{70} & \texttt{1.20} \\
\bottomrule
\end{tabular}
\end{threeparttable}
}
\end{table}

\begin{table}[t!]
\caption{
\footnotesize
\textbf{Full benchmark results (success rate).}
}
\label{table:benchmark_sr}
\centering
\small
\setlength{\tabcolsep}{4pt}
\scalebox{0.84}{%
\begin{tabular}{lrrr}
\toprule
\texttt{Task} & \texttt{BC} & \texttt{FBC} & \texttt{FSQBC} \\
\midrule
\texttt{block-single-task1-v0} & \texttt{88} {\tiny $\pm$\texttt{0}} & \texttt{90} {\tiny $\pm$\texttt{1}} & \texttt{89} {\tiny $\pm$\texttt{0}} \\
\texttt{block-double-task1-v0} & \texttt{51} {\tiny $\pm$\texttt{5}} & \texttt{60} {\tiny $\pm$\texttt{1}} & \texttt{72} {\tiny $\pm$\texttt{0}} \\
\texttt{block-double-task2-v0} & \texttt{68} {\tiny $\pm$\texttt{3}} & \texttt{71} {\tiny $\pm$\texttt{2}} & \texttt{74} {\tiny $\pm$\texttt{2}} \\
\texttt{block-triple-task1-v0} & \texttt{1} {\tiny $\pm$\texttt{0}} & \texttt{2} {\tiny $\pm$\texttt{1}} & \texttt{29} {\tiny $\pm$\texttt{2}} \\
\texttt{block-triple-task2-v0} & \texttt{27} {\tiny $\pm$\texttt{2}} & \texttt{8} {\tiny $\pm$\texttt{1}} & \texttt{30} {\tiny $\pm$\texttt{8}} \\
\texttt{block-quadruple-task1-v0} & \texttt{0} {\tiny $\pm$\texttt{0}} & \texttt{0} {\tiny $\pm$\texttt{0}} & \texttt{0} {\tiny $\pm$\texttt{0}} \\
\texttt{block-quadruple-task2-v0} & \texttt{3} {\tiny $\pm$\texttt{0}} & \texttt{0} {\tiny $\pm$\texttt{0}} & \texttt{6} {\tiny $\pm$\texttt{2}} \\
\texttt{block-quadruple-task3-v0} & \texttt{2} {\tiny $\pm$\texttt{0}} & \texttt{8} {\tiny $\pm$\texttt{3}} & \texttt{25} {\tiny $\pm$\texttt{2}} \\
\midrule
\texttt{chamber-easy-task1-v0} & \texttt{65} {\tiny $\pm$\texttt{8}} & \texttt{73} {\tiny $\pm$\texttt{9}} & \texttt{88} {\tiny $\pm$\texttt{5}} \\
\texttt{chamber-easy-task2-v0} & \texttt{40} {\tiny $\pm$\texttt{0}} & \texttt{53} {\tiny $\pm$\texttt{0}} & \texttt{48} {\tiny $\pm$\texttt{5}} \\
\texttt{chamber-easy-task3-v0} & \texttt{42} {\tiny $\pm$\texttt{0}} & \texttt{48} {\tiny $\pm$\texttt{6}} & \texttt{38} {\tiny $\pm$\texttt{4}} \\
\texttt{chamber-medium-task1-v0} & \texttt{7} {\tiny $\pm$\texttt{1}} & \texttt{12} {\tiny $\pm$\texttt{0}} & \texttt{27} {\tiny $\pm$\texttt{1}} \\
\texttt{chamber-medium-task2-v0} & \texttt{10} {\tiny $\pm$\texttt{3}} & \texttt{5} {\tiny $\pm$\texttt{0}} & \texttt{23} {\tiny $\pm$\texttt{1}} \\
\texttt{chamber-hard-task1-v0} & \texttt{0} {\tiny $\pm$\texttt{0}} & \texttt{1} {\tiny $\pm$\texttt{1}} & \texttt{2} {\tiny $\pm$\texttt{0}} \\
\texttt{chamber-hard-task2-v0} & \texttt{1} {\tiny $\pm$\texttt{0}} & \texttt{0} {\tiny $\pm$\texttt{0}} & \texttt{8} {\tiny $\pm$\texttt{4}} \\
\midrule
\texttt{switch-3x3-task1-v0} & \texttt{21} {\tiny $\pm$\texttt{0}} & \texttt{16} {\tiny $\pm$\texttt{1}} & \texttt{20} {\tiny $\pm$\texttt{2}} \\
\texttt{switch-3x3-task2-v0} & \texttt{25} {\tiny $\pm$\texttt{3}} & \texttt{16} {\tiny $\pm$\texttt{0}} & \texttt{19} {\tiny $\pm$\texttt{3}} \\
\texttt{switch-4x4-task1-v0} & \texttt{8} {\tiny $\pm$\texttt{0}} & \texttt{5} {\tiny $\pm$\texttt{0}} & \texttt{6} {\tiny $\pm$\texttt{0}} \\
\texttt{switch-4x4-task2-v0} & \texttt{2} {\tiny $\pm$\texttt{0}} & \texttt{1} {\tiny $\pm$\texttt{0}} & \texttt{1} {\tiny $\pm$\texttt{0}} \\
\texttt{switch-5x5-task1-v0} & \texttt{0} {\tiny $\pm$\texttt{0}} & \texttt{0} {\tiny $\pm$\texttt{0}} & \texttt{0} {\tiny $\pm$\texttt{0}} \\
\texttt{switch-5x5-task2-v0} & \texttt{0} {\tiny $\pm$\texttt{0}} & \texttt{0} {\tiny $\pm$\texttt{0}} & \texttt{0} {\tiny $\pm$\texttt{0}} \\
\midrule
\texttt{hanoi-single-task1-v0} & \texttt{56} {\tiny $\pm$\texttt{1}} & \texttt{41} {\tiny $\pm$\texttt{31}} & \texttt{84} {\tiny $\pm$\texttt{10}} \\
\texttt{hanoi-single-task2-v0} & \texttt{62} {\tiny $\pm$\texttt{1}} & \texttt{74} {\tiny $\pm$\texttt{0}} & \texttt{66} {\tiny $\pm$\texttt{2}} \\
\texttt{hanoi-double-task1-v0} & \texttt{23} {\tiny $\pm$\texttt{2}} & \texttt{21} {\tiny $\pm$\texttt{3}} & \texttt{20} {\tiny $\pm$\texttt{2}} \\
\texttt{hanoi-double-task2-v0} & \texttt{28} {\tiny $\pm$\texttt{2}} & \texttt{16} {\tiny $\pm$\texttt{11}} & \texttt{18} {\tiny $\pm$\texttt{3}} \\
\texttt{hanoi-triple-task1-v0} & \texttt{18} {\tiny $\pm$\texttt{0}} & \texttt{4} {\tiny $\pm$\texttt{2}} & \texttt{1} {\tiny $\pm$\texttt{0}} \\
\texttt{hanoi-triple-task2-v0} & \texttt{14} {\tiny $\pm$\texttt{1}} & \texttt{8} {\tiny $\pm$\texttt{1}} & \texttt{2} {\tiny $\pm$\texttt{0}} \\
\midrule
\texttt{bowling-task1-v0} & \texttt{39} {\tiny $\pm$\texttt{1}} & \texttt{35} {\tiny $\pm$\texttt{1}} & \texttt{32} {\tiny $\pm$\texttt{1}} \\
\midrule
\texttt{block-lite-single-task1-v0} & \texttt{93} {\tiny $\pm$\texttt{0}} & \texttt{64} {\tiny $\pm$\texttt{9}} & \texttt{96} {\tiny $\pm$\texttt{0}} \\
\texttt{block-lite-double-task1-v0} & \texttt{26} {\tiny $\pm$\texttt{6}} & \texttt{41} {\tiny $\pm$\texttt{2}} & \texttt{73} {\tiny $\pm$\texttt{2}} \\
\texttt{block-lite-double-task2-v0} & \texttt{46} {\tiny $\pm$\texttt{5}} & \texttt{53} {\tiny $\pm$\texttt{3}} & \texttt{74} {\tiny $\pm$\texttt{8}} \\
\texttt{block-lite-triple-task1-v0} & \texttt{1} {\tiny $\pm$\texttt{1}} & \texttt{1} {\tiny $\pm$\texttt{1}} & \texttt{0} {\tiny $\pm$\texttt{0}} \\
\texttt{block-lite-triple-task2-v0} & \texttt{7} {\tiny $\pm$\texttt{1}} & \texttt{22} {\tiny $\pm$\texttt{6}} & \texttt{8} {\tiny $\pm$\texttt{2}} \\
\texttt{block-lite-quadruple-task1-v0} & \texttt{0} {\tiny $\pm$\texttt{0}} & \texttt{0} {\tiny $\pm$\texttt{0}} & \texttt{0} {\tiny $\pm$\texttt{0}} \\
\texttt{block-lite-quadruple-task2-v0} & \texttt{0} {\tiny $\pm$\texttt{0}} & \texttt{0} {\tiny $\pm$\texttt{0}} & \texttt{0} {\tiny $\pm$\texttt{0}} \\
\texttt{block-lite-quadruple-task3-v0} & \texttt{0} {\tiny $\pm$\texttt{0}} & \texttt{4} {\tiny $\pm$\texttt{4}} & \texttt{3} {\tiny $\pm$\texttt{1}} \\
\midrule
\texttt{chamber-lite-easy-task1-v0} & \texttt{27} {\tiny $\pm$\texttt{2}} & \texttt{78} {\tiny $\pm$\texttt{9}} & \texttt{66} {\tiny $\pm$\texttt{4}} \\
\texttt{chamber-lite-easy-task2-v0} & \texttt{11} {\tiny $\pm$\texttt{0}} & \texttt{35} {\tiny $\pm$\texttt{4}} & \texttt{22} {\tiny $\pm$\texttt{5}} \\
\texttt{chamber-lite-easy-task3-v0} & \texttt{9} {\tiny $\pm$\texttt{1}} & \texttt{26} {\tiny $\pm$\texttt{6}} & \texttt{17} {\tiny $\pm$\texttt{4}} \\
\texttt{chamber-lite-medium-task1-v0} & \texttt{1} {\tiny $\pm$\texttt{0}} & \texttt{3} {\tiny $\pm$\texttt{1}} & \texttt{1} {\tiny $\pm$\texttt{0}} \\
\texttt{chamber-lite-medium-task2-v0} & \texttt{0} {\tiny $\pm$\texttt{0}} & \texttt{2} {\tiny $\pm$\texttt{1}} & \texttt{0} {\tiny $\pm$\texttt{0}} \\
\texttt{chamber-lite-hard-task1-v0} & \texttt{0} {\tiny $\pm$\texttt{0}} & \texttt{0} {\tiny $\pm$\texttt{0}} & \texttt{0} {\tiny $\pm$\texttt{0}} \\
\texttt{chamber-lite-hard-task2-v0} & \texttt{0} {\tiny $\pm$\texttt{0}} & \texttt{0} {\tiny $\pm$\texttt{0}} & \texttt{0} {\tiny $\pm$\texttt{0}} \\
\midrule
\texttt{switch-lite-3x3-task1-v0} & \texttt{2} {\tiny $\pm$\texttt{1}} & \texttt{6} {\tiny $\pm$\texttt{1}} & \texttt{1} {\tiny $\pm$\texttt{1}} \\
\texttt{switch-lite-3x3-task2-v0} & \texttt{4} {\tiny $\pm$\texttt{0}} & \texttt{6} {\tiny $\pm$\texttt{1}} & \texttt{3} {\tiny $\pm$\texttt{1}} \\
\texttt{switch-lite-4x4-task1-v0} & \texttt{1} {\tiny $\pm$\texttt{0}} & \texttt{2} {\tiny $\pm$\texttt{1}} & \texttt{1} {\tiny $\pm$\texttt{0}} \\
\texttt{switch-lite-4x4-task2-v0} & \texttt{1} {\tiny $\pm$\texttt{0}} & \texttt{1} {\tiny $\pm$\texttt{0}} & \texttt{0} {\tiny $\pm$\texttt{0}} \\
\texttt{switch-lite-5x5-task1-v0} & \texttt{0} {\tiny $\pm$\texttt{0}} & \texttt{0} {\tiny $\pm$\texttt{0}} & \texttt{0} {\tiny $\pm$\texttt{0}} \\
\texttt{switch-lite-5x5-task2-v0} & \texttt{0} {\tiny $\pm$\texttt{0}} & \texttt{0} {\tiny $\pm$\texttt{0}} & \texttt{0} {\tiny $\pm$\texttt{0}} \\
\midrule
\texttt{hanoi-lite-single-task1-v0} & \texttt{59} {\tiny $\pm$\texttt{6}} & \texttt{9} {\tiny $\pm$\texttt{3}} & \texttt{84} {\tiny $\pm$\texttt{0}} \\
\texttt{hanoi-lite-single-task2-v0} & \texttt{67} {\tiny $\pm$\texttt{2}} & \texttt{2} {\tiny $\pm$\texttt{1}} & \texttt{56} {\tiny $\pm$\texttt{2}} \\
\texttt{hanoi-lite-double-task1-v0} & \texttt{9} {\tiny $\pm$\texttt{2}} & \texttt{0} {\tiny $\pm$\texttt{0}} & \texttt{20} {\tiny $\pm$\texttt{1}} \\
\texttt{hanoi-lite-double-task2-v0} & \texttt{0} {\tiny $\pm$\texttt{0}} & \texttt{0} {\tiny $\pm$\texttt{0}} & \texttt{0} {\tiny $\pm$\texttt{0}} \\
\texttt{hanoi-lite-triple-task1-v0} & \texttt{0} {\tiny $\pm$\texttt{0}} & \texttt{0} {\tiny $\pm$\texttt{0}} & \texttt{0} {\tiny $\pm$\texttt{0}} \\
\texttt{hanoi-lite-triple-task2-v0} & \texttt{0} {\tiny $\pm$\texttt{0}} & \texttt{0} {\tiny $\pm$\texttt{0}} & \texttt{0} {\tiny $\pm$\texttt{0}} \\
\bottomrule
\end{tabular}
}
\end{table}

\begin{table}[t!]
\caption{
\footnotesize
\textbf{Full benchmark results (throughput).}
}
\label{table:benchmark_throughput}
\centering
\small
\setlength{\tabcolsep}{4pt}
\scalebox{0.84}{%
\begin{tabular}{lrrr}
\toprule
\texttt{Task} & \texttt{BC} & \texttt{FBC} & \texttt{FSQBC} \\
\midrule
\texttt{block-single-task1-v0} & \texttt{1.41} {\tiny $\pm$\texttt{0.01}} & \texttt{1.45} {\tiny $\pm$\texttt{0.04}} & \texttt{1.40} {\tiny $\pm$\texttt{0.00}} \\
\texttt{block-double-task1-v0} & \texttt{0.58} {\tiny $\pm$\texttt{0.07}} & \texttt{0.71} {\tiny $\pm$\texttt{0.02}} & \texttt{0.88} {\tiny $\pm$\texttt{0.00}} \\
\texttt{block-double-task2-v0} & \texttt{0.95} {\tiny $\pm$\texttt{0.05}} & \texttt{1.00} {\tiny $\pm$\texttt{0.04}} & \texttt{1.07} {\tiny $\pm$\texttt{0.05}} \\
\texttt{block-triple-task1-v0} & \texttt{0.01} {\tiny $\pm$\texttt{0.00}} & \texttt{0.02} {\tiny $\pm$\texttt{0.01}} & \texttt{0.30} {\tiny $\pm$\texttt{0.03}} \\
\texttt{block-triple-task2-v0} & \texttt{0.29} {\tiny $\pm$\texttt{0.02}} & \texttt{0.09} {\tiny $\pm$\texttt{0.01}} & \texttt{0.33} {\tiny $\pm$\texttt{0.10}} \\
\texttt{block-quadruple-task1-v0} & \texttt{0.00} {\tiny $\pm$\texttt{0.00}} & \texttt{0.00} {\tiny $\pm$\texttt{0.00}} & \texttt{0.00} {\tiny $\pm$\texttt{0.00}} \\
\texttt{block-quadruple-task2-v0} & \texttt{0.03} {\tiny $\pm$\texttt{0.00}} & \texttt{0.00} {\tiny $\pm$\texttt{0.00}} & \texttt{0.06} {\tiny $\pm$\texttt{0.02}} \\
\texttt{block-quadruple-task3-v0} & \texttt{0.02} {\tiny $\pm$\texttt{0.00}} & \texttt{0.08} {\tiny $\pm$\texttt{0.03}} & \texttt{0.26} {\tiny $\pm$\texttt{0.02}} \\
\midrule
\texttt{chamber-easy-task1-v0} & \texttt{0.89} {\tiny $\pm$\texttt{0.16}} & \texttt{1.07} {\tiny $\pm$\texttt{0.18}} & \texttt{1.54} {\tiny $\pm$\texttt{0.15}} \\
\texttt{chamber-easy-task2-v0} & \texttt{0.43} {\tiny $\pm$\texttt{0.00}} & \texttt{0.61} {\tiny $\pm$\texttt{0.01}} & \texttt{0.53} {\tiny $\pm$\texttt{0.06}} \\
\texttt{chamber-easy-task3-v0} & \texttt{0.46} {\tiny $\pm$\texttt{0.01}} & \texttt{0.53} {\tiny $\pm$\texttt{0.07}} & \texttt{0.42} {\tiny $\pm$\texttt{0.05}} \\
\texttt{chamber-medium-task1-v0} & \texttt{0.07} {\tiny $\pm$\texttt{0.01}} & \texttt{0.12} {\tiny $\pm$\texttt{0.00}} & \texttt{0.28} {\tiny $\pm$\texttt{0.01}} \\
\texttt{chamber-medium-task2-v0} & \texttt{0.10} {\tiny $\pm$\texttt{0.03}} & \texttt{0.05} {\tiny $\pm$\texttt{0.00}} & \texttt{0.24} {\tiny $\pm$\texttt{0.01}} \\
\texttt{chamber-hard-task1-v0} & \texttt{0.00} {\tiny $\pm$\texttt{0.00}} & \texttt{0.01} {\tiny $\pm$\texttt{0.01}} & \texttt{0.02} {\tiny $\pm$\texttt{0.00}} \\
\texttt{chamber-hard-task2-v0} & \texttt{0.01} {\tiny $\pm$\texttt{0.00}} & \texttt{0.00} {\tiny $\pm$\texttt{0.00}} & \texttt{0.08} {\tiny $\pm$\texttt{0.05}} \\
\midrule
\texttt{switch-3x3-task1-v0} & \texttt{0.23} {\tiny $\pm$\texttt{0.00}} & \texttt{0.17} {\tiny $\pm$\texttt{0.01}} & \texttt{0.22} {\tiny $\pm$\texttt{0.02}} \\
\texttt{switch-3x3-task2-v0} & \texttt{0.27} {\tiny $\pm$\texttt{0.03}} & \texttt{0.17} {\tiny $\pm$\texttt{0.00}} & \texttt{0.21} {\tiny $\pm$\texttt{0.03}} \\
\texttt{switch-4x4-task1-v0} & \texttt{0.08} {\tiny $\pm$\texttt{0.00}} & \texttt{0.05} {\tiny $\pm$\texttt{0.00}} & \texttt{0.06} {\tiny $\pm$\texttt{0.00}} \\
\texttt{switch-4x4-task2-v0} & \texttt{0.02} {\tiny $\pm$\texttt{0.00}} & \texttt{0.01} {\tiny $\pm$\texttt{0.00}} & \texttt{0.01} {\tiny $\pm$\texttt{0.00}} \\
\texttt{switch-5x5-task1-v0} & \texttt{0.00} {\tiny $\pm$\texttt{0.00}} & \texttt{0.00} {\tiny $\pm$\texttt{0.00}} & \texttt{0.00} {\tiny $\pm$\texttt{0.00}} \\
\texttt{switch-5x5-task2-v0} & \texttt{0.00} {\tiny $\pm$\texttt{0.00}} & \texttt{0.00} {\tiny $\pm$\texttt{0.00}} & \texttt{0.00} {\tiny $\pm$\texttt{0.00}} \\
\midrule
\texttt{hanoi-single-task1-v0} & \texttt{0.63} {\tiny $\pm$\texttt{0.02}} & \texttt{0.48} {\tiny $\pm$\texttt{0.38}} & \texttt{1.08} {\tiny $\pm$\texttt{0.16}} \\
\texttt{hanoi-single-task2-v0} & \texttt{0.75} {\tiny $\pm$\texttt{0.02}} & \texttt{0.93} {\tiny $\pm$\texttt{0.00}} & \texttt{0.81} {\tiny $\pm$\texttt{0.03}} \\
\texttt{hanoi-double-task1-v0} & \texttt{0.24} {\tiny $\pm$\texttt{0.03}} & \texttt{0.22} {\tiny $\pm$\texttt{0.03}} & \texttt{0.21} {\tiny $\pm$\texttt{0.02}} \\
\texttt{hanoi-double-task2-v0} & \texttt{0.30} {\tiny $\pm$\texttt{0.02}} & \texttt{0.17} {\tiny $\pm$\texttt{0.12}} & \texttt{0.19} {\tiny $\pm$\texttt{0.03}} \\
\texttt{hanoi-triple-task1-v0} & \texttt{0.21} {\tiny $\pm$\texttt{0.00}} & \texttt{0.05} {\tiny $\pm$\texttt{0.02}} & \texttt{0.01} {\tiny $\pm$\texttt{0.00}} \\
\texttt{hanoi-triple-task2-v0} & \texttt{0.15} {\tiny $\pm$\texttt{0.01}} & \texttt{0.08} {\tiny $\pm$\texttt{0.01}} & \texttt{0.02} {\tiny $\pm$\texttt{0.00}} \\
\midrule
\texttt{block-lite-single-task1-v0} & \texttt{1.93} {\tiny $\pm$\texttt{0.00}} & \texttt{0.93} {\tiny $\pm$\texttt{0.18}} & \texttt{2.07} {\tiny $\pm$\texttt{0.02}} \\
\texttt{block-lite-double-task1-v0} & \texttt{0.28} {\tiny $\pm$\texttt{0.07}} & \texttt{0.48} {\tiny $\pm$\texttt{0.02}} & \texttt{1.02} {\tiny $\pm$\texttt{0.07}} \\
\texttt{block-lite-double-task2-v0} & \texttt{0.59} {\tiny $\pm$\texttt{0.09}} & \texttt{0.72} {\tiny $\pm$\texttt{0.07}} & \texttt{1.19} {\tiny $\pm$\texttt{0.21}} \\
\texttt{block-lite-triple-task1-v0} & \texttt{0.01} {\tiny $\pm$\texttt{0.01}} & \texttt{0.01} {\tiny $\pm$\texttt{0.01}} & \texttt{0.00} {\tiny $\pm$\texttt{0.00}} \\
\texttt{block-lite-triple-task2-v0} & \texttt{0.07} {\tiny $\pm$\texttt{0.01}} & \texttt{0.24} {\tiny $\pm$\texttt{0.07}} & \texttt{0.08} {\tiny $\pm$\texttt{0.02}} \\
\texttt{block-lite-quadruple-task1-v0} & \texttt{0.00} {\tiny $\pm$\texttt{0.00}} & \texttt{0.00} {\tiny $\pm$\texttt{0.00}} & \texttt{0.00} {\tiny $\pm$\texttt{0.00}} \\
\texttt{block-lite-quadruple-task2-v0} & \texttt{0.00} {\tiny $\pm$\texttt{0.00}} & \texttt{0.00} {\tiny $\pm$\texttt{0.00}} & \texttt{0.00} {\tiny $\pm$\texttt{0.00}} \\
\texttt{block-lite-quadruple-task3-v0} & \texttt{0.00} {\tiny $\pm$\texttt{0.00}} & \texttt{0.05} {\tiny $\pm$\texttt{0.04}} & \texttt{0.03} {\tiny $\pm$\texttt{0.01}} \\
\midrule
\texttt{chamber-lite-easy-task1-v0} & \texttt{0.31} {\tiny $\pm$\texttt{0.03}} & \texttt{1.24} {\tiny $\pm$\texttt{0.25}} & \texttt{0.96} {\tiny $\pm$\texttt{0.08}} \\
\texttt{chamber-lite-easy-task2-v0} & \texttt{0.11} {\tiny $\pm$\texttt{0.00}} & \texttt{0.39} {\tiny $\pm$\texttt{0.04}} & \texttt{0.23} {\tiny $\pm$\texttt{0.05}} \\
\texttt{chamber-lite-easy-task3-v0} & \texttt{0.09} {\tiny $\pm$\texttt{0.01}} & \texttt{0.28} {\tiny $\pm$\texttt{0.07}} & \texttt{0.18} {\tiny $\pm$\texttt{0.04}} \\
\texttt{chamber-lite-medium-task1-v0} & \texttt{0.01} {\tiny $\pm$\texttt{0.00}} & \texttt{0.03} {\tiny $\pm$\texttt{0.01}} & \texttt{0.01} {\tiny $\pm$\texttt{0.00}} \\
\texttt{chamber-lite-medium-task2-v0} & \texttt{0.00} {\tiny $\pm$\texttt{0.00}} & \texttt{0.02} {\tiny $\pm$\texttt{0.01}} & \texttt{0.00} {\tiny $\pm$\texttt{0.00}} \\
\texttt{chamber-lite-hard-task1-v0} & \texttt{0.00} {\tiny $\pm$\texttt{0.00}} & \texttt{0.00} {\tiny $\pm$\texttt{0.00}} & \texttt{0.00} {\tiny $\pm$\texttt{0.00}} \\
\texttt{chamber-lite-hard-task2-v0} & \texttt{0.00} {\tiny $\pm$\texttt{0.00}} & \texttt{0.00} {\tiny $\pm$\texttt{0.00}} & \texttt{0.00} {\tiny $\pm$\texttt{0.00}} \\
\midrule
\texttt{switch-lite-3x3-task1-v0} & \texttt{0.02} {\tiny $\pm$\texttt{0.01}} & \texttt{0.06} {\tiny $\pm$\texttt{0.02}} & \texttt{0.01} {\tiny $\pm$\texttt{0.01}} \\
\texttt{switch-lite-3x3-task2-v0} & \texttt{0.04} {\tiny $\pm$\texttt{0.00}} & \texttt{0.07} {\tiny $\pm$\texttt{0.01}} & \texttt{0.04} {\tiny $\pm$\texttt{0.01}} \\
\texttt{switch-lite-4x4-task1-v0} & \texttt{0.01} {\tiny $\pm$\texttt{0.00}} & \texttt{0.02} {\tiny $\pm$\texttt{0.01}} & \texttt{0.01} {\tiny $\pm$\texttt{0.00}} \\
\texttt{switch-lite-4x4-task2-v0} & \texttt{0.01} {\tiny $\pm$\texttt{0.00}} & \texttt{0.01} {\tiny $\pm$\texttt{0.00}} & \texttt{0.00} {\tiny $\pm$\texttt{0.00}} \\
\texttt{switch-lite-5x5-task1-v0} & \texttt{0.00} {\tiny $\pm$\texttt{0.00}} & \texttt{0.00} {\tiny $\pm$\texttt{0.00}} & \texttt{0.00} {\tiny $\pm$\texttt{0.00}} \\
\texttt{switch-lite-5x5-task2-v0} & \texttt{0.00} {\tiny $\pm$\texttt{0.00}} & \texttt{0.00} {\tiny $\pm$\texttt{0.00}} & \texttt{0.00} {\tiny $\pm$\texttt{0.00}} \\
\midrule
\texttt{hanoi-lite-single-task1-v0} & \texttt{0.85} {\tiny $\pm$\texttt{0.14}} & \texttt{0.11} {\tiny $\pm$\texttt{0.04}} & \texttt{1.49} {\tiny $\pm$\texttt{0.00}} \\
\texttt{hanoi-lite-single-task2-v0} & \texttt{1.02} {\tiny $\pm$\texttt{0.05}} & \texttt{0.02} {\tiny $\pm$\texttt{0.01}} & \texttt{0.80} {\tiny $\pm$\texttt{0.03}} \\
\texttt{hanoi-lite-double-task1-v0} & \texttt{0.10} {\tiny $\pm$\texttt{0.02}} & \texttt{0.00} {\tiny $\pm$\texttt{0.00}} & \texttt{0.23} {\tiny $\pm$\texttt{0.01}} \\
\texttt{hanoi-lite-double-task2-v0} & \texttt{0.00} {\tiny $\pm$\texttt{0.00}} & \texttt{0.00} {\tiny $\pm$\texttt{0.00}} & \texttt{0.00} {\tiny $\pm$\texttt{0.00}} \\
\texttt{hanoi-lite-triple-task1-v0} & \texttt{0.00} {\tiny $\pm$\texttt{0.00}} & \texttt{0.00} {\tiny $\pm$\texttt{0.00}} & \texttt{0.00} {\tiny $\pm$\texttt{0.00}} \\
\texttt{hanoi-lite-triple-task2-v0} & \texttt{0.00} {\tiny $\pm$\texttt{0.00}} & \texttt{0.00} {\tiny $\pm$\texttt{0.00}} & \texttt{0.00} {\tiny $\pm$\texttt{0.00}} \\
\bottomrule
\end{tabular}
}
\end{table}

\clearpage

\end{document}